%% file: 3dgen4robot_survey.tex
\documentclass[letterpaper,journal,compsoc]{IEEEtran}

\usepackage{amsmath,amsfonts,amssymb}
\usepackage{algorithmic}
\usepackage{algorithm}
\usepackage{array}
\usepackage[caption=false,font=normalsize,labelfont=sf,textfont=sf]{subfig}
\usepackage{textcomp}
\usepackage{stfloats}
\usepackage{url}
\usepackage{verbatim}
\usepackage{cite}
\usepackage{multirow}
\usepackage{graphicx}
\usepackage{fontawesome5}
\usepackage[table]{xcolor}
\usepackage{pifont}
\usepackage{enumitem}
\usepackage{tabularx}
\usepackage{wrapfig}
\usepackage{tikz}
\usepackage{booktabs}
\usepackage{hyperref}

\newcommand{\cmark}{\ding{51}}%
\newcommand{\xmark}{\ding{55}}%
\newcommand{\pmark}{\ding{115}}%

\begin{document}

\title{3D Generation for Embodied AI \\ and Robotic Simulation: A Survey}

\author{Tianwei Ye, Yifan Mao, Minwen Liao, Jian Liu, Chunchao Guo, \\ Dazhao Du, Quanxin Shou, Fangqi Zhu, and Song Guo%
\IEEEcompsocitemizethanks{%
\IEEEcompsocthanksitem Tianwei Ye, and Minwen Liao performed this work during an internship at HKUST. Jian Liu is the project lead.
\IEEEcompsocthanksitem Tianwei Ye is with the Hong Kong University of Science and Technology and Wuhan University (email: twye2001@gmail.com). Yifan Mao is with Harbin Institute of Technology. Minwen Liao is with the Hong Kong University of Science and Technology and Xinjiang University. Chunchao Guo is an Independent Researcher. Jian Liu, Dazhao Du, Quanxin Shou, Fangqi Zhu, and Song Guo are with the Hong Kong University of Science and Technology.
\IEEEcompsocthanksitem Corresponding author: Song Guo, e-mail: songguo@cse.ust.hk.}%
}


\IEEEtitleabstractindextext{%
\begin{abstract}
Embodied AI and robotic systems increasingly depend on scalable, diverse, and physically grounded 3D content for simulation-based training and real-world deployment. While 3D generative modeling has advanced rapidly, embodied applications impose requirements far beyond visual realism: generated objects must carry kinematic structure and material properties, scenes must support interaction and task execution, and the resulting content must bridge the gap between simulation and reality. This survey reviews 3D generation for embodied AI and organizes the literature around three roles that 3D generation plays in embodied systems. In \emph{Data Generator}, 3D generation produces simulation-ready objects and assets, including articulated, physically grounded, and deformable content for downstream interaction; in \emph{Simulation Environments}, it constructs interactive and task-oriented worlds, spanning structure-aware, controllable, and agentic scene generation; and in \emph{Sim2Real Bridge}, it supports digital twin reconstruction, data augmentation, and synthetic demonstrations for downstream robot learning and real-world transfer. We also show that the field is shifting from visual realism toward interaction readiness, and we identify the main bottlenecks, including limited physical annotations, the gap between geometric quality and physical validity, fragmented evaluation, and the persistent sim-to-real divide, that must be addressed for 3D generation to become a dependable foundation for embodied intelligence. Our project page is at \url{https://3dgen4robot.github.io}.
\end{abstract}

}

\maketitle

\input{sec/introduction}
\input{sec/prelim}
\input{sec/data_generator}

\input{sec/simulation_environments}
\input{sec/sim2real_bridge}
\input{sec/datasets}
\input{sec/challenges}
\input{sec/conclusion}

\bibliographystyle{IEEEtran}
\bibliography{ref}

\input{sec/biography}
\vfill

\end{document}

%% file: sec/introduction.tex
\section{Introduction}
\label{sec:intro}

\begin{figure*}[t]
    \centering
    \includegraphics[width=0.95\linewidth]{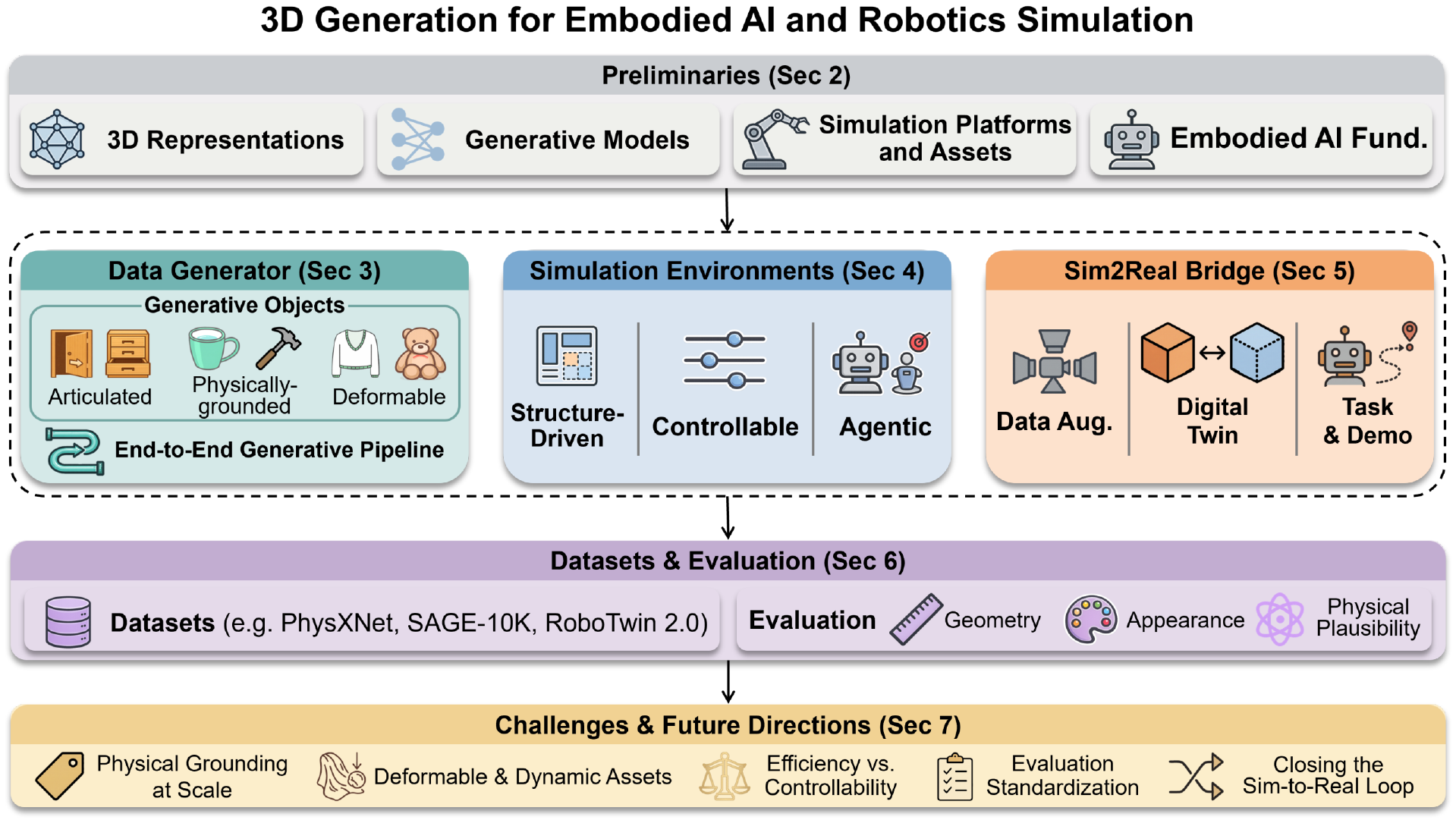}
    \caption{\textbf{Outline of the survey.} Built on \emph{Preliminaries} (Sec.~\ref{sec:pre}), the three core modules---\emph{Data Generator} (Sec.~\ref{sec:data_generator}), \emph{Simulation Environments} (Sec.~\ref{sec:sim_env}), and \emph{Sim2Real Bridge} (Sec.~\ref{sec:sim2real})---address object-level asset creation, scene-level environment synthesis, and the simulation-to-reality transfer loop, respectively. \emph{Datasets \& Evaluation} (Sec.~\ref{sec:dataset}) and \emph{Challenges} (Sec.~\ref{sec:challenges}) complete the survey.}
    \label{fig:outline}
\end{figure*}

\IEEEPARstart{E}{mbodied} AI and robotic systems are increasingly expected to perceive, reason, and act in open-ended physical environments~\cite{liu2025aligning, long2025survey}. Recent progress in large-scale policy learning~\cite{chi2025diffusion, zhao2023learning}, vision-language-action models~\cite{zitkovich2023rt, kim2024openvla, black2024pi0, kim2025fine, intelligence2025pi_}, and high-fidelity simulation~\cite{tao2024maniskill3, makoviychuk2021isaac, chen2025robotwin} has significantly expanded what these systems can do. However, their performance remains fundamentally constrained by the availability of scalable, diverse, and interaction-ready 3D assets and environments. In contrast to conventional 3D generation, which often prioritizes appearance realism or static geometry, embodied applications require assets that can be manipulated, simulated, and transferred across tasks. A cabinet is useful not only because it looks realistic, but because its doors rotate around plausible joints~\cite{le2024articulate, li2025urdf}; a cloth is valuable not only because it has the right shape, but because it deforms under contact~\cite{li2025dress, lu2024garmentlab}; a scene is effective not only because it is visually coherent, but because it supports navigation, interaction, and task execution under physical constraints~\cite{physcene}. These requirements make 3D generation for embodied AI a substantially broader problem than visual content synthesis alone.

This shift has led to a rapid convergence between 3D generative modeling and robotics-oriented simulation. On the one hand, advances in diffusion models, reconstruction pipelines, large language models, and multimodal foundation models have greatly improved the ability to generate geometry, texture, structure, and semantics from sparse inputs such as text, images, or demonstrations~\cite{li2024advances, tang2025recent, wang2025diffusion}. On the other hand, robotic simulation places additional demands on generated outputs, including kinematic structure, physical parameters, material properties, affordance-related semantics, and compatibility with execution formats such as URDF, MJCF, and simulator-native representations. Concretely, methods such as URDF-Anything~\cite{li2025urdf} and PhysX-3D~\cite{cao2025physx3d} demonstrate that useful generated assets must encode not just visual appearance but also joint configurations, mass distributions, and friction coefficients that enable physical simulation. As a result, the central question is no longer simply how to generate plausible 3D content, but how to generate \emph{simulation-ready} 3D content that supports embodied perception, planning, control, and sim-to-real transfer.

Existing literature on this topic is growing quickly, but it remains fragmented across multiple communities, including computer vision, computer graphics, robotics, embodied AI, and simulation systems. Prior work often studies only one layer of the problem: object generation without downstream executability~\cite{lei2023nap, liu2024cage}, scene generation without interaction semantics~\cite{wen20253d}, or simulation platforms without generative scalability~\cite{tao2024maniskill3}. Meanwhile, recent methods increasingly blur these boundaries by combining asset generation, scene synthesis, simulator feedback, and agentic planning into unified pipelines~\cite{wang2025embodiedgen, seed2025seed3d}. This makes it timely to revisit the field from the perspective of embodied use, focusing not only on what can be generated, but also on whether the generated results are controllable, interactive, physically grounded, and practically useful for robotic simulation.

In this survey, we present a structured review of 3D generation for embodied AI and robotic simulation. Rather than organizing the literature purely by generative backbone, we center the survey on a single question: \emph{what roles does 3D generation play in embodied AI?} This perspective leads to three core sections that also define our taxonomy. \emph{Data Generator} (Sec.~\ref{sec:data_generator}) treats 3D generation as the engine for producing simulation-ready objects and assets, including articulated, physically grounded, and deformable content as well as end-to-end pipelines from raw inputs to executable formats. \emph{Simulation Environments} (Sec.~\ref{sec:sim_env}) treats 3D generation as the mechanism for building interactive worlds, tracing the evolution from structure-driven scene synthesis to controllable and agentic environment generation. \emph{Sim2Real Bridge} (Sec.~\ref{sec:sim2real}) treats 3D generation as the link between simulation and deployment, covering data augmentation, digital twin construction, and task or demonstration generation for downstream robot learning. Around these three roles, we further summarize the technical preliminaries, datasets, evaluation protocols, and open challenges that shape this field.

The three roles are distinguished by their primary objective rather than by the underlying technique. \emph{Data Generator} covers methods whose goal is to \emph{create new} simulation-ready assets---learning generative priors over geometry, articulation, or physics to synthesize novel objects that do not yet exist in any asset library. \emph{Simulation Environments} covers methods that \emph{compose} objects into interactive scenes, whether by procedural rules, learned layout priors, or agentic planning. \emph{Sim2Real Bridge} covers methods that \emph{reconstruct or augment} existing real-world content for deployment---building digital twins from observations, augmenting recorded demonstrations, or synthesizing training trajectories. When a technique spans multiple roles (e.g., articulated object methods that both generate new assets and reconstruct real-world instances), we classify it by the role that best reflects its primary contribution and discuss its connections to related sections explicitly.

\noindent\textbf{Contributions.}
This survey makes the following contributions:

\begin{itemize}[nosep, leftmargin=1.2em]
  \item We clarify what distinguishes embodied-oriented 3D generation from traditional 3D generation, establishing simulation readiness (encompassing geometric validity, physical parameterization, kinematic executability, and simulator compatibility) as the central evaluation criterion. This reframes the literature around deployment requirements rather than visual metrics alone.

  \item We provide the first taxonomy organized around the three core roles of 3D generation in embodied AI, \emph{Data Generator}, \emph{Simulation Environments}, and \emph{Sim2Real Bridge}, thereby connecting research threads from computer vision, computer graphics, robotics, and simulation systems that are typically discussed in isolation.

  \item We identify and systematize the key technical gaps that still limit practical deployment: the lack of standardized physical annotations at scale, the tension between generation quality and simulator compatibility, the difficulty of evaluating embodied usefulness, and the remaining domain gap in sim-to-real transfer. We translate these into concrete research directions for the community.
\end{itemize}

\begin{figure*}[t]
    \centering
    \includegraphics[width=0.85\linewidth]{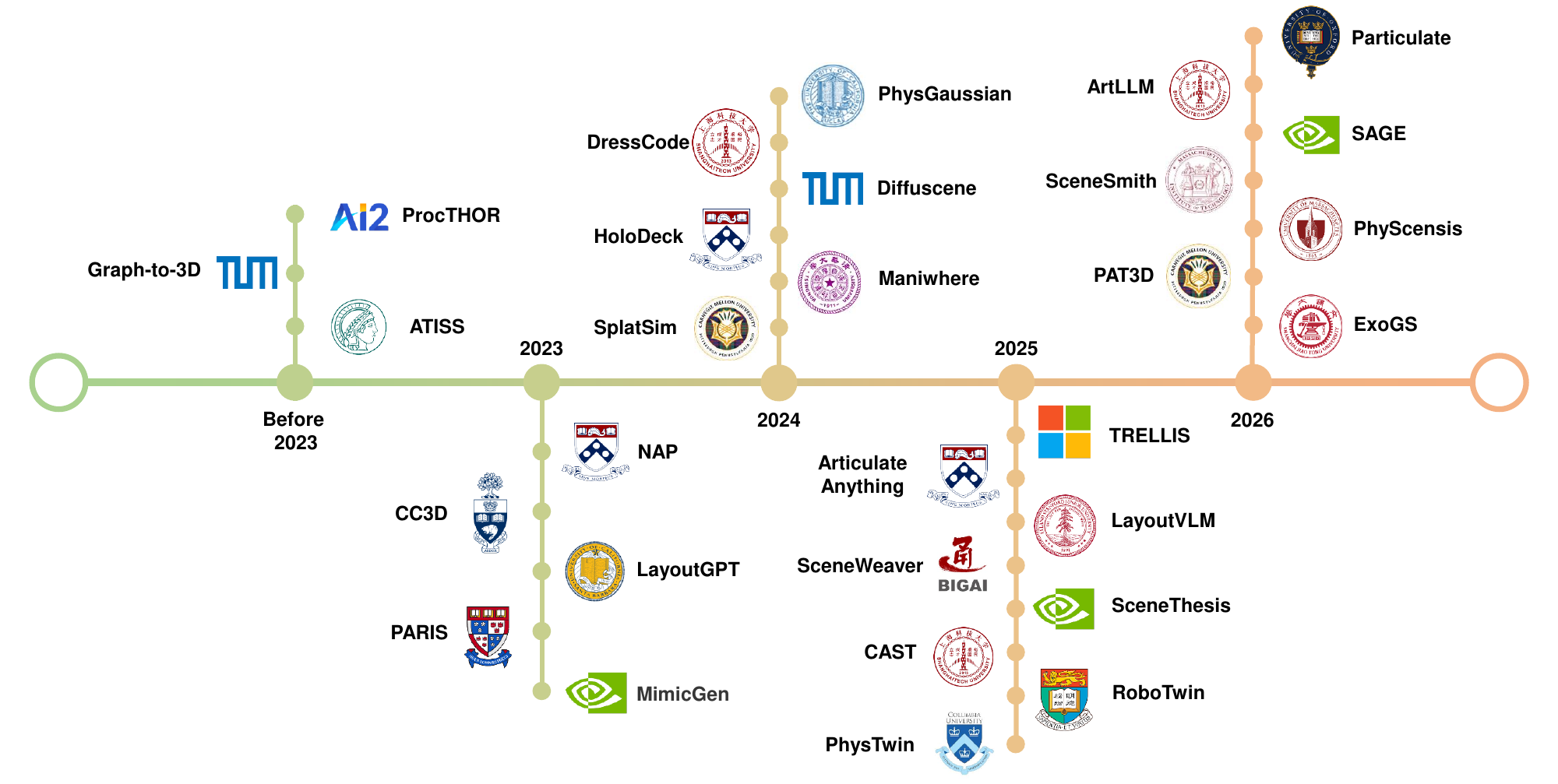}
    \caption{\textbf{Timeline of representative methods for 3D generation in embodied AI.} The figure highlights key milestones across object generation, scene generation, and Sim2Real-related pipelines from before 2023 to 2026.}
\end{figure*}

\noindent\textbf{Scope of This Survey.}
This survey focuses on \emph{simulation-centric} 3D asset generation---methods whose outputs are intended for deployment in physical simulators or embodied AI environments. We include object-level and scene-level generation addressing geometry, articulation, or physical properties, as well as Real-to-Sim and Sim-to-Real transfer pipelines. We exclude purely 2D methods, novel view synthesis without simulation-ready geometry, robot policy methods unless 3D generation is their central contribution, and outdoor/autonomous driving scene generation which operates on fundamentally different asset granularities and simulator ecosystems.

\noindent\textbf{Relation to Other Surveys.}
Existing surveys cover adjacent areas: 3D generation surveys~\cite{li2024advances, tang2025recent, wang2025diffusion} focus on generative backbones and visual metrics without addressing simulator compatibility or embodied evaluation; scene generation surveys~\cite{wen20253d} omit object-level simulation-ready generation and sim-to-real transfer; embodied AI surveys~\cite{liu2025aligning} treat 3D assets as given infrastructure; and generative AI for manipulation~\cite{zhang2025generative} covers policy and planning but not the upstream asset generation pipeline. To our knowledge, no prior survey jointly covers object generation, scene synthesis, and sim-to-real transfer through the lens of embodied AI.

\noindent\textbf{Organization.}
The remainder of this survey is organized as follows. 
Sec.~\ref{sec:pre} introduces the technical preliminaries, covering 3D representations, generative model foundations, and embodied AI fundamentals.
Sec.~\ref{sec:data_generator} presents \emph{Data Generator}, where 3D generation serves as the source of simulation-ready objects and assets, including articulated, physically grounded, and deformable object generation as well as end-to-end simulation-ready pipelines.
Sec.~\ref{sec:sim_env} presents \emph{Simulation Environments}, where 3D generation serves as the builder of interactive worlds, from structure-driven and controllable scene synthesis to agentic and task-conditioned environment creation. 
Sec.~\ref{sec:sim2real} presents \emph{Sim2Real Bridge}, where 3D generation serves as the connector between simulation and deployment through data augmentation, digital twin construction, and task and demonstration generation for robot learning.
Sec.~\ref{sec:dataset} summarizes datasets and evaluation protocols relevant to 3D asset generation for embodied AI.
Sec.~\ref{sec:challenges} identifies the major open challenges and future directions.
Sec.~\ref{sec:conclusion} concludes the survey.

%% file: sec/prelim.tex
\section{Preliminaries}
\label{sec:pre}

\subsection{3D Representations}
\label{subsec:3d_rep}

\begin{figure*}
    \centering
    \includegraphics[width=0.95\linewidth]{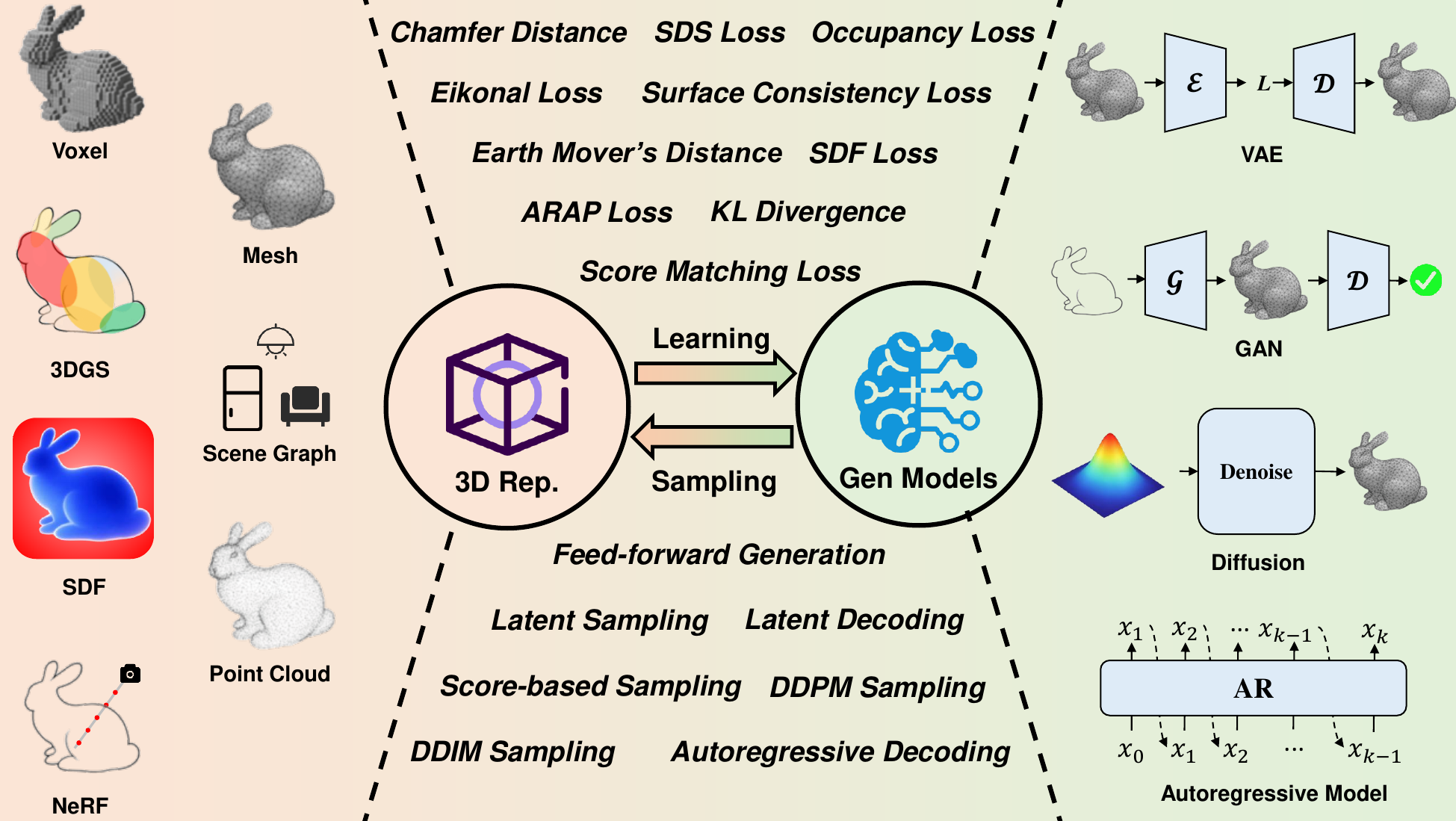}
    \caption{\textbf{3D representations and generative modeling paradigms.} \textbf{\textit{Left:}} common 3D representations. \textbf{\textit{Center:}} the bidirectional interplay between representations and generative models through learning and sampling. \textbf{\textit{Right:}} major generative model families used for 3D content synthesis.}
    \label{fig:3d-rep-methods}
\end{figure*}

The choice of 3D representation determines what physical information an asset can carry and how it interfaces with downstream simulation and generation pipelines. We briefly review the representations that recur throughout this survey, emphasizing their roles in embodied AI rather than their general formulations (see~\cite{li2024advances, tang2025recent}).

\noindent\textbf{Explicit Geometric Representations.}
A \textbf{voxel grid} $\mathbf{V}\!\in\!\mathbb{R}^{H\times W\times D}$ discretizes space into regular cells storing occupancy or distance values, facilitating 3D convolutions and collision checking in simulation, though its cubic memory cost limits resolution.
A \textbf{point cloud} $\mathbf{P}=\{p_i\!\in\!\mathbb{R}^3\}_{i=1}^N$ is the natural output of depth sensors and LiDAR, serving as the primary input for robotic grasping and manipulation pipelines~\cite{sfm}; its unordered, lightweight structure makes it a common bridge between perception and generation.
A \textbf{polygonal mesh} $\mathbf{M}=(M_V,M_E,M_F)$ explicitly encodes surface topology and connectivity, making it the standard representation for physics engines and the target output format for most simulation-ready generation pipelines.

\noindent\textbf{Structured and Implicit Representations.}
A \textbf{scene graph} $\mathbf{G}=(O,R)$ encodes objects and their spatial or functional relationships (e.g., support, adjacency), providing the structured interface used by scene generation methods for layout planning and task-conditioned composition~\cite{graph_to_3d}.
A \textbf{signed distance function (SDF)} defines geometry as a continuous field $f\!:\!\mathbb{R}^3\!\to\!\mathbb{R}$ whose zero-level set is the surface~\cite{deepsdf}; neural SDFs guarantee watertight outputs, which is critical for collision geometry in simulation.

\noindent\textbf{Neural Rendering Representations.}
\textbf{Neural Radiance Fields (NeRF)}~\cite{nerf} map position and viewing direction to density and color via a neural function $F(x,d)=(\sigma,c)$, synthesizing images through differentiable volume rendering. In embodied AI, NeRFs are widely used for viewpoint augmentation and scene reconstruction in sim-to-real pipelines.
\textbf{3D Gaussian Splatting (3DGS)}~\cite{3dgs} represents scenes as sets of anisotropic Gaussian primitives $\{(\mu_i,\Sigma_i,c_i,\alpha_i)\}$ rendered via differentiable rasterization, achieving real-time speeds. Beyond visual reconstruction, 3DGS has become a key enabler for digital twin construction and data augmentation, as each Gaussian can additionally carry physical attributes for unified rendering and physics simulation~\cite{xie2024physgaussian}.

\subsection{Generative Models}
\label{subsec:generative_models}

Generative models provide the methodological backbone for 3D content synthesis. Here we summarize the four paradigms most relevant to embodied 3D generation; each is characterized by how it models the data distribution and what conditioning and controllability it affords.

\noindent\textbf{VAE.}
Variational Autoencoders learn a continuous latent space via encoder--decoder pairs, enabling smooth interpolation and conditional generation. In embodied 3D generation, VAEs serve primarily as the \emph{latent encoding backbone} in end-to-end simulation-ready pipelines~\cite{xiang2025structured, cao2025physx3d}, compressing shapes, point clouds~\cite{pointcloud}, or meshes~\cite{meshdeform, meshvae} into compact codes over which downstream diffusion or autoregressive models operate.

\noindent\textbf{GAN.}
Generative Adversarial Networks have been applied to 3D shapes~\cite{3dgan, sdfstylegan} and 3D-aware image synthesis~\cite{graf, pigan}, but have been largely superseded in embodied 3D generation by diffusion-based approaches with superior training stability and conditioning flexibility.

\noindent\textbf{Diffusion and Flow-Based Models.}
Diffusion models---the currently dominant paradigm---learn to reverse a gradual noising process, operating on native 3D data (point clouds~\cite{pvd, lion}, latent fields~\cite{shape2vecset}) or leveraging 2D priors via Score Distillation Sampling~\cite{dreamfusion}. The closely related \textit{flow matching} formulation~\cite{flowmatching} learns straight-trajectory transport from noise to data, yielding faster sampling and has been adopted in recent simulation-ready pipelines~\cite{xiang2025structured}. 

\noindent\textbf{Autoregressive Model.}
Autoregressive models factorize the joint distribution into sequential conditional predictions over tokenized shapes, layouts, or kinematic structures. This paradigm is particularly suited to \emph{structured} generation tasks where compositional reasoning matters---scene arrangement~\cite{atiss, g3pt}, skeleton hierarchy prediction~\cite{song2025magicarticulate}, and URDF parameter generation~\cite{wu2026urdf}---and integrates naturally with Transformer architectures for text- or instruction-conditioned synthesis.

\subsection{Simulation Platforms and Assets}
\label{subsec:platform}

\begin{figure*}
    \centering
    \includegraphics[width=0.95\linewidth]{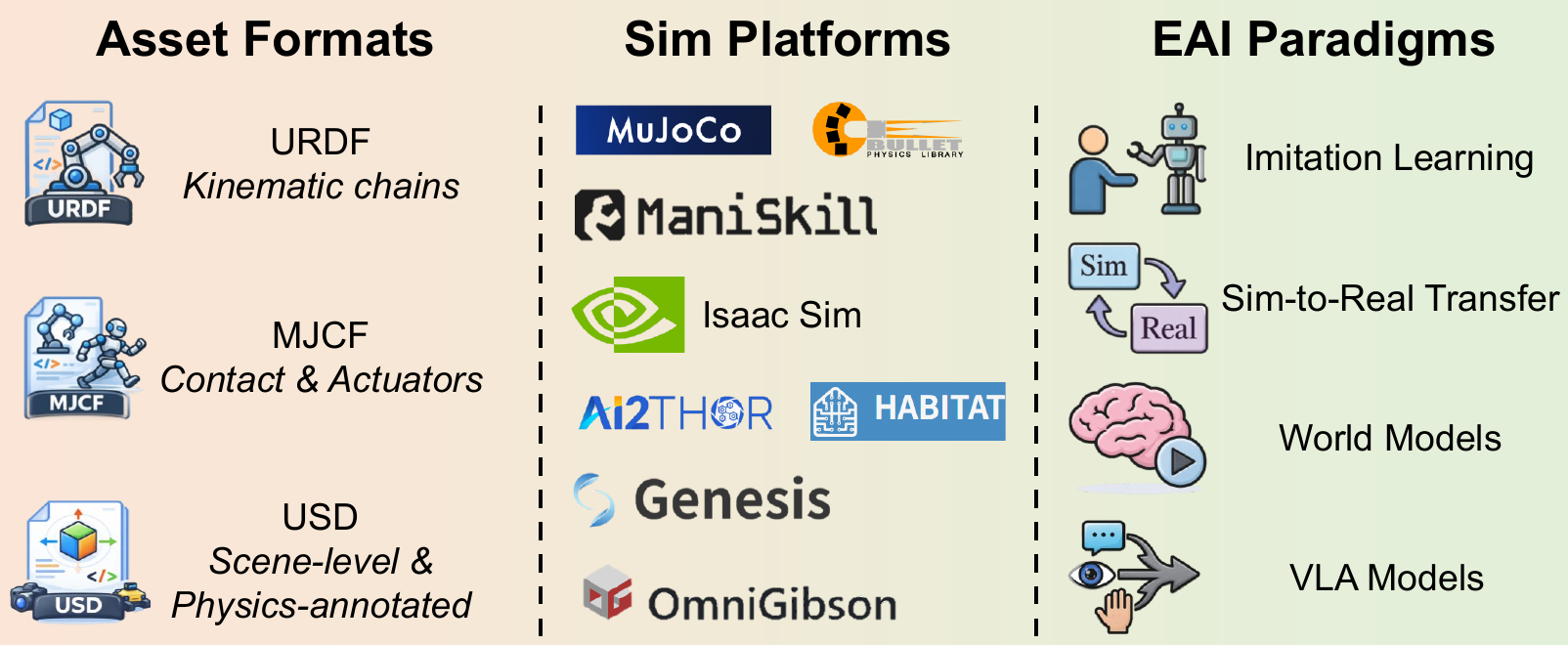}
    \caption{\textbf{Infrastructure connecting generated 3D content to embodied AI.} \textbf{\textit{Left:}} simulation-ready asset formats. \textbf{\textit{Center:}} representative simulation platforms. \textbf{\textit{Right:}} downstream embodied AI paradigms that consume simulated assets and environments.}
    \label{fig:eai_fund}
\end{figure*}

For generated 3D content to be useful in embodied AI (EAI), it must ultimately be loaded into a physics simulator that supports rendering, contact dynamics, and agent interaction. This requires both a \emph{simulation-compatible format} that encodes geometry, kinematics, and physical properties in a machine-readable specification, and a \emph{simulation platform} that consumes such formats and provides the execution environment for robot learning.

\noindent\textbf{Simulation Readiness Definition.}
We use \emph{simulation readiness} to describe whether generated 3D assets or scenes can be consumed by physics-based simulators with minimal manual post-processing. Beyond visual plausibility, simulation-ready content should satisfy four requirements:
\begin{enumerate}[nosep, leftmargin=1.5em, label=(\roman*)]
  \item \emph{Geometric validity}: consistent scale, clean topology, part-level structure when applicable, and usable collision geometry for contact computation.
  \item \emph{Physical parameterization}: simulation-relevant attributes such as mass, inertia, density, friction, restitution, center of mass, and material properties.
  \item \emph{Kinematic executability}: articulated structures, joint types, joint axes, joint limits, and motion constraints when objects contain movable parts.
  \item \emph{Simulator compatibility}: serialization into one or more executable asset formats consumable by target physics engines.
\end{enumerate}
\noindent This definition generalizes existing SimReady-style asset standards and recent simulation-ready generation pipelines by emphasizing runtime-consumable geometry, physics, kinematics, and metadata rather than visual fidelity alone.

\noindent\textbf{Simulation-Ready Formats.}
The choice of asset format determines what physical information can be expressed and which simulators can consume the output.
The \textbf{Unified Robot Description Format (URDF)} remains the standard for kinematic structures within the ROS ecosystem, encoding link--joint trees with geometry, inertia, and joint limits. However, URDF is limited to tree-structured kinematic chains and lacks native support for complex contact models or closed loops.
The \textbf{MuJoCo XML Format (MJCF)} extends beyond simple kinematic trees by orienting its structure around bodies, joints, and rich physical interactions---including tendons, actuator models, and contact parameters---making it the preferred format for high-fidelity manipulation and locomotion simulation.
At the scene level, the \textbf{Universal Scene Description (USD)} provides a comprehensive scene graph that natively supports material bindings, lighting, animations, and sensor attributes. The associated \textbf{SimReady} standard further annotates USD assets with semantic labels, collision geometry, and rigid-body physical properties (mass, center of mass, friction, restitution) to enable immediate deployment in simulators such as Isaac Sim.
A method's practical value for embodied AI depends on whether its output can be serialized into one of these formats without extensive manual post-processing.

\noindent\textbf{Simulation Platforms.}
Modern robotic simulators have evolved into GPU-accelerated ecosystems. Table~\ref{tab:sim_platforms} summarizes the major platforms, which can be broadly categorized by focus: \textit{high-throughput physics} (MuJoCo~\cite{todorov2012mujoco}, ManiSkill3~\cite{tao2024maniskill3}), \textit{photorealistic sensor simulation} (Isaac Sim~\cite{makoviychuk2021isaac}), \textit{semantic and interactive environments} (Habitat~\cite{savva2019habitat}, AI2-THOR~\cite{kolve2017ai2thor}, OmniGibson~\cite{li2023behavior}), \textit{lightweight prototyping} (PyBullet~\cite{pybullet}), and \textit{differentiable multi-physics} (Genesis~\cite{genesis}).

\begin{table}[t]
  \centering
  \small
  \caption{Major robotic simulation platforms for embodied AI.}
  \label{tab:sim_platforms}
  \renewcommand{\arraystretch}{1.15}
  \rowcolors{2}{gray!20}{white}
  \resizebox{\columnwidth}{!}{%
  \begin{tabular}{l l l l l l}
    \toprule
    \textbf{Platform} & \textbf{Engine} & \textbf{Format} & \textbf{Rendering} & \textbf{Diff.\ Phys.} & \textbf{Primary Strength} \\
    \midrule
    MuJoCo      & MuJoCo/MJX  & MJCF/URDF & Native        & \cmark~(MJX/JAX)   & Fast contact; RL \\
    Isaac Sim   & PhysX 5     & USD/URDF  & RTX ray-tracing & \xmark            & Photorealism; sensor sim \\
    Habitat 3.0 & Bullet      & URDF/GLB  & Rasterization & \xmark            & Navigation; human-robot \\
    AI2-THOR    & Unity       & Custom    & Unity URP     & \xmark            & Semantic states; VLN \\
    OmniGibson  & PhysX 5     & USD       & RTX ray-tracing & \xmark            & Affordance; soft-body \\
    PyBullet    & Bullet      & URDF      & OpenGL        & \pmark~(TDS)      & Lightweight; prototyping \\
    ManiSkill3  & SAPIEN      & URDF/USD  & Vulkan/RT     & \xmark            & Heterogeneous GPU parallel \\
    Genesis     & Diff.\ MPM/FEM & MJCF/URDF & Rasterization & \cmark~(Analytical) & Multi-physics; gradients \\
    \bottomrule
  \end{tabular}}
\end{table}

\subsection{Embodied AI Fundamentals}
This section introduces the downstream robot learning paradigms that determine \emph{why} scalable 3D generation is needed for embodied AI.

\noindent\textbf{Imitation Learning and Demonstration Data.}
Imitation learning trains robot policies by mimicking expert behavior, most commonly through \textit{behavior cloning} (BC), which frames policy learning as supervised regression from observations to actions~\cite{chi2025diffusion, zhao2023learning}.
A robot demonstration typically consists of a sequence of observation--action pairs $\{(o_t, a_t)\}_{t=1}^{T}$, where observations may include RGB images, depth maps, or point clouds, and actions encode end-effector poses, joint commands, or gripper states.
The practical bottleneck is data collection: high-quality demonstrations require human teleoperation or kinesthetic teaching, both of which are slow, expensive, and difficult to scale.

\noindent\textbf{Sim-to-Real Transfer.}
Training robot policies in simulation offers scalability and safety, but introduces a \textit{domain gap}---the distributional mismatch between simulated and real-world observations, physics, and dynamics---that can cause policies to fail upon deployment.
Three complementary strategies address this gap.
\textit{Domain randomization}~\cite{tobin2017domain} varies visual and physical parameters during training so that the policy learns invariances that transfer to the real world; subsequent work has refined this idea through learned randomization distributions~\cite{mehta2020adr}, Bayesian parameter inference~\cite{ramos2019bayessim}, closed-loop range adaptation~\cite{chebotar2019simopt}, and visual canonicalization networks~\cite{james2019rcan}.
\textit{Real-to-Sim (R2S) reconstruction} builds high-fidelity digital replicas of real environments to reduce the domain gap at its source.
\textit{Real-to-Sim-to-Real (R2S2R) pipelines} combine both directions: reconstructing the real world into simulation, generating or augmenting data within it, and transferring the resulting policies back to reality~\cite{robogsim, wang2026exogs}.

\noindent\textbf{World Models.}
\textit{World models}~\cite{ha2018world} learn to simulate environment dynamics from experience, enabling large-scale imaginary rollouts for long-horizon planning without real-world cost. Recent methods leverage action-conditioned video diffusion~\cite{dreamgen, RoboScape} or 3D scene generation to serve as neural simulators.

\noindent\textbf{Vision-Language-Action Models.}
Vision-Language-Action (VLA) models~\cite{zitkovich2023rt, kim2024openvla, black2024pi0, intelligence2025pi_} take visual observations and language instructions as input and directly output robot actions, promising open-vocabulary generalization across tasks and objects. Their data appetite is correspondingly large, creating a direct demand for scalable 3D generation at every level---object assets, scene configurations, and synthetic demonstrations.

%% file: sec/data_generator.tex
\section{Data Generator}
\label{sec:data_generator}

\begin{figure*}[t]
    \centering
    \includegraphics[width=0.95\linewidth]{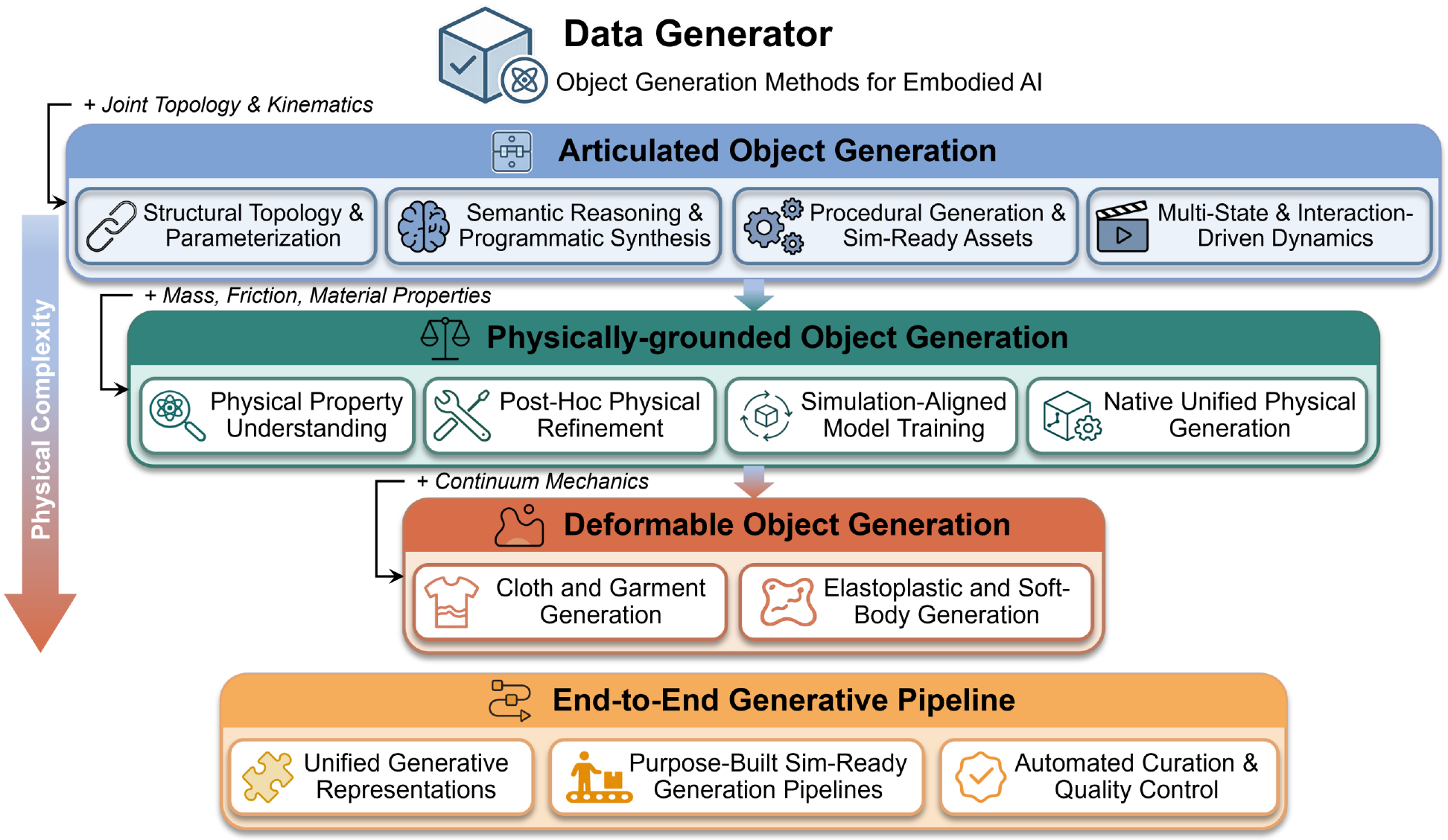}
    \caption{\textbf{Overview of generative 3D object methods for embodied AI}, organized by increasing physical complexity: \textit{Articulated} (joint topology and kinematics), \textit{Physically Grounded} (mass, friction, material properties), \textit{Deformable} (continuum mechanics for cloth and soft bodies), and \textit{End-to-End Pipelines} (automated simulation-ready workflows).}
    \label{fig:data-generator}
\end{figure*}

This section examines how generative models produce \textit{simulation-ready} objects deployable in physics engines without post-processing. The central challenge is not visual plausibility but physical deployability: generated assets must carry kinematic constraints, material attributes, and simulator-compatible formats alongside geometry. We focus on methods that \emph{create new} assets by learning generative priors; methods that \emph{reconstruct} specific real-world instances are covered in Sec.~\ref{sec:digital_twin}.
This section is organized by increasing physical complexity: articulated objects (Sec.~\ref{sec:articulated}), physically grounded objects (Sec.~\ref{sec:physical}), deformable objects (Sec.~\ref{sec:deformable}), and end-to-end simulation-ready pipelines (Sec.~\ref{sec:pipeline}).

\subsection{Generative Objects}

\begin{table*}[!t]
\caption{
  Summary of generative 3D object methods for embodied AI.}
\label{tab:generative_objects}

\vspace{-5pt}
\begin{minipage}{\textwidth}
    \scriptsize
    \begin{itemize}[leftmargin=1.2em, nosep, itemsep=2pt, label=\small$\bullet$]
        \item \textbf{Input}:
          \textcolor{gray}{\faFont}~Text,
          \textcolor{orange}{\faImage}~Image,
          \textcolor{blue}{\faBraille}~Point Cloud,
          \textcolor{cyan}{\faCube}~Mesh,
          \textcolor{magenta}{\faVideo}~Video/Multi-view,
          \textcolor{brown}{\faHandPointer}~Interaction
        \item \textbf{Output}:
          \textcolor{cyan}{\faCube}~Mesh,
          \textcolor{teal}{\faAtom}~3DGS/RF,
          \textcolor{violet}{\faWaveSquare}~Field, 
          \textcolor{purple}{\faProjectDiagram}~Kinematic/URDF,
          \textcolor{green!60!black}{\faFileCode}~Code,
          \textcolor{olive}{\faBalanceScale}~Physics Params,
          \textcolor{red}{\faWind}~Trajectory,
        \item \textbf{Architectures (Arch):}
          \textit{\textbf{Diff}}: Diffusion Model,
          \textit{\textbf{AR}}: Autoregressive,
          \textit{\textbf{Transf}}: Transformer,
          \textit{\textbf{GNN}}: Graph Neural Network,
          \textit{\textbf{Impl}}: Implicit Repr.,
          \textit{\textbf{3DGS}}: 3D Gaussian Splatting,
          \textit{\textbf{LLM/VLM}}: Large/Vision-Language Model,
          \textit{\textbf{FF}}: Feed-forward Network,
          \textit{\textbf{VAE}}: Variational Autoencoder,
          \textit{\textbf{Sp.DiT}}: Sparse Diffusion Transformer,
          \textit{\textbf{MPM}}: Material Point Method,
          \textit{\textbf{FEM}}: Finite Element Method,
          \textit{\textbf{MCTS}}: Monte Carlo Tree Search,
          \textit{\textbf{Proc}}: Procedural Generation,
          \textit{\textbf{Voxel}}: Voxel-based Repr.
        \item \textbf{Sim-Ready}: 
          \textit{\textbf{Partial}}: intermediate representations requiring additional processing before full simulator deployment, \textit{\textbf{Kin.}}: Kinematic, 
          \textit{\textbf{Traj.}}: Trajectory, 
          \textit{\textbf{Dyn.}}: Dynamic, 
          \textit{\textbf{Sew.~Pat.}}: Sewing Pattern, 
          \textit{\textbf{RF}}: Radiance Field.
        \item \textbf{Categories (Cat)}:
          \textcolor{blue}{\ding{182}}~Articulated (Sec.~\ref{sec:articulated}),
          \textcolor{red}{\ding{183}}~Physical (Sec.~\ref{sec:physical}),
          \textcolor{teal}{\ding{184}}~Deformable (Sec.~\ref{sec:deformable}),
          \textcolor{orange}{\ding{185}}~Pipeline (Sec.~\ref{sec:pipeline}).
    \end{itemize}
\end{minipage}
\rowcolors{2}{gray!20}{white}
\resizebox{\textwidth}{!}{%
\scriptsize
\begin{tabular}{@{}c l l c c c c c c c@{}}
\toprule
\textbf{\#} & \textbf{Method} & \textbf{Venue}
  & \textbf{Input} & \textbf{Output}
  & \textbf{Arch} & \textbf{Supervision}
  & \textbf{Sim-Ready} & \textbf{Cat} & \textbf{URL} \\
\midrule

1 & NAP~\cite{lei2023nap} & NeurIPS '23
  & \textcolor{gray}{\faFont}
  & \textcolor{purple}{\faProjectDiagram}
  & Diff+GNN & 3D GT & Kin. Graph (Partial)
  & \textcolor{blue}{\ding{182}}
  & \href{https://arxiv.org/abs/2305.16315}{\textcolor{blue}{\faExternalLink*}} \\

2 & CAGE~\cite{liu2024cage} & CVPR '24
  & \textcolor{gray}{\faFont}
  & \textcolor{cyan}{\faCube}~\textcolor{purple}{\faProjectDiagram}
  & Diff+GNN & 3D GT & Mesh + Kin.
  & \textcolor{blue}{\ding{182}}
  & \href{https://3dlg-hcvc.github.io/cage/}{\textcolor{blue}{\faExternalLink*}} \\

3 & URDFormer~\cite{chen2024urdformer} & RSS '24
  & \textcolor{orange}{\faImage}
  & \textcolor{purple}{\faProjectDiagram}
  & Transf & 3D GT & URDF
  & \textcolor{blue}{\ding{182}}
  & \href{https://urdformer.github.io/}{\textcolor{blue}{\faExternalLink*}} \\

4 & SINGAPO~\cite{liu2024singapo} & ICLR '25
  & \textcolor{orange}{\faImage}
  & \textcolor{cyan}{\faCube}~\textcolor{purple}{\faProjectDiagram}
  & Diffusion & 3D GT & URDF
  & \textcolor{blue}{\ding{182}}
  & \href{https://3dlg-hcvc.github.io/singapo}{\textcolor{blue}{\faExternalLink*}} \\

5 & Real2Code~\cite{mandi2024real2code} & ICLR '25
  & \textcolor{gray}{\faFont}~\textcolor{orange}{\faImage}
  & \textcolor{green!60!black}{\faFileCode}
  & LLM/VLM & Text/Image & Code (Python)
  & \textcolor{blue}{\ding{182}}
  & \href{https://real2code.github.io/}{\textcolor{blue}{\faExternalLink*}} \\

6 & Articulate-Anything~\cite{le2024articulate} & ICLR '25
  & \textcolor{gray}{\faFont}~\textcolor{orange}{\faImage}
  & \textcolor{green!60!black}{\faFileCode}
  & LLM/VLM & Text/Image & Code (URDF)
  & \textcolor{blue}{\ding{182}}
  & \href{https://articulate-anything.github.io/}{\textcolor{blue}{\faExternalLink*}} \\

7 & MagicArticulate~\cite{song2025magicarticulate} & CVPR '25
  & \textcolor{cyan}{\faCube}
  & \textcolor{cyan}{\faCube}~\textcolor{purple}{\faProjectDiagram}
  & AR & 3D GT & Rigged Asset
  & \textcolor{blue}{\ding{182}}
  & \href{https://chaoyuesong.github.io/MagicArticulate/}{\textcolor{blue}{\faExternalLink*}} \\

8 & MeshArt~\cite{gao2025meshart} & CVPR '25
  & \textcolor{gray}{\faFont}
  & \textcolor{cyan}{\faCube}~\textcolor{purple}{\faProjectDiagram}
  & AR & 3D GT & Mesh + Kin.
  & \textcolor{blue}{\ding{182}}
  & \href{https://daoyig.github.io/Mesh_Art/}{\textcolor{blue}{\faExternalLink*}} \\

9 & PartRM~\cite{gao2025partrm} & CVPR '25
  & \textcolor{orange}{\faImage}~\textcolor{brown}{\faHandPointer}
  & \textcolor{red}{\faWind}
  & Diff & Interaction & Traj. (Partial)
  & \textcolor{blue}{\ding{182}}
  & \href{https://partrm.c7w.tech/}{\textcolor{blue}{\faExternalLink*}} \\

10 & ArtFormer~\cite{su2025artformer} & CVPR '25
  & \textcolor{gray}{\faFont}~\textcolor{orange}{\faImage}
  & \textcolor{cyan}{\faCube}~\textcolor{purple}{\faProjectDiagram}
  & Diffusion & 3D GT & URDF
  & \textcolor{blue}{\ding{182}}
  & \href{https://arxiv.org/abs/2412.07237}{\textcolor{blue}{\faExternalLink*}} \\

11 & ArtiWorld~\cite{yang2025artiworld} & arXiv '25
  & \textcolor{gray}{\faFont}~\textcolor{orange}{\faImage}
  & \textcolor{purple}{\faProjectDiagram}
  & LLM+Diffusion & Weak & URDF
  & \textcolor{blue}{\ding{182}}
  & \href{https://arxiv.org/abs/2511.12977}{\textcolor{blue}{\faExternalLink*}} \\

12 & Articulate AnyMesh~\cite{qiu2025articulate} & CoRL '25
  & \textcolor{gray}{\faFont}~\textcolor{cyan}{\faCube}
  & \textcolor{purple}{\faProjectDiagram}
  & LLM & Weak & URDF
  & \textcolor{blue}{\ding{182}}
  & \href{https://articulate-anymesh.github.io/}{\textcolor{blue}{\faExternalLink*}} \\

13 & ATOP~\cite{vora2025articulate} & TOG '26
  & \textcolor{gray}{\faFont}~\textcolor{orange}{\faImage}
  & \textcolor{purple}{\faProjectDiagram}
  & VLM & Weak & URDF
  & \textcolor{blue}{\ding{182}}
  & \href{https://arxiv.org/abs/2502.07278}{\textcolor{blue}{\faExternalLink*}} \\

14 & URDF-Anything~\cite{li2025urdf} & NeurIPS '25
  & \textcolor{orange}{\faImage}
  & \textcolor{purple}{\faProjectDiagram}
  & LLM/VLM & Text/Image & URDF
  & \textcolor{blue}{\ding{182}}
  & \href{https://lzvsdy.github.io/URDF-Anything/}{\textcolor{blue}{\faExternalLink*}} \\

15 & ArtiLatent~\cite{chen2025artilatent} & SIG. Asia '25
  & \textcolor{gray}{\faFont}~\textcolor{orange}{\faImage}
  & \textcolor{cyan}{\faCube}~\textcolor{purple}{\faProjectDiagram}
  & Diff+VAE & 3D GT & Mesh + Kin.
  & \textcolor{blue}{\ding{182}}
  & \href{https://chenhonghua.github.io/MyProjects/ArtiLatent/}{\textcolor{blue}{\faExternalLink*}} \\

16 & ArtGen~\cite{wang2025artgen} & arXiv '25
  & \textcolor{gray}{\faFont}~\textcolor{orange}{\faImage}
  & \textcolor{cyan}{\faCube}~\textcolor{purple}{\faProjectDiagram}
  & Diff (MoE) & 3D GT & Mesh + Kin.
  & \textcolor{blue}{\ding{182}}
  & \href{https://arxiv.org/abs/2512.12395}{\textcolor{blue}{\faExternalLink*}} \\

17 & DreamArt~\cite{lu2025dreamart} & arXiv '25
  & \textcolor{orange}{\faImage}
  & \textcolor{cyan}{\faCube}
  & Diff & Text/Image & Mesh
  & \textcolor{blue}{\ding{182}}
  & \href{https://arxiv.org/abs/2507.05763}{\textcolor{blue}{\faExternalLink*}} \\

18 & GAOT~\cite{sun2025gaot} & ACM MM Asia '25
  & \textcolor{blue}{\faBraille}~\textcolor{gray}{\faFont}
  & \textcolor{purple}{\faProjectDiagram}
  & GNN & 3D GT & Kin. Graph
  & \textcolor{blue}{\ding{182}}
  & \href{https://arxiv.org/abs/2512.03566}{\textcolor{blue}{\faExternalLink*}} \\

19 & Kinematify~\cite{wang2025kinematify} & ICRA '26
  & \textcolor{cyan}{\faCube}
  & \textcolor{purple}{\faProjectDiagram}
  & MCTS & Search & Kin. Tree
  & \textcolor{blue}{\ding{182}}
  & \href{https://sites.google.com/deemos.com/kinematify}{\textcolor{blue}{\faExternalLink*}} \\

20 & Particulate~\cite{li2025particulate} & CVPR '26
  & \textcolor{orange}{\faImage}
  & \textcolor{cyan}{\faCube}~\textcolor{purple}{\faProjectDiagram}
  & Transf & 3D GT & Mesh + Kin. (Partial)
  & \textcolor{blue}{\ding{182}}
  & \href{https://ruiningli.com/particulate}{\textcolor{blue}{\faExternalLink*}} \\

21 & UniArt~\cite{jin2025uniart} & arXiv '25
  & \textcolor{blue}{\faBraille}
  & \textcolor{purple}{\faProjectDiagram}
  & Voxel+GNN & 3D GT & URDF
  & \textcolor{blue}{\ding{182}}
  & \href{https://arxiv.org/abs/2511.21887}{\textcolor{blue}{\faExternalLink*}} \\

22 & Infinite Mobility~\cite{lian2025infinite} & arXiv '25
  & \textcolor{gray}{\faFont}
  & \textcolor{cyan}{\faCube}~\textcolor{purple}{\faProjectDiagram}
  & Proc & Rules & URDF/USD
  & \textcolor{blue}{\ding{182}}
  & \href{https://infinite-mobility.github.io/}{\textcolor{blue}{\faExternalLink*}} \\

23 & URDF-Anything+~\cite{wu2026urdf} & NeurIPS '26
  & \textcolor{gray}{\faFont}~\textcolor{orange}{\faImage}
  & \textcolor{purple}{\faProjectDiagram}
  & AR+LLM/VLM & Text/Image & URDF
  & \textcolor{blue}{\ding{182}}
  & \href{https://urdf-anything-plus.github.io/}{\textcolor{blue}{\faExternalLink*}} \\

24 & ArtLLM~\cite{wang2026artllm} & CVPR '26
  & \textcolor{gray}{\faFont}~\textcolor{orange}{\faImage}
  & \textcolor{purple}{\faProjectDiagram}
  & LLM & Weak & URDF
  & \textcolor{blue}{\ding{182}}
  & \href{https://arxiv.org/abs/2603.01142}{\textcolor{blue}{\faExternalLink*}} \\

25 & SPARK~\cite{he2025spark} & CVPR '26
  & \textcolor{gray}{\faFont}~\textcolor{orange}{\faImage}
  & \textcolor{purple}{\faProjectDiagram}
  & VLM+Sim & Weak & URDF
  & \textcolor{blue}{\ding{182}}
  & \href{https://heyumeng.com/SPARK-web/}{\textcolor{blue}{\faExternalLink*}} \\

26 & PAct~\cite{liu2026pact} & arXiv '26
  & \textcolor{orange}{\faImage}
  & \textcolor{cyan}{\faCube}~\textcolor{purple}{\faProjectDiagram}
  & FF & Supervised & URDF
  & \textcolor{blue}{\ding{182}}
  & \href{https://pact-project.github.io/}{\textcolor{blue}{\faExternalLink*}} \\

\midrule
27 & NeRF2Physics~\cite{zhai2024physical} & CVPR '24
  & \textcolor{orange}{\faImage}
  & \textcolor{olive}{\faBalanceScale}
  & Impl+LLM/VLM & Zero-shot & Params Only
  & \textcolor{red}{\ding{183}}
  & \href{https://ajzhai.github.io/NeRF2Physics/}{\textcolor{blue}{\faExternalLink*}} \\

28 & Atlas3D~\cite{chen2024atlas3d} & NeurIPS '24
  & \textcolor{gray}{\faFont}~\textcolor{orange}{\faImage}
  & \textcolor{cyan}{\faCube}
  & Diff & Text/Image & Mesh
  & \textcolor{red}{\ding{183}}
  & \href{https://yunuoch.github.io/Atlas3D/}{\textcolor{blue}{\faExternalLink*}} \\

29 & PhysComp~\cite{guo2024physically} & NeurIPS '24
  & \textcolor{orange}{\faImage}
  & \textcolor{cyan}{\faCube}
  & FEM & 3D GT & Mesh
  & \textcolor{red}{\ding{183}}
  & \href{https://arxiv.org/abs/2405.20510}{\textcolor{blue}{\faExternalLink*}} \\

30 & PhyCAGE~\cite{yan2024phycage} & NeurIPS '24
  & \textcolor{orange}{\faImage}
  & \textcolor{cyan}{\faCube}~\textcolor{teal}{\faAtom} 
  & 3DGS+Diff & 3D GT & Mesh
  & \textcolor{red}{\ding{183}}
  & \href{https://wolfball.github.io/phycage/}{\textcolor{blue}{\faExternalLink*}} \\

31 & PhysGaussian~\cite{xie2024physgaussian} & CVPR '24
  & \textcolor{orange}{\faImage}
  & \textcolor{violet}{\faWaveSquare} 
  & 3DGS+MPM & Manual Params & Dyn. Field
  & \textcolor{red}{\ding{183}}
  & \href{https://arxiv.org/abs/2311.12198}{\textcolor{blue}{\faExternalLink*}} \\

32 & PhysPart~\cite{luo2025physpart} & ICRA '25
  & \textcolor{orange}{\faImage}~\textcolor{gray}{\faFont}
  & \textcolor{cyan}{\faCube}~\textcolor{red}{\faWind}
  & Diff & Text/Image & Mesh (Partial)
  & \textcolor{red}{\ding{183}}
  & \href{https://red-fairy.github.io/physpart-webpage/}{\textcolor{blue}{\faExternalLink*}} \\

33 & DSO~\cite{li2025dso} & ICCV '25
  & \textcolor{orange}{\faImage}~\textcolor{gray}{\faFont}
  & \textcolor{cyan}{\faCube}
  & Diff & Sim-Reward & Mesh
  & \textcolor{red}{\ding{183}}
  & \href{https://ruiningli.com/dso}{\textcolor{blue}{\faExternalLink*}} \\

34 & GaussianProperty~\cite{xu2025gaussianproperty} & ICCV '25
  & \textcolor{orange}{\faImage}
  & \textcolor{olive}{\faBalanceScale}
  & 3DGS+LLM/VLM & Zero-shot & Params Only
  & \textcolor{red}{\ding{183}}
  & \href{https://arxiv.org/abs/2412.11258}{\textcolor{blue}{\faExternalLink*}} \\

35 & PhysX-3D~\cite{cao2025physx3d} & NeurIPS '25
  & \textcolor{orange}{\faImage}~\textcolor{gray}{\faFont}
  & \textcolor{cyan}{\faCube}~\textcolor{olive}{\faBalanceScale}
  & VAE & 3D GT & URDF (Partial)
  & \textcolor{red}{\ding{183}}
  & \href{https://physx-3d.github.io/}{\textcolor{blue}{\faExternalLink*}} \\

36 & SOPHY~\cite{cao2025sophy} & WACV '26
  & \textcolor{orange}{\faImage}~\textcolor{gray}{\faFont}
  & \textcolor{cyan}{\faCube}~\textcolor{violet}{\faWaveSquare}
  & Diff & 3D GT & URDF/MJCF
  & \textcolor{red}{\ding{183}}
  & \href{https://xjay18.github.io/SOPHY_page/}{\textcolor{blue}{\faExternalLink*}} \\

37 & Pixie~\cite{le2025pixie} & arXiv '25
  & \textcolor{orange}{\faImage}
  & \textcolor{olive}{\faBalanceScale}
  & FF & Supervised & Params Only
  & \textcolor{red}{\ding{183}}
  & \href{https://pixie-3d.github.io/}{\textcolor{blue}{\faExternalLink*}} \\

38 & DensiCrafter~\cite{dang2026densicrafter} & AAAI '26
  & \textcolor{gray}{\faFont}
  & \textcolor{violet}{\faWaveSquare}
  & SDS+Optimization & None & Dyn. Field
  & \textcolor{red}{\ding{183}}
  & \href{https://arxiv.org/abs/2511.09298}{\textcolor{blue}{\faExternalLink*}} \\

39 & PhysX-Anything~\cite{cao2025physx} & CVPR '26
  & \textcolor{orange}{\faImage}
  & \textcolor{cyan}{\faCube}~\textcolor{purple}{\faProjectDiagram}~\textcolor{olive}{\faBalanceScale}
  & LLM/VLM & Text/Image & URDF/MJCF
  & \textcolor{red}{\ding{183}}
  & \href{https://physx-anything.github.io/}{\textcolor{blue}{\faExternalLink*}} \\

\midrule
40 & DressCode~\cite{he2024dresscode} & ToG (SIG.~'24)
  & \textcolor{gray}{\faFont}
  & \textcolor{cyan}{\faCube}
  & AR+Diff & Text/3D GT & Sew. Pat.
  & \textcolor{teal}{\ding{184}}
  & \href{https://ihe-kaii.github.io/DressCode/}{\textcolor{blue}{\faExternalLink*}} \\

41 & PhysDreamer~\cite{zhang2024physdreamer} & ECCV '24
  & \textcolor{cyan}{\faCube}~\textcolor{magenta}{\faVideo}
  & \textcolor{violet}{\faWaveSquare} 
  & Diff+MPM & Self-sup. & Dyn. Field
  & \textcolor{teal}{\ding{184}}
  & \href{https://physdreamer.github.io/}{\textcolor{blue}{\faExternalLink*}} \\

42 & GarmentDreamer~\cite{li2025garmentdreamer} & 3DV '25
  & \textcolor{gray}{\faFont}
  & \textcolor{cyan}{\faCube}~\textcolor{teal}{\faAtom} 
  & Diff+3DGS & Text/3DGS & Mesh
  & \textcolor{teal}{\ding{184}}
  & \href{https://arxiv.org/abs/2405.12420}{\textcolor{blue}{\faExternalLink*}} \\

43 & Dress-1-to-3~\cite{li2025dress} & ToG (SIG.~'25)
  & \textcolor{orange}{\faImage}
  & \textcolor{cyan}{\faCube}~\textcolor{olive}{\faBalanceScale}
  & FEM+Diff & Sim-Feedback & Sew. Pat.
  & \textcolor{teal}{\ding{184}}
  & \href{https://dress-1-to-3.github.io/}{\textcolor{blue}{\faExternalLink*}} \\

44 & Image2Garment~\cite{can2026image2garment} & arXiv '26
  & \textcolor{orange}{\faImage}
  & \textcolor{cyan}{\faCube}~\textcolor{olive}{\faBalanceScale}
  & FF & Supervised & Mesh + Params
  & \textcolor{teal}{\ding{184}}
  & \href{https://image2garment.github.io/}{\textcolor{blue}{\faExternalLink*}} \\

\midrule
45 & TRELLIS~\cite{xiang2025structured} & CVPR '25
  & \textcolor{gray}{\faFont}~\textcolor{orange}{\faImage}
  & \textcolor{cyan}{\faCube}~\textcolor{teal}{\faAtom}
  & Sp.DiT+VAE & 3D GT & Mesh/3DGS/RF
  & \textcolor{orange}{\ding{185}}
  & \href{https://microsoft.github.io/TRELLIS/}{\textcolor{blue}{\faExternalLink*}} \\

46 & TRELLIS.2~\cite{xiang2025native} & CVPR '26
  & \textcolor{gray}{\faFont}~\textcolor{orange}{\faImage}
  & \textcolor{cyan}{\faCube}~\textcolor{teal}{\faAtom}
  & Sp.DiT+VAE & 3D GT & Mesh/3DGS/RF
  & \textcolor{orange}{\ding{185}}
  & \href{https://microsoft.github.io/TRELLIS.2/}{\textcolor{blue}{\faExternalLink*}} \\

47 & EmbodiedGen~\cite{wang2025embodiedgen} & NeurIPS '25
  & \textcolor{gray}{\faFont}~\textcolor{orange}{\faImage}
  & \textcolor{purple}{\faProjectDiagram}~\textcolor{olive}{\faBalanceScale}
  & LLM/VLM+Diff & Text/Image & URDF/MJCF
  & \textcolor{orange}{\ding{185}}
  & \href{https://arxiv.org/abs/2506.10600}{\textcolor{blue}{\faExternalLink*}} \\

48 & Seed3D~\cite{seed2025seed3d} & arXiv '25
  & \textcolor{orange}{\faImage}
  & \textcolor{cyan}{\faCube}
  & Diff+VAE & 3D GT & Isaac Sim Ready
  & \textcolor{orange}{\ding{185}}
  & \href{https://seed.bytedance.com/en/seed3d}{\textcolor{blue}{\faExternalLink*}} \\

\bottomrule
\end{tabular}%
}
\end{table*}

\subsubsection{Articulated Objects}
\label{sec:articulated}

Embodied interaction requires kinematic structure specifying how parts move and constrain one another. Reconstruction-based approaches~\cite{paris, wu2025reartgs++} recover articulation from observations but generalize poorly to unseen categories. Generative approaches instead learn priors over geometry and motion structure for scalable articulated asset creation~\cite{chen2025artilatent, lei2023nap, lian2025infinite}. The key distinction is whether the method learns a \emph{category-level generative prior} or recovers \emph{instance-specific} parameters from dynamic observations of a particular object (Sec.~\ref{sec:digital_twin}).

\noindent\textbf{Structural Topology and Parameterization.}
The topology problem is inherently combinatorial: the parent-child graph among parts must be determined before joint axes and limits can be specified. Kinematify~\cite{wang2025kinematify} tackles this with MCTS using physical rewards to prune infeasible structures. GAOT~\cite{sun2025gaot} and NAP~\cite{lei2023nap} employ hierarchical learning for part connectivity, while UniArt~\cite{jin2025uniart} introduces a Joint-to-Voxel representation mapping kinematic constraints into continuous 3D volumes.
For motion parameter estimation, diffusion-based methods (CAGE~\cite{liu2024cage}, ArtFormer~\cite{su2025artformer}, ArtiLatent~\cite{chen2025artilatent}) jointly model geometry and kinematic distributions. PAct~\cite{liu2026pact} and SINGAPO~\cite{liu2024singapo} handle sparse-view settings via part-aware pipelines. MagicArticulate~\cite{song2025magicarticulate} reformulates skeleton generation as autoregressive sequence modeling with functional diffusion for skinning weights, converting static meshes into fully rigged assets. MeshArt~\cite{gao2025meshart} adopts a similar hierarchical autoregressive strategy, first predicting part-level bounding boxes and articulation modes before generating detailed mesh geometry with coherent inter-part connectivity. Particulate~\cite{li2025particulate} further simplifies the input requirement by using a feed-forward transformer to predict part segmentation, kinematic tree, and motion parameters directly from a single static mesh.

\noindent\textbf{Semantic Reasoning and Programmatic Synthesis.}
Purely geometric approaches struggle to infer latent functional relations from shape alone. Recent work incorporates LLM/VLM semantic priors: Articulate-Anything~\cite{le2024articulate} uses a VLM Actor-Critic framework with simulator feedback; URDF-Anything~\cite{li2025urdf} integrates segmentation-aware tokens for joint part decomposition and motion prediction; ArtLLM~\cite{wang2026artllm}, ArtiWorld~\cite{yang2025artiworld}, and SPARK~\cite{he2025spark} further leverage semantic reasoning for articulation inference.
Semantic reasoning naturally supports translation into simulator-executable formats. URDFormer~\cite{chen2024urdformer} regresses URDF parameters from images; Real2Code~\cite{mandi2024real2code} synthesizes executable motion programs via LLMs; URDF-Anything+~\cite{wu2026urdf} extends this to autoregressive end-to-end URDF generation. ArtGen~\cite{wang2025artgen} incorporates Chain-of-Thought reasoning, while Articulate AnyMesh~\cite{qiu2025articulate} and ATOP~\cite{vora2025articulate} demonstrate open-vocabulary articulation transfer to arbitrary meshes.

\noindent\textbf{Procedural Generation and Simulation-Ready Assets.}
Procedural approaches target simulator readiness directly by constructing assets with explicit structural and physical validity. Infinite Mobility~\cite{lian2025infinite} combines layout constraint solving with physics-aware procedural rules to synthesize articulated objects at scale. PhysX-Anything~\cite{cao2025physx} jointly generates geometry, articulation, and physical properties (mass, friction, inertia, material) from a single image, exporting directly in URDF and MJCF formats---marking a shift toward pipelines that produce directly deployable assets.

\noindent\textbf{Multi-State and Interaction-Driven Dynamics.}
Static geometry alone cannot fully reveal degrees of freedom. Recent methods incorporate cross-state observations and interaction cues: PartRM~\cite{gao2025partrm} models part-level 4D dynamics conditioned on user drag prompts; DreamArt~\cite{lu2025dreamart} injects motion priors from video generation models. These methods mark a transition toward temporally coherent and interaction-aware articulated synthesis.

\subsubsection{Physically-grounded Objects}
\label{sec:physical}

Simulation-oriented assets require compliance with physical constraints---structural stability, mass distribution, material response, and contact dynamics---that purely geometric pipelines ignore. Early efforts~\cite{physaware_gen, physlayer} introduced physical priors through inductive biases but remained limited to geometry-level consistency. Recent research has progressively moved from post-hoc refinement toward native physical generation.

\noindent\textbf{Physical Property Understanding.}
NeRF2Physics~\cite{zhai2024physical} distills physical attributes (density, friction, hardness) from language-embedded feature fields for zero-shot property prediction. GaussianProperty~\cite{xu2025gaussianproperty} combines SAM segmentation with GPT-4V reasoning to assign per-part material properties to 3DGS assets via multi-view voting. These methods establish physical property understanding as a foundation but still treat physical attributes as post-reconstruction annotation rather than an intrinsic generative component.

\noindent\textbf{Post-Hoc Physical Refinement.}
This strategy generates visually plausible assets first and then enforces physical consistency. Atlas3D~\cite{chen2024atlas3d} introduces differentiable simulation signals into text-to-3D optimization to penalize stability violations. PhysComp~\cite{guo2024physically} optimizes tetrahedral meshes under mechanical equilibrium constraints. DensiCrafter~\cite{dang2026densicrafter} refines internal density fields for stability without altering visible geometry. PhysPart~\cite{luo2025physpart} injects collision and contact losses into articulated generation sampling. While effective, post-hoc physics correction incurs substantial test-time cost and introduces tension between geometric fidelity and physical soundness.

\noindent\textbf{Simulation-Aligned Model Training.}
A more principled direction incorporates simulator feedback into training. DSO~\cite{li2025dso} constructs a preference dataset by labeling generated objects with stability scores from a non-differentiable simulator, then fine-tunes the generator via preference optimization to internalize physical stability. PhyCAGE~\cite{yan2024phycage} couples 3D Gaussian representations with physics-guided score distillation for compositional multi-object generation. These methods shift physics from a post-hoc repair signal to a supervisory signal shaping the generator itself.

\noindent\textbf{Native Unified Physical Generation.}
The most recent direction models geometry, appearance, and physical properties jointly within a single framework. PhysX-3D~\cite{cao2025physx3d} encodes geometry with scale, material, and kinematic attributes; PhysX-Anything~\cite{cao2025physx} extends this to image-conditioned generation with URDF/MJCF export. SOPHY~\cite{cao2025sophy} highlights the importance of spatially varying material fields for heterogeneous objects. PhysGaussian~\cite{xie2024physgaussian} couples 3DGS with MPM, unifying rendering and simulation within the same representation (Sec.~\ref{sec:deformable}). PIXIE~\cite{le2025pixie} takes a supervised feed-forward route, predicting physical parameters from large-scale annotated data. These methods embed physical reasoning into the representation itself, reducing reliance on external correction.

\begin{figure*}
    \centering
    \includegraphics[width=0.95\linewidth]{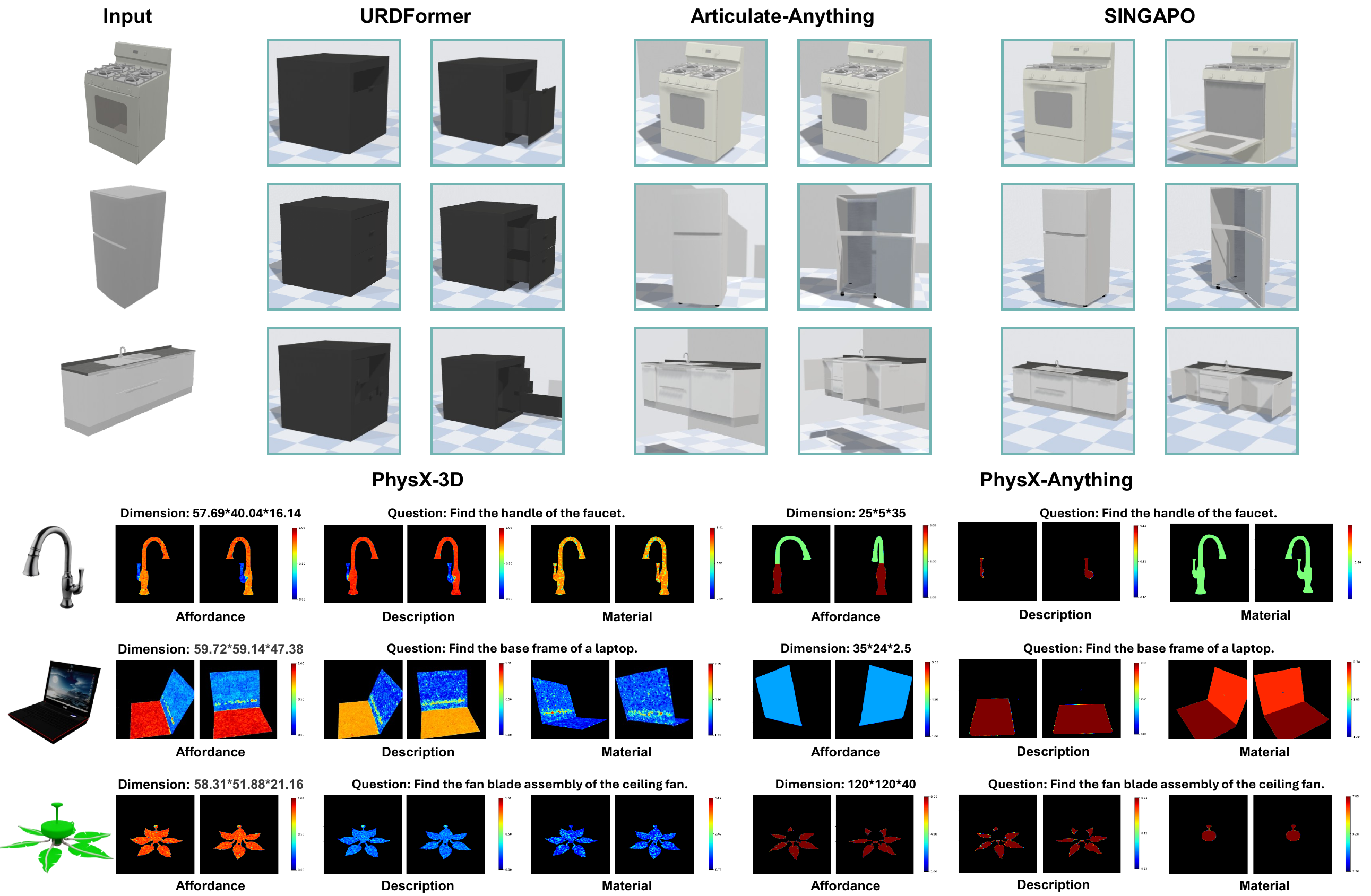}
    \caption{\textbf{Qualitative results of data generation methods.} \textit{Top}: articulated object generation results from URDFormer, Articulate-Anything, and SINGAPO. \textit{Bottom}: physically grounded object generation results from PhysX-3D and PhysX-Anything, showing affordance, description, and material heatmaps.}
    \label{fig:data-generator-example}
\end{figure*}

\subsubsection{Deformable Objects}
\label{sec:deformable}

Deformable objects require high-dimensional continuous state spaces governed by continuum mechanics (FEM, MPM, PBD) rather than discrete joints. Simulation-ready deformable assets need both rest-state geometry and material constitutive laws.
Classical simulation foundations underpin all downstream generative work: Baraff and Witkin~\cite{baraff1998large} introduced implicit integration for large-step cloth simulation, ArcSim~\cite{narain2012arcsim} contributed adaptive anisotropic remeshing, and DiffCloth~\cite{li2022diffcloth} enabled gradient-based optimization of material parameters through differentiable contact modeling.
Standardized benchmarks---SoftGym~\cite{lin2021softgym} for deformable manipulation, PlasticineLab~\cite{huang2021plasticinelab} for differentiable soft-body tasks, FluidLab~\cite{wu2023fluidlab} for fluid manipulation, and SoftZoo~\cite{wang2023softzoo} for soft robot co-design---define the evaluation landscape within which generative deformable methods must operate.
Progress in generative methods follows two directions: \textit{cloth and garment generation} for humanoid manipulation, and \textit{elastoplastic and soft-body generation} for contact-rich interactions.

\noindent\textbf{Cloth and Garment Generation.}
A key insight is that \textit{sewing patterns} provide a natural simulation-ready intermediate directly compatible with cloth simulators. DressCode~\cite{he2024dresscode} pioneers text-driven garment generation by producing sewing pattern sequences from language prompts with PBR texture synthesis. GarmentDreamer~\cite{li2025garmentdreamer} extends this to text-to-3D generation using 3DGS-based multi-view guidance while preserving simulation-compatible topology.
Dress-1-to-3~\cite{li2025dress} embeds a differentiable cloth simulator into the optimization loop to jointly recover sewing patterns and material properties from a single image. Image2Garment~\cite{can2026image2garment} improves scalability with a feed-forward framework predicting material composition in one pass. NeuralClothSim~\cite{bertiche2022neural} learns cloth dynamics directly from data, bypassing explicit constitutive modeling while preserving physically plausible behavior.
Downstream garment manipulation~\cite{ha2022flingbot, canberk2023clothfunnels} imposes additional requirements on generated assets: beyond geometric and material fidelity, compatibility with contact-rich dynamics is essential. GarmentLab~\cite{lu2024garmentlab} provides a GPU-parallel benchmark on PBD and FEM backends for evaluating generated garment assets under such manipulation scenarios.

\noindent\textbf{Elastoplastic and Soft-Body Generation.}
For soft objects (foam, ropes, food, granular materials), the central difficulty is \textit{material field estimation}---inferring constitutive parameters from observations.
PhysGaussian~\cite{xie2024physgaussian} couples 3DGS with MPM so that each Gaussian serves as both rendering primitive and simulation particle, supporting elastic, plastic, fluid-like, and granular dynamics without geometry conversion.
PhysDreamer~\cite{zhang2024physdreamer} removes manual material specification by optimizing spatial material fields so that MPM-simulated motion matches dynamics from a video generation prior. SOPHY~\cite{cao2025sophy} (Sec.~\ref{sec:physical}) further advances physics-prior-guided generation of spatially varying material fields.
While sewing-pattern pipelines have begun to make cloth assets simulation-ready at scale, general soft-body generation still requires costly per-instance optimization.

\subsection{End-to-End Generative Pipeline}
\label{sec:pipeline}

The main bottleneck for practical deployment is the \textit{integration gap}: even high-quality meshes require repair, UV mapping, material decomposition, scale calibration, collision geometry, and format serialization before entering physics engines. End-to-end pipelines address this by automating the full path from generation to simulator-ready output.

\noindent\textbf{Unified Generative Representations for Format-Versatile Output.}
A key requirement is a generative backbone that unifies geometry and appearance while supporting different output formats---earlier methods were fragmented between geometry-oriented models lacking realistic appearance and radiance-field models offering photorealism but limited simulator compatibility. TRELLIS~\cite{xiang2025structured} addresses this with a structured latent representation jointly encoding 3D structure and visual detail, enabling multiple decoders to produce Radiance Fields, 3DGS, or meshes from a shared code. Its successor TRELLIS.2~\cite{xiang2025native} adds geometric sharpness and native PBR support. These unified backbones serve both rendering quality and downstream physical deployment.

\noindent\textbf{Purpose-Built Simulation-Ready Generation Pipelines.}
Building on such backbones, complete toolchains transform raw inputs into simulation-ready assets. Seed3D 1.0~\cite{seed2025seed3d} couples geometry generation with multi-view texture synthesis, PBR estimation, UV completion, and data standardization, producing watertight textured meshes with scale metadata importable into Isaac Sim. EmbodiedGen~\cite{wang2025embodiedgen} takes a modular approach integrating image/text-to-3D generation, quality control, physical attribute estimation, URDF serialization, and layout construction into a unified toolkit. These pipelines show how generation systems are evolving from visual content creators into embodied asset engines. Their importance is not only automation, but also interface alignment: once scale, articulation, materials, and metadata are serialized together, the asset can move across simulators and downstream tasks with much less manual intervention.

\noindent\textbf{Automated Curation and Quality Control.}
Large-scale generation inevitably introduces defects, making quality assurance critical. Seed3D incorporates topology, semantic, and view-consistency checks to filter assets before simulation deployment. EmbodiedGen adopts a self-correcting loop that detects and re-triggers generation. Effective simulation-ready pipelines must combine strong generative backbones with closed-loop quality assessment and structured serialization. Beyond removing obvious failures, such curation determines whether generated assets are reliable enough for benchmark construction, policy training, and reproducible cross-method evaluation.

%% file: sec/simulation_environments.tex
\section{Simulation Environments}
\label{sec:sim_env}

Embodied agent training requires executable 3D environments that support perception, navigation, and manipulation at scale. Because building and labeling real environments is expensive, slow, and limited, automatic scene generation has become essential for scaling data and task diversity. Beyond geometry and appearance, generated environments must provide semantic structure, physically valid setups, interaction affordances, and the runtime metadata needed by simulators. We focus on indoor scenes, as most existing embodied AI tasks and platforms target indoor settings.

\begin{figure*}
    \centering
    \includegraphics[width=0.9\linewidth]{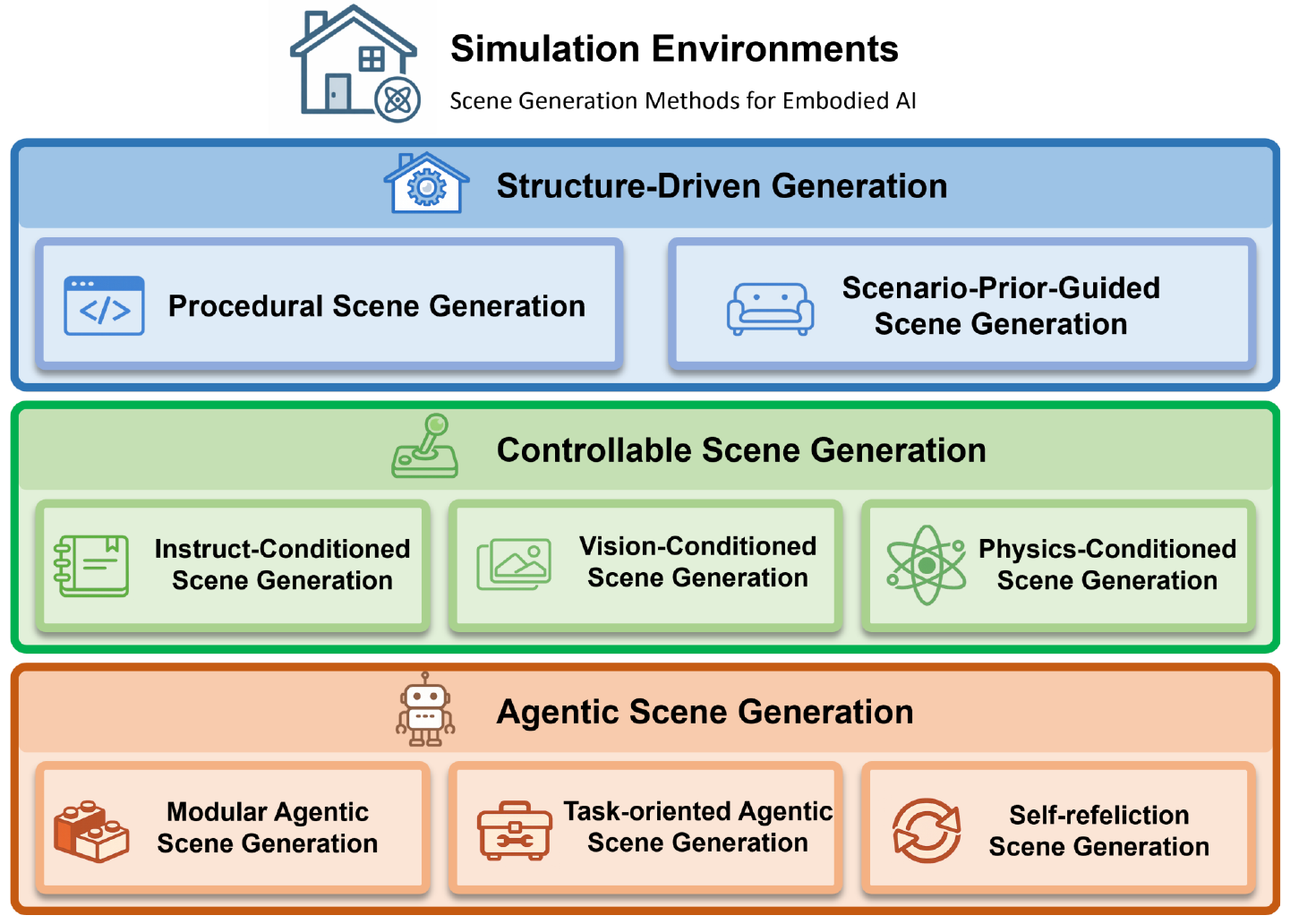}
    \caption{\textbf{Overview of indoor scene generation methods for embodied AI.} Methods progress from \textit{Structure-Driven} generation (procedural rules or learned priors), through \textit{Controllable} generation (instruction, vision, or physics conditioning), to \textit{Agentic} generation (foundation-model planning with self-correction and simulation feedback).}
    \label{fig:scene_generation}
\end{figure*}

\newcommand{\condN}{\textcolor{black}{\mbox{\large$\varnothing$}}}
\newcommand{\condT}{\textcolor{gray}{\faFont}}
\newcommand{\condI}{\textcolor{orange}{\faImage}}
\newcommand{\condP}{\textcolor{purple}{\faBalanceScale}}
\newcommand{\condTa}{\textcolor{red!70!black}{\raisebox{0.15ex}{\scriptsize\textsf{Ta}}}}
\newcommand{\condS}{\textcolor{blue!70!black}{\faCrosshairs}}
\newcommand{\condL}{\textcolor{teal!70!black}{\faTh}}

\newcommand{\simNA}{\textcolor{gray!70!black}{\raisebox{0.15ex}{\scriptsize\textsf{--}}}}
\newcommand{\simThor}{\textcolor{magenta!80!black}{\raisebox{0.15ex}{\scriptsize\textsf{THOR}}}}
\newcommand{\simHab}{\textcolor{violet!80!black}{\raisebox{0.15ex}{\scriptsize\textsf{Hab}}}}
\newcommand{\simIsaac}{\textcolor{olive!80!black}{\raisebox{0.15ex}{\scriptsize\textsf{Isaac}}}}
\newcommand{\simMujoco}{\textcolor{red!70!black}{\raisebox{0.15ex}{\scriptsize\textsf{MJ}}}}
\newcommand{\simBullet}{\textcolor{cyan!60!black}{\raisebox{0.15ex}{\scriptsize\textsf{Bullet}}}}
\newcommand{\simGenesis}{\textcolor{orange!85!black}{\raisebox{0.15ex}{\scriptsize\textsf{Genesis}}}}
\newcommand{\simMani}{\textcolor{green!50!black}{\raisebox{0.15ex}{\scriptsize\textsf{Mani3}}}}
\newcommand{\simBlender}{\textcolor{brown!75!black}{\raisebox{0.15ex}{\scriptsize\textsf{Blender}}}}
\newcommand{\simUnreal}{\textcolor{blue!60!black}{\raisebox{0.15ex}{\scriptsize\textsf{Unreal}}}}
\newcommand{\simDrake}{\textcolor{black!80!blue}{\raisebox{0.15ex}{\scriptsize\textsf{Drake}}}}
\newcommand{\simOmniverse}{\textcolor{pink!80!black}{\raisebox{0.15ex}{\scriptsize\textsf{Omniverse}}}}
\newcommand{\simLibulPC}{\textcolor{purple!80!black}{\raisebox{0.15ex}{\scriptsize\textsf{LibulPC}}}}

\begin{table*}[!t]
\caption{Summary of indoor scene generation methods for Embodied AI.}
\label{tab:embodied_scene_generation_summary}

\vspace{-10pt}
\begin{minipage}{\textwidth}
    \scriptsize
    \begin{itemize}[leftmargin=1.2em, nosep, itemsep=2pt, label=\small$\bullet$]
        \item \textbf{Input}:
          \condN~None,
          \condT~Text,
          \condI~Image,
          \condP~Physics,
          \condTa~Task specification,
          \condL~Layout,
          \condS~Scan.
        \item \textbf{Architectures (Arch):}
          \textit{\textbf{Proc}}: Procedural / rule-based generation,
          \textit{\textbf{AR}}: Autoregressive,
          \textit{\textbf{Diff}}: Diffusion model,
          \textit{\textbf{Transf}}: Transformer,
          \textit{\textbf{GNN}}: Graph neural network,
          \textit{\textbf{Impl}}: Implicit / neural field model,
          \textit{\textbf{LLM/VLM}}: Large / vision-language model,
          \textit{\textbf{FF}}: Feed-forward model,
          \textit{\textbf{Opt}}: Optimization-based,
          \textit{\textbf{Agent}}: Agentic multi-stage pipeline,
          \textit{\textbf{Search}}: Inference-time search.
        \item \textbf{Platform}:
          \simThor~AI2-THOR,
          \simHab~Habitat / Habitat~3.0,
          \simIsaac~Isaac~Sim,
          \simMujoco~MuJoCo,
          \simBullet~PyBullet,
          \simGenesis~Genesis,
          \simMani~ManiSkill3,
          \simBlender~Blender,
          \simUnreal~Unreal Engine,
          \simDrake~Drake,
          \simOmniverse~Omniverse,
          \simLibulPC~LibulPC,
          \simNA~no explicit simulator / platform named in the paper or official repo.
        \item \textbf{Categories (Cat)}:
          \textcolor{blue}{\ding{182}}~Structure-Driven (Sec.~\ref{sec:structure-driven-scene}),
          \textcolor{teal}{\ding{183}}~Controllable (Sec.~\ref{sec:controllable-scene-generation}),
          \textcolor{red}{\ding{184}}~Agentic (Sec.~\ref{sec:agentic-scene}).
    \end{itemize}
\end{minipage}
\rowcolors{2}{gray!20}{white}
\resizebox{\textwidth}{!}{%
\scriptsize
\begin{tabular}{@{}c l l c c c c c@{}}
\toprule
\textbf{\#} & \textbf{Method} & \textbf{Venue} & \textbf{Input} & \textbf{Arch} & \textbf{Platform} & \textbf{Cat} & \textbf{URL} \\
\midrule
1 & Graph-to-3D~\cite{graph_to_3d} & ICCV '21 & \condL & GNN+VAE & \simNA & \textcolor{blue}{\ding{182}} & \href{https://he-dhamo.github.io/Graphto3D/}{\textcolor{blue}{\faExternalLink*}} \\

2 & ATISS~\cite{atiss} & NeurIPS '21 & \condL & AR+Transf & \simOmniverse & \textcolor{blue}{\ding{182}} & \href{https://research.nvidia.com/labs/toronto-ai/ATISS/}{\textcolor{blue}{\faExternalLink*}} \\

3 & ProcTHOR~\cite{procthor} & NeurIPS '22 & \condN & Proc & \simThor & \textcolor{blue}{\ding{182}} & \href{https://procthor.allenai.org/}{\textcolor{blue}{\faExternalLink*}} \\

4 & CC3D~\cite{cc3d} & ICCV '23 & \condI & Impl & \simNA & \textcolor{blue}{\ding{182}} & \href{https://sherwinbahmani.github.io/cc3d/}{\textcolor{blue}{\faExternalLink*}} \\

5 & LayoutGPT~\cite{layoutgpt} & NeurIPS '23 & \condT & LLM & \simNA & \textcolor{blue}{\ding{182}} & \href{https://layoutgpt.github.io/}{\textcolor{blue}{\faExternalLink*}} \\

6 & Infinigen Indoors~\cite{infinigen_indoors} & CVPR '24 & \condN & Proc & \simBlender & \textcolor{blue}{\ding{182}} & \href{https://arxiv.org/abs/2406.11824}{\textcolor{blue}{\faExternalLink*}} \\

7 & Holodeck~\cite{holodeck} & CVPR '24 & \condT & LLM & \simThor & \textcolor{blue}{\ding{182}} & \href{https://yueyang1996.github.io/holodeck/}{\textcolor{blue}{\faExternalLink*}} \\

8 & DiffuScene~\cite{diffuscene} & CVPR '24 & \condN & Diff & \simNA & \textcolor{blue}{\ding{182}} & \href{https://tangjiapeng.github.io/projects/DiffuScene/}{\textcolor{blue}{\faExternalLink*}} \\

\midrule

9 & PhyScene~\cite{physcene} & CVPR '24 & \condP~\condL & Diff & \simNA & \textcolor{teal}{\ding{183}} & \href{https://physcene.github.io/}{\textcolor{blue}{\faExternalLink*}} \\

10 & InstructScene~\cite{instructscene} & ICLR '24 & \condT & Diff+GNN & \simNA & \textcolor{teal}{\ding{183}} & \href{https://chenguolin.github.io/projects/InstructScene/}{\textcolor{blue}{\faExternalLink*}} \\

11 & DepR~\cite{DepR} & ICCV '25 & \condI & Diff & \simNA & \textcolor{teal}{\ding{183}} & \href{https://mlpc-ucsd.github.io/DepR/}{\textcolor{blue}{\faExternalLink*}} \\

12 & MIDI~\cite{midi} & CVPR '25 & \condI & Diff & \simNA & \textcolor{teal}{\ding{183}} & \href{https://huanngzh.github.io/MIDI-Page/}{\textcolor{blue}{\faExternalLink*}} \\

13 & DynScene~\cite{dynscene} & CVPR '25 & \condT & LLM/VLM & \simIsaac & \textcolor{teal}{\ding{183}} & \href{https://openaccess.thecvf.com/content/CVPR2025/papers/Lee_DynScene_Scalable_Generation_of_Dynamic_Robotic_Manipulation_Scenes_for_Embodied_CVPR_2025_paper.pdf}{\textcolor{blue}{\faExternalLink*}} \\

14 & FactoredScenes~\cite{factoredscenes} & NeurIPS '25 & \condL & Proc & \simNA & \textcolor{teal}{\ding{183}} & \href{https://github.com/joyhsu0504/factored_scenes}{\textcolor{blue}{\faExternalLink*}} \\

15 & Steerable Scene Generation~\cite{steerable_scene_generation} & CoRL '25 & \condT & Diff+Search & \simDrake & \textcolor{teal}{\ding{183}} & \href{https://steerable-scene-generation.github.io/}{\textcolor{blue}{\faExternalLink*}} \\

16 & SceneFoundry~\cite{scenefoundry} & arXiv '26 & \condT & LLM+Diff & \simBlender & \textcolor{teal}{\ding{183}} & \href{https://arxiv.org/abs/2601.05810}{\textcolor{blue}{\faExternalLink*}} \\

17 & SceneGen~\cite{scenegen} & 3DV '26 & \condI & FF & \simBlender & \textcolor{teal}{\ding{183}} & \href{https://mengmouxu.github.io/SceneGen/}{\textcolor{blue}{\faExternalLink*}} \\

\midrule

18 & SceneCraft~\cite{scenecraft} & ICML '24 & \condT & LLM+Agent & \simBlender & \textcolor{red}{\ding{184}} & \href{https://arxiv.org/abs/2403.01248}{\textcolor{blue}{\faExternalLink*}} \\

19 & Architect~\cite{architect} & NeurIPS '24 & \condT~\condI & Diff & \simGenesis & \textcolor{red}{\ding{184}} & \href{https://wangyian-me.github.io/Architect/}{\textcolor{blue}{\faExternalLink*}} \\

20 & OptiScene~\cite{optiscene} & NeurIPS '25 & \condT & LLM+Opt & \simNA & \textcolor{red}{\ding{184}} & \href{https://polysummit.github.io/optiscene.github.io/}{\textcolor{blue}{\faExternalLink*}} \\

21 & LayoutVLM~\cite{layoutvlm} & CVPR '25 & \condT~\condI & VLM+Opt & \simNA & \textcolor{red}{\ding{184}} & \href{https://ai.stanford.edu/~sunfanyun/layoutvlm/}{\textcolor{blue}{\faExternalLink*}} \\

22 & SceneWeaver~\cite{sceneweaver} & NeurIPS '25 & \condT & Agent & \simBlender~\simIsaac & \textcolor{red}{\ding{184}} & \href{https://scene-weaver.github.io/}{\textcolor{blue}{\faExternalLink*}} \\

23 & CAST~\cite{cast} & TOG '25 & \condI & Diff+VLM & \simNA & \textcolor{red}{\ding{184}} & \href{https://sites.google.com/view/cast4}{\textcolor{blue}{\faExternalLink*}} \\

24 & 3D-RE-GEN~\cite{3dregen} & CVPR Find. '26 & \condI & Diff+Opt & \simNA & \textcolor{red}{\ding{184}} & \href{https://3dregen.jdihlmann.com/}{\textcolor{blue}{\faExternalLink*}} \\

25 & MetaScenes~\cite{metascenes} & CVPR '25 & \condS & VLM+Opt & \simHab & \textcolor{red}{\ding{184}} & \href{https://meta-scenes.github.io/}{\textcolor{blue}{\faExternalLink*}} \\

26 & DisCo-Layout~\cite{disco} & arXiv '25 & \condT & Agent & \simNA & \textcolor{red}{\ding{184}} & \href{https://arxiv.org/abs/2510.02178}{\textcolor{blue}{\faExternalLink*}} \\

27 & MarketGen~\cite{marketgen} & arXiv '25 & \condTa & Proc+Agent & \simIsaac~\simUnreal & \textcolor{red}{\ding{184}} & \href{https://xuhu0529.github.io/MarketGen}{\textcolor{blue}{\faExternalLink*}} \\

28 & MesaTask~\cite{mesatask} & NeurIPS '25 & \condTa & LLM & \simIsaac & \textcolor{red}{\ding{184}} & \href{https://mesatask.github.io/}{\textcolor{blue}{\faExternalLink*}} \\

29 & TabletopGen~\cite{tabletopgen} & ECCV '26 & \condTa~\condI~\condT & VLM+Diff & \simIsaac & \textcolor{red}{\ding{184}} & \href{https://d-robotics-ai-lab.github.io/TabletopGen.project/}{\textcolor{blue}{\faExternalLink*}} \\

30 & PAT3D~\cite{pat3d} & ICLR '26 & \condT & Diff+Phys & \simLibulPC & \textcolor{red}{\ding{184}} & \href{https://arxiv.org/abs/2511.21978}{\textcolor{blue}{\faExternalLink*}} \\

31 & PhyScensis~\cite{physcensis} & ICLR '26 & \condT~\condP & LLM+Phys & \simGenesis & \textcolor{red}{\ding{184}} & \href{https://physcensis.github.io/}{\textcolor{blue}{\faExternalLink*}} \\

32 & SAGE~\cite{sage} & CVPR '26 & \condT & Agent & \simIsaac & \textcolor{red}{\ding{184}} & \href{https://nvlabs.github.io/sage/}{\textcolor{blue}{\faExternalLink*}} \\

33 & SceneSmith~\cite{scenesmith} & ICML '26 & \condT & Agent & \simDrake~\simMujoco~\simIsaac & \textcolor{red}{\ding{184}} & \href{https://scenesmith.github.io/}{\textcolor{blue}{\faExternalLink*}} \\

34 & Scenethesis~\cite{scenethesis} & ICLR '26 & \condT & Agent & \simBlender & \textcolor{red}{\ding{184}} & \href{https://arxiv.org/abs/2505.02836}{\textcolor{blue}{\faExternalLink*}} \\

35 & 3D-Generalist~\cite{3d-generalist} & 3DV '26 & \condT~\condI & VLM & \simBlender & \textcolor{red}{\ding{184}} & \href{https://arxiv.org/abs/2507.06484}{\textcolor{blue}{\faExternalLink*}} \\

\bottomrule
\end{tabular}}
\end{table*}

\subsection{Structure-Driven Scene Generation}
\label{sec:structure-driven-scene}
Early methods produce a machine-readable scene specification (room layout, object list with poses and sizes) and then retrieve and instantiate assets from an existing library.
Prior to the current wave of embodied-AI--focused generation, the graphics and vision communities developed foundational techniques for data-driven indoor scene synthesis. Deep Convolutional Priors~\cite{wang2018deep} learned spatial object co-occurrence statistics from exemplar scenes. Fast \& Flexible~\cite{ritchie2019fast} introduced deep generative models for rapid indoor layout sampling. 3D-SLN~\cite{luo2020end} proposed end-to-end differentiable scene layout optimization. SceneFormer~\cite{wang2021sceneformer} applied transformers to autoregressively generate indoor scenes conditioned on room geometry, while LEGO-Net~\cite{wei2023legonet} tackled regular rearrangement via iterative refinement. These works established the core technical vocabulary---object-centric generation, layout optimization, and learned spatial priors---upon which current embodied scene generation builds.
Structure-driven methods fall into two categories: \textit{procedural-based} methods that encode feasibility through explicit rules, and \textit{scenario-prior--guided} methods that learn layout distributions from data.
\subsubsection{Procedural-Based Scene Generation}
ProcTHOR~\cite{procthor} samples floor plans, room topologies, and object placements under explicit constraints to generate large-scale interactive indoor environments.
Infinigen Indoors~\cite{infinigen_indoors} uses hierarchical procedural rules in Blender to produce diverse, photorealistic scenes exportable with metadata.
LayoutGPT~\cite{layoutgpt} uses an LLM to translate text prompts into structured layout specifications grounded via asset retrieval.
Holodeck~\cite{holodeck} converts language descriptions into relational constraints and optimizes object configurations before instantiation.
These methods differ primarily in control granularity: ProcTHOR and Infinigen Indoors operate at the room-topology and object-geometry levels through hand-coded rules, while LayoutGPT and Holodeck elevate control to natural language, trading fine-grained geometric specification for open-vocabulary flexibility.
Despite this range, their scalability is ultimately bounded by hand-crafted pipelines and manually specified rules.

\subsubsection{Scenario-Prior-Guided Scene Generation}
Instead of hand-crafted rules, these methods learn statistical regularities over structured scene representations and sample layouts from the learned distribution, with feasibility handled through post-processing or hybrid constraints.
Graph-to-3D~\cite{graph_to_3d} predicts object-level 3D representations from relational scene graphs.
ATISS~\cite{atiss} autoregressively generates indoor scenes as sets of object bounding boxes and attributes.
CC3D~\cite{cc3d} synthesizes compositional 3D scenes conditioned on 2D semantic layouts.
DiffuScene~\cite{diffuscene} performs denoising in object-attribute space to generate diverse configurations before asset instantiation.
These methods span different generative spaces---relational graphs (Graph-to-3D), sequential object attributes (ATISS), 2D semantic maps (CC3D), and continuous denoising fields (DiffuScene)---each offering distinct trade-offs between structural expressiveness and sampling diversity.
However, plausibility alone does not ensure embodied executability, and additional validation and asset-level metadata are often necessary. Critically, all structure-driven methods depend on external asset libraries (e.g., 3D-FUTURE, Objaverse) for object instantiation, meaning that scene diversity is ultimately bounded by asset coverage rather than by the layout generator itself.

\subsection{Controllable Scene Generation}
\label{sec:controllable-scene-generation}
Structure-driven methods offer only coarse control. Controllable scene generation introduces flexible conditioning signals---language instructions, images, or physics constraints---to steer layout synthesis toward specific requirements. This direction builds on a rich line of prior work in conditioned 3D scene synthesis: Text2Room~\cite{hollein2023text2room} and SceneScape~\cite{fridman2023scenescape} demonstrated text-driven iterative 3D room generation via 2D diffusion priors; CommonScenes~\cite{zhai2023commonscenes} converted scene graphs into 3D scenes through graph-conditioned diffusion; Set-the-Scene~\cite{cohenbar2023setthescene} enabled compositional NeRF scene generation with global-local training; and Ctrl-Room~\cite{fang2023ctrlroom} generated room-scale meshes from layout constraints. While these methods target visual content creation rather than simulator-ready output, they established the conditioning paradigms (text, scene graph, layout, image) that embodied scene generation inherits.

\subsubsection{Instruction-Conditioned Scene Generation}
Instruction-conditioned methods use natural language as the primary control channel to translate high-level intent into structured scene specifications.
InstructScene~\cite{instructscene} introduces an intermediate structured representation to align text instructions with 3D indoor layouts.
SceneFoundry~\cite{scenefoundry} targets apartment-scale generation by combining language guidance with diffusion-based synthesis and physical feasibility signals.
Steerable Scene Generation~\cite{steerable_scene_generation} adapts a generative model toward downstream objectives via post-training and inference-time search, providing a general mechanism for steering generation toward task-relevant constraints.

\subsubsection{Vision-Conditioned Scene Generation}
Vision-conditioned methods generate scenes consistent with observed geometry from images, making them natural for reconstruction and completion.
SceneGen~\cite{scenegen} generates multi-object 3D scenes from a single image by predicting geometry, texture, and spatial arrangement compositionally.
DepR~\cite{DepR} conditions on depth cues for instance-level generation to improve observation alignment.
MIDI~\cite{midi} extends diffusion to multi-instance scene generation from a single image with cross-object interaction modeling.

\subsubsection{Physics-Conditioned Scene Generation}
PhyScene~\cite{physcene} incorporates physics and interactivity guidance into layout generation to bridge the gap between visual plausibility and simulator usability. DynScene~\cite{dynscene} targets dynamic manipulation scenarios by decomposing generation into static scene synthesis and action-related components with physical refinement. FactoredScenes~\cite{factoredscenes} factorizes room structure and object poses using program-like generation to control reusable placement patterns. Across these three conditioning modalities, instruction-based methods offer the broadest user accessibility but rely on language grounding accuracy, vision-based methods provide geometric specificity but require robust depth and segmentation, and physics-based methods most directly target simulator usability but depend on accurate physical priors. Practical embodied readiness requires combining these signals, as no single modality simultaneously addresses grounding quality, geometric fidelity, and simulator-specific asset metadata.

\begin{figure*}[t]
    \centering
    \includegraphics[width=0.9\linewidth]{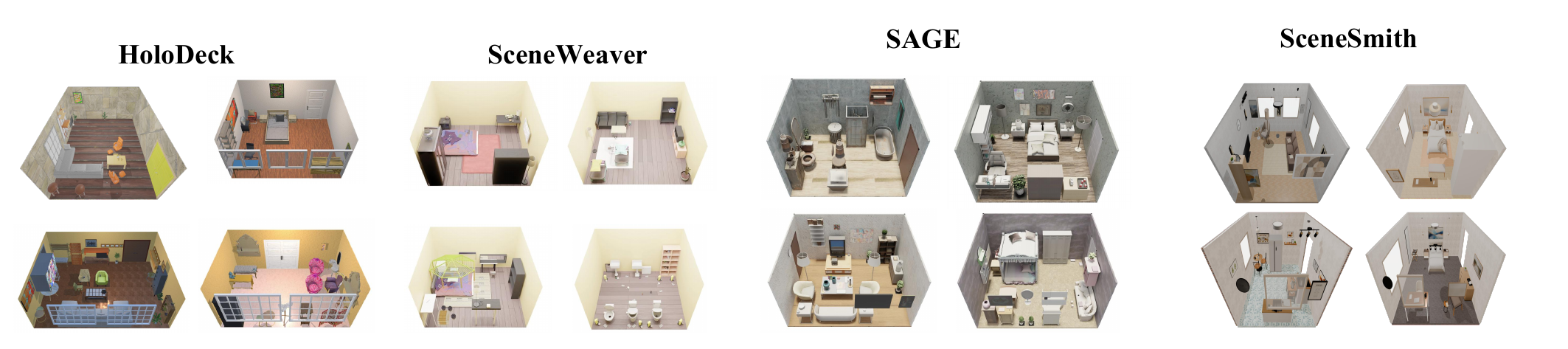}
    \caption{\textbf{Representative examples of simulation environment generation for embodied AI.} HoloDeck typically produces structured layouts with relatively constrained object arrangements. In contrast, SceneWeaver, SAGE, and SceneSmith generate more diverse and compositionally rich scenes, featuring improved object variety and spatial complexity. Samples in this figure are from SAGE and SceneSmith.}
    \label{fig:scene_generation-example}
\end{figure*}

\subsection{Agentic Scene Generation}
\label{sec:agentic-scene}
Foundation models enable agentic scene generation that translates high-level intent into structured specifications and continuously corrects semantic and physical mismatches through tool use and feedback, improving robustness over single-pass methods.

\subsubsection{Modular Agentic Scene Generation}
Modular agents connect planning, retrieval, layout solving, and verification through structured intermediate representations.
OptiScene~\cite{optiscene} improves layout alignment via staged preference optimization.
LayoutVLM~\cite{layoutvlm} combines vision-language grounding with differentiable optimization to refine layouts toward physical plausibility.
Scenethesis~\cite{scenethesis} uses an LLM to draft coarse layouts and then applies vision-guided refinement with a judging step to mitigate interpenetrations.
SceneCraft~\cite{scenecraft} takes a code-generation approach, using an LLM agent to produce Blender-executable scripts from text descriptions with iterative visual feedback and library learning for continuous self-improvement.
Beyond language-first pipelines, vision-based modular methods reconstruct editable scenes from observations:
CAST~\cite{cast} reconstructs 3D scenes from a single RGB image via segmentation, depth estimation, and physics-aware correction.
3D-RE-GEN~\cite{3dregen} integrates segmentation, inpainting, and constrained optimization for separable scene representations.
MetaScenes~\cite{metascenes} automates simulator-ready replica creation from scans via scalable asset replacement.
3D-Generalist~\cite{3d-generalist} formulates environment building as sequential decision-making with VLM-driven world-building actions.
Architect~\cite{architect} uses hierarchical 2D inpainting to iteratively generate layouts before lifting to 3D.
While modular agents separate planning from geometric grounding, performance depends on the weakest module and interface errors can accumulate across stages.

\subsubsection{Self-reflection Agentic Scene Generation}
Self-reflection methods incorporate explicit evaluation and repair loops, using foundation-model critics or simulation feedback to diagnose and fix semantic or physical mismatches.
SceneWeaver~\cite{sceneweaver} unifies multiple synthesis tools into a reflective framework with iterative planning, evaluation, and targeted repair.
SAGE~\cite{sage} couples layout and object-composition generators with learned critics for semantic plausibility and physical stability.
DisCo-Layout~\cite{disco} separates semantic from physical refinement via coordinated multi-agent roles.
Several works further incorporate physics-in-the-loop:
PAT3D~\cite{pat3d} integrates rigid-body simulation into text-to-3D generation to reduce interpenetration.
PhyScensis~\cite{physcensis} formulates physical relationships as predicates executed by a physics solver, using failure diagnostics to drive LLM-agent revisions.
SceneSmith~\cite{scenesmith} uses staged role-based agentic construction with verification and physical property estimation across stages.
The closed-loop process improves robustness in open-vocabulary and cluttered settings but incurs substantial computational overhead, and correction quality depends on evaluator reliability.

\subsubsection{Task-oriented Agentic Scene Generation}
Task-driven methods bind generation to downstream training or evaluation objectives, using task specifications (subgoals, reachability, manipulation feasibility) rather than appearance descriptions as control signals.
MarketGen~\cite{marketgen} generates scalable supermarket environments via agent-based procedural generation coupled with domain-specific benchmark tasks.
MesaTask~\cite{mesatask} generates structured tabletop layouts by mapping task descriptions into object lists and relations via a reasoning chain.
TabletopGen~\cite{tabletopgen} constructs interactive tabletop scenes from text or a single image using a compositional pipeline with simulator readiness.
Task-driven methods align scene generation directly with embodied training needs, but the generated distribution may be constrained by available assets and task design, and cross-simulator reuse requires careful validation. More broadly, the progression from structure-driven to agentic methods in this section reveals an increasing reliance on foundation-model inference at generation time: while this improves semantic alignment, few works report generation cost or throughput, making it unclear whether agentic pipelines can produce the thousands of environments needed for large-scale policy training.

%% file: sec/sim2real_bridge.tex
\section{Sim2Real Bridge}
\label{sec:sim2real}

\begin{figure*}
    \centering
    \includegraphics[width=0.95\linewidth]{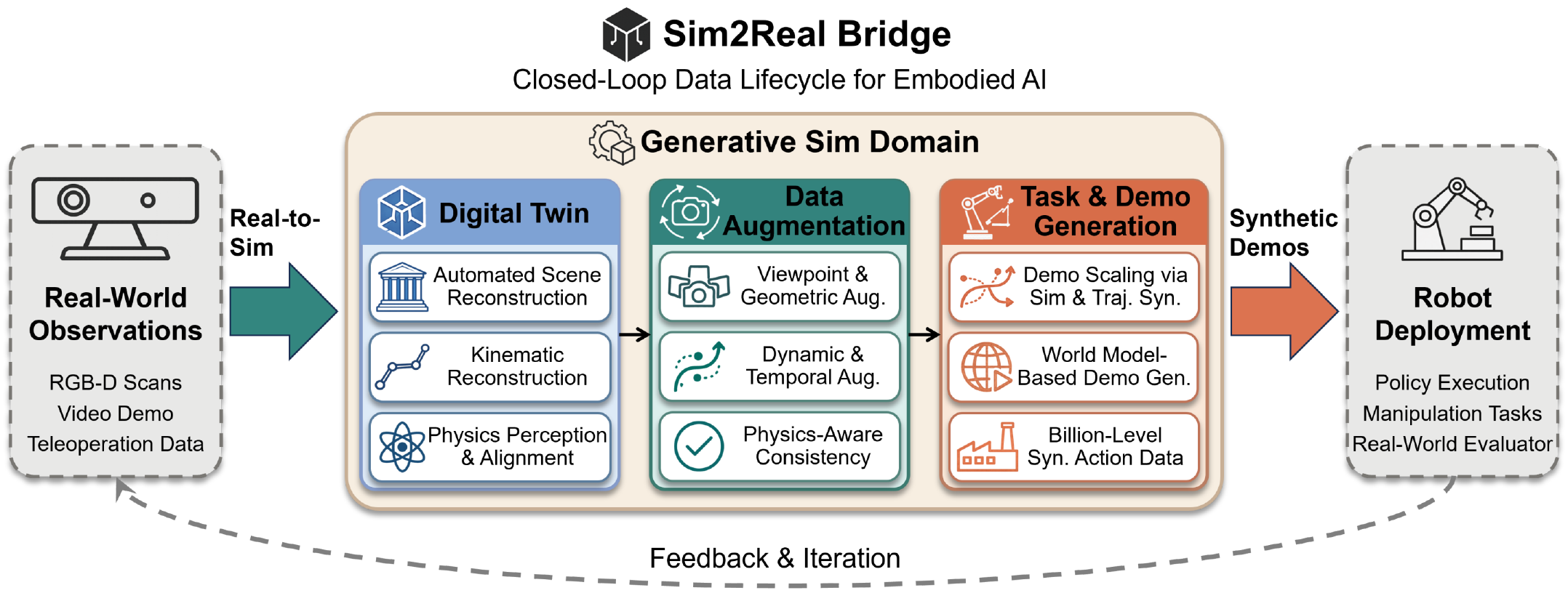}
    \caption{\textbf{Overview of the Sim2Real Bridge as a closed-loop data lifecycle for embodied AI.} Real-world observations are converted into \textit{Digital Twins} (Sec.~\ref{sec:digital_twin}), augmented through 3D-grounded \textit{Data Augmentation} (Sec.~\ref{sec:data_aug}), and scaled via \textit{Task \& Demo Generation} (Sec.~\ref{sec:gen_task_demo}). The resulting synthetic data trains real-robot policies, whose execution feedback iteratively refines the simulation domain.}
    \label{fig:sim2real}
\end{figure*}

While previous sections focus on generating \emph{new} 3D assets (Sec.~\ref{sec:data_generator}) and composing them into interactive environments (Sec.~\ref{sec:sim_env}), deploying these capabilities for robot learning requires bridging the gap between synthetic generation and real-world execution. This section examines how generative 3D representations support this bridge across three sequential dimensions: constructing physically faithful digital twins by \emph{reconstructing} real-world objects and scenes from observations (Sec.~\ref{sec:digital_twin}), augmenting limited demonstrations via 3D-grounded data synthesis based on these reconstructed assets (Sec.~\ref{sec:data_aug}), and scaling task-relevant demonstration data through simulation and generative models for real-world deployment (Sec.~\ref{sec:gen_task_demo}). Note that some underlying techniques---particularly for articulated objects---overlap with those in Sec.~\ref{sec:data_generator}; the distinction is that methods here target instance-level reconstruction of \emph{specific} real-world objects rather than category-level generation of \emph{novel} assets.

\subsection{Generative Digital Twin}
\label{sec:digital_twin}
The goal of generative digital twins is to automate the construction of simulation-ready replicas from real-world observations, replacing manual modeling with generation pipelines that produce not only visually faithful but also physically interactable environments.

\subsubsection{Automated Scene Reconstruction}
The most direct route to digital twin construction is automated pipelines that build interactive 3D assets from RGB-D inputs without manual intervention. DRAWER~\cite{drawer} and LiteReality~\cite{litereality} convert raw visual and depth data into interactive environments suitable for graphics engines. LatticeWorld~\cite{latticeworld} further integrates LLMs to coordinate multi-step reconstruction, producing semantically annotated virtual replicas from multimodal inputs. RoboSimGS~\cite{robosimgs} constructs a hybrid 3DGS-mesh representation from multi-view images and leverages a multimodal LLM to infer physical properties, enabling zero-shot sim-to-real manipulation transfer. Scalable Real2Sim~\cite{scalable_robosimgs} automates asset creation within robotic pick-and-place setups, estimating collision geometry and mass-inertial parameters from force-torque sensing and visual inputs for scalable simulation-ready asset production.
 
\subsubsection{Kinematic Reconstruction}
Functional digital twins must articulate correctly: revolute and prismatic joint parameters need to be recovered from observations so that simulated objects mirror motion constraints of their real-world counterparts. While the generative articulation methods in Sec.~\ref{sec:articulated} learn category-level priors to synthesize novel assets, kinematic reconstruction targets instance-level recovery from observations of a specific real-world object. The two share technical building blocks but differ in their input assumptions and generalization goals.

Ditto~\cite{jiang2022ditto} jointly optimizes geometry and articulation models from interactive 3D point clouds using implicit neural representations. For dynamic observation sequences, PARIS~\cite{paris} and ArticFlow~\cite{lin2025articflow} employ flow matching and deformation fields to establish dense correspondences across motion states, decoupling geometric reconstruction from motion estimation to resolve ambiguities in partial or single-view inputs. More recently, 3DGS-based methods have improved both rendering fidelity and kinematic tracking for digital twins. ArticulatedGS~\cite{guo2025articulatedgs} associates Gaussian primitives with rigid parts to enable joint state optimization alongside real-time rendering. ArtGS~\cite{liu2025artgs} and REArtGS++~\cite{wu2025reartgs++} extend this approach to reconstruct complex joint topologies and view-dependent appearances from multi-view or monocular videos, coupling kinematic priors with radiance field rendering to produce reconstructions suitable for downstream physics simulation and manipulation.

\begin{table*}[!t]
\caption{Summary of Sim2Real Bridge methods for embodied AI.}
\label{tab:sim2real_summary}

\vspace{-5pt}
\begin{minipage}{\textwidth}
    \scriptsize
    \begin{itemize}[leftmargin=1.2em, nosep, itemsep=2pt, label=\small$\bullet$]
        \item \textbf{Input/Output}:
          \textcolor{orange}{\faImage}~Img,
          \textcolor{brown}{\faCubes}~RGB-D,
          \textcolor{magenta}{\faVideo}~Video,
          \textcolor{blue}{\faBraille}~PC,
          \textcolor{cyan}{\faCube}~Mesh/Sim-Ready,
          \textcolor{gray}{\faFont}~Text,
          \textcolor{red}{\faWind}~Demo/Traj,
          \textcolor{olive}{\faBalanceScale}~Force,
          \textcolor{purple}{\faProjectDiagram}~URDF,
          \textcolor{teal}{\faAtom}~Digital Twin,
          \textcolor{green!60!black}{\faFileCode}~Action.
        \item \textbf{3D Rep}:
          3DGS, NeRF, \textit{\textbf{Impl}}: Implicit, Mesh, SDF, PC,
          \textit{\textbf{Occ}}: Occupancy, \textit{\textbf{Vid}}: Video Diffusion, \textit{\textbf{Sim}}: Simulator.
        \item \textbf{Dir}:
          S{\textrightarrow}R = Sim-to-Real,
          R{\textrightarrow}S = Real-to-Sim.
        \item \textbf{Cat}:
          \textcolor{teal}{\ding{182}}~Digital Twin (Sec.~\ref{sec:digital_twin}),
          \textcolor{cyan}{\ding{183}}~Data Aug.\ (Sec.~\ref{sec:data_aug}),
          \textcolor{red}{\ding{184}}~Task \& Demo (Sec.~\ref{sec:gen_task_demo}).
    \end{itemize}
\end{minipage}
\rowcolors{2}{gray!20}{white}
{
\renewcommand{\arraystretch}{0.95}
\resizebox{\textwidth}{!}{%
\scriptsize
\begin{tabular}{@{}c l l c c c c c c@{}}
\toprule
\textbf{\#} & \textbf{Method} & \textbf{Venue} & \textbf{Input} & \textbf{3D Rep} & \textbf{Dir} & \textbf{Output} & \textbf{Cat} & \textbf{URL} \\
\midrule
1 & Ditto~\cite{jiang2022ditto} & CVPR '22
  & \textcolor{blue}{\faBraille}
  & Impl & R\textrightarrow S
  & \textcolor{cyan}{\faCube}~\textcolor{purple}{\faProjectDiagram}
  & \textcolor{teal}{\ding{182}}
  & \href{https://ut-austin-rpl.github.io/Ditto/}{\textcolor{blue}{\faExternalLink*}} \\

2 & PARIS~\cite{paris} & ICCV '23
  & \textcolor{orange}{\faImage}
  & Impl & R\textrightarrow S
  & \textcolor{cyan}{\faCube}~\textcolor{purple}{\faProjectDiagram}
  & \textcolor{teal}{\ding{182}}
  & \href{https://3dlg-hcvc.github.io/paris/}{\textcolor{blue}{\faExternalLink*}} \\

3 & DRAWER~\cite{drawer} & CVPR '25
  & \textcolor{brown}{\faCubes}
  & Mesh & R\textrightarrow S
  & \textcolor{cyan}{\faCube}
  & \textcolor{teal}{\ding{182}}
  & \href{https://xiahongchi.github.io/DRAWER/}{\textcolor{blue}{\faExternalLink*}} \\

4 & LiteReality~\cite{litereality} & NeurIPS '25
  & \textcolor{brown}{\faCubes}
  & Mesh & R\textrightarrow S
  & \textcolor{cyan}{\faCube}
  & \textcolor{teal}{\ding{182}}
  & \href{https://litereality.github.io/}{\textcolor{blue}{\faExternalLink*}} \\

5 & LatticeWorld~\cite{latticeworld} & arXiv '25
  & \textcolor{brown}{\faCubes}~\textcolor{gray}{\faFont}
  & Mesh & R\textrightarrow S
  & \textcolor{cyan}{\faCube}~\textcolor{teal}{\faAtom}
  & \textcolor{teal}{\ding{182}}
  & \href{https://arxiv.org/abs/2509.05263}{\textcolor{blue}{\faExternalLink*}} \\

6 & RoboSimGS~\cite{robosimgs} & RA-L '26
  & \textcolor{orange}{\faImage}
  & 3DGS+Mesh & R\textrightarrow S\textrightarrow R
  & \textcolor{cyan}{\faCube}~\textcolor{teal}{\faAtom}
  & \textcolor{teal}{\ding{182}}
  & \href{https://robosimgs.github.io/}{\textcolor{blue}{\faExternalLink*}} \\

7 & Real-is-Sim~\cite{realissim} & ICRA '26
  & \textcolor{brown}{\faCubes}
  & 3DGS & R\textrightarrow S\textrightarrow R
  & \textcolor{teal}{\faAtom}
  & \textcolor{teal}{\ding{182}}
  & \href{https://arxiv.org/abs/2504.03597}{\textcolor{blue}{\faExternalLink*}} \\

8 & ArticFlow~\cite{lin2025articflow} & arXiv '25
  & \textcolor{orange}{\faImage}
  & Impl & R\textrightarrow S
  & \textcolor{purple}{\faProjectDiagram}
  & \textcolor{teal}{\ding{182}}
  & \href{https://arxiv.org/abs/2511.17883}{\textcolor{blue}{\faExternalLink*}} \\

9 & ArticulatedGS~\cite{guo2025articulatedgs} & CVPR '25
  & \textcolor{magenta}{\faVideo}
  & 3DGS & R\textrightarrow S
  & \textcolor{teal}{\faAtom}~\textcolor{purple}{\faProjectDiagram}
  & \textcolor{teal}{\ding{182}}
  & \href{https://arxiv.org/abs/2503.08135}{\textcolor{blue}{\faExternalLink*}} \\

10 & ArtGS~\cite{liu2025artgs} & ICLR '25
  & \textcolor{orange}{\faImage}~\textcolor{magenta}{\faVideo}
  & 3DGS & R\textrightarrow S
  & \textcolor{teal}{\faAtom}~\textcolor{purple}{\faProjectDiagram}
  & \textcolor{teal}{\ding{182}}
  & \href{https://articulate-gs.github.io/}{\textcolor{blue}{\faExternalLink*}} \\

11 & PhysTwin~\cite{jiang2025phystwin} & ICCV '25
  & \textcolor{magenta}{\faVideo}
  & 3DGS+Spring & R\textrightarrow S
  & \textcolor{teal}{\faAtom}~\textcolor{olive}{\faBalanceScale}
  & \textcolor{teal}{\ding{182}}
  & \href{https://jianghanxiao.github.io/phystwin-web/}{\textcolor{blue}{\faExternalLink*}} \\

12 & RoboScape~\cite{RoboScape} & NeurIPS '25
  & \textcolor{magenta}{\faVideo}
  & Vid+Depth & R\textrightarrow S
  & \textcolor{olive}{\faBalanceScale}
  & \textcolor{teal}{\ding{182}}
  & \href{https://arxiv.org/abs/2506.23135}{\textcolor{blue}{\faExternalLink*}} \\

13 & TwinAligner~\cite{twinaligner} & arXiv '25
  & \textcolor{brown}{\faCubes}
  & SDF+3DGS & R\textrightarrow S\textrightarrow R
  & \textcolor{teal}{\faAtom}~\textcolor{olive}{\faBalanceScale}
  & \textcolor{teal}{\ding{182}}
  & \href{https://twin-aligner.github.io/}{\textcolor{blue}{\faExternalLink*}} \\

14 & Scalable Real2Sim~\cite{scalable_robosimgs} & IROS '25
  & \textcolor{orange}{\faImage}~\textcolor{olive}{\faBalanceScale}
  & NeRF/3DGS & R\textrightarrow S
  & \textcolor{cyan}{\faCube}~\textcolor{olive}{\faBalanceScale}
  & \textcolor{teal}{\ding{182}}
  & \href{https://scalable-real2sim.github.io/}{\textcolor{blue}{\faExternalLink*}} \\

15 & REArtGS++~\cite{wu2025reartgs++} & CVPR '26
  & \textcolor{magenta}{\faVideo}
  & 3DGS & R\textrightarrow S
  & \textcolor{teal}{\faAtom}~\textcolor{purple}{\faProjectDiagram}
  & \textcolor{teal}{\ding{182}}
  & \href{https://sites.google.com/view/reartgs2/home/}{\textcolor{blue}{\faExternalLink*}} \\

\midrule

16 & RoboGSim~\cite{robogsim} & arXiv '24
  & \textcolor{orange}{\faImage}
  & 3DGS+Sim & R\textrightarrow S\textrightarrow R
  & \textcolor{red}{\faWind}~\textcolor{orange}{\faImage}
  & \textcolor{cyan}{\ding{183}}
  & \href{https://robogsim.github.io/}{\textcolor{blue}{\faExternalLink*}} \\

17 & Splat-MOVER~\cite{shorinwa2024splat} & CoRL '24
  & \textcolor{brown}{\faCubes}
  & 3DGS & R\textrightarrow S
  & \textcolor{teal}{\faAtom}
  & \textcolor{cyan}{\ding{183}}
  & \href{https://arxiv.org/abs/2405.04378}{\textcolor{blue}{\faExternalLink*}} \\

18 & Maniwhere~\cite{yuan2024learning} & CoRL '24
  & \textcolor{orange}{\faImage}~\textcolor{red}{\faWind}
  & 3DGS & S\textrightarrow R
  & \textcolor{red}{\faWind}
  & \textcolor{cyan}{\ding{183}}
  & \href{https://arxiv.org/abs/2407.15815}{\textcolor{blue}{\faExternalLink*}} \\

19 & RoboSplat~\cite{novel_demo_gs} & RSS '25
  & \textcolor{orange}{\faImage}~\textcolor{red}{\faWind}
  & 3DGS & S\textrightarrow R
  & \textcolor{orange}{\faImage}~\textcolor{red}{\faWind}
  & \textcolor{cyan}{\ding{183}}
  & \href{https://yangsizhe.github.io/robosplat/}{\textcolor{blue}{\faExternalLink*}} \\

20 & SplatSim~\cite{splatsim} & ICRA '25
  & \textcolor{orange}{\faImage}
  & 3DGS & S\textrightarrow R
  & \textcolor{orange}{\faImage}
  & \textcolor{cyan}{\ding{183}}
  & \href{https://splatsim.github.io/}{\textcolor{blue}{\faExternalLink*}} \\

21 & SIGHT~\cite{gavryushin2025sight} & arXiv '25
  & \textcolor{gray}{\faFont}~\textcolor{orange}{\faImage}
  & 3DGS & S\textrightarrow R
  & \textcolor{red}{\faWind}
  & \textcolor{cyan}{\ding{183}}
  & \href{https://arxiv.org/abs/2503.22869}{\textcolor{blue}{\faExternalLink*}} \\

22 & RoboTransfer~\cite{liu2025robotransfer} & CVPR Find. '26
  & \textcolor{magenta}{\faVideo}
  & Vid & S\textrightarrow R
  & \textcolor{magenta}{\faVideo}
  & \textcolor{cyan}{\ding{183}}
  & \href{https://arxiv.org/abs/2505.23171}{\textcolor{blue}{\faExternalLink*}} \\

23 & GAF~\cite{gaf} & ICRA '26
  & \textcolor{orange}{\faImage}
  & 3DGS & S\textrightarrow R
  & \textcolor{green!60!black}{\faFileCode}
  & \textcolor{cyan}{\ding{183}}
  & \href{https://arxiv.org/abs/2506.14135}{\textcolor{blue}{\faExternalLink*}} \\

24 & EgoDemoGen~\cite{xu2025egodemogen} & arXiv '25
  & \textcolor{orange}{\faImage}~\textcolor{red}{\faWind}
  & 3DGS & S\textrightarrow R
  & \textcolor{orange}{\faImage}~\textcolor{red}{\faWind}
  & \textcolor{cyan}{\ding{183}}
  & \href{https://arxiv.org/abs/2509.22578}{\textcolor{blue}{\faExternalLink*}} \\

25 & ExoGS~\cite{wang2026exogs} & arXiv '26
  & \textcolor{magenta}{\faVideo}
  & 3DGS & R\textrightarrow S\textrightarrow R
  & \textcolor{red}{\faWind}~\textcolor{orange}{\faImage}
  & \textcolor{cyan}{\ding{183}}
  & \href{https://arxiv.org/abs/2601.18629}{\textcolor{blue}{\faExternalLink*}} \\

\midrule

26 & MimicGen~\cite{mandlekar2023mimicgen} & CoRL '23
  & \textcolor{red}{\faWind}
  & Sim & S\textrightarrow R
  & \textcolor{red}{\faWind}
  & \textcolor{red}{\ding{184}}
  & \href{https://mimicgen.github.io/}{\textcolor{blue}{\faExternalLink*}} \\

27 & Gen2Sim~\cite{gen2sim} & ICRA '24
  & \textcolor{orange}{\faImage}
  & Sim+LLM & S\textrightarrow R
  & \textcolor{red}{\faWind}
  & \textcolor{red}{\ding{184}}
  & \href{https://gen2sim.github.io/}{\textcolor{blue}{\faExternalLink*}} \\

28 & GenSim2~\cite{gensim2} & CoRL '24
  & \textcolor{orange}{\faImage}~\textcolor{gray}{\faFont}
  & Sim+LLM & S\textrightarrow R
  & \textcolor{red}{\faWind}
  & \textcolor{red}{\ding{184}}
  & \href{https://gensim2.github.io/}{\textcolor{blue}{\faExternalLink*}} \\

29 & DemoGen~\cite{demogen} & RSS '25
  & \textcolor{blue}{\faBraille}~\textcolor{red}{\faWind}
  & PC & S\textrightarrow R
  & \textcolor{red}{\faWind}
  & \textcolor{red}{\ding{184}}
  & \href{https://demo-generation.github.io/}{\textcolor{blue}{\faExternalLink*}} \\

30 & R2R2R~\cite{r2r2r} & CoRL '25
  & \textcolor{magenta}{\faVideo}
  & 3DGS\textrightarrow Mesh & R\textrightarrow S\textrightarrow R
  & \textcolor{red}{\faWind}
  & \textcolor{red}{\ding{184}}
  & \href{https://arxiv.org/abs/2505.09601}{\textcolor{blue}{\faExternalLink*}} \\

31 & ManipDreamer3D~\cite{manipdreamer3d} & AAAI '26
  & \textcolor{orange}{\faImage}
  & Occ+Vid & S\textrightarrow R
  & \textcolor{red}{\faWind}~\textcolor{magenta}{\faVideo}
  & \textcolor{red}{\ding{184}}
  & \href{https://arxiv.org/abs/2509.05314}{\textcolor{blue}{\faExternalLink*}} \\

32 & DreamGen~\cite{dreamgen} & CoRL '25
  & \textcolor{orange}{\faImage}
  & Vid & S\textrightarrow R
  & \textcolor{red}{\faWind}~\textcolor{magenta}{\faVideo}
  & \textcolor{red}{\ding{184}}
  & \href{https://research.nvidia.com/labs/gear/dreamgen/}{\textcolor{blue}{\faExternalLink*}} \\

33 & AnchorDream~\cite{anchordream} & ICRA '26
  & \textcolor{orange}{\faImage}
  & Vid & S\textrightarrow R
  & \textcolor{magenta}{\faVideo}
  & \textcolor{red}{\ding{184}}
  & \href{https://junjieye.com/AnchorDream/}{\textcolor{blue}{\faExternalLink*}} \\

34 & PhysWorld~\cite{physworld} & arXiv '25
  & \textcolor{orange}{\faImage}~\textcolor{gray}{\faFont}
  & Vid+3D & S\textrightarrow R
  & \textcolor{green!60!black}{\faFileCode}
  & \textcolor{red}{\ding{184}}
  & \href{https://jiageng.me/PhysWorld_Web/}{\textcolor{blue}{\faExternalLink*}} \\

35 & DRAW2ACT~\cite{draw2act} & ICRA '26
  & \textcolor{orange}{\faImage}~\textcolor{red}{\faWind}
  & Vid+Depth & S\textrightarrow R
  & \textcolor{magenta}{\faVideo}
  & \textcolor{red}{\ding{184}}
  & \href{https://arxiv.org/abs/2512.14217}{\textcolor{blue}{\faExternalLink*}} \\

36 & Video2Act~\cite{video2act} & arXiv '25
  & \textcolor{magenta}{\faVideo}
  & Vid & S\textrightarrow R
  & \textcolor{green!60!black}{\faFileCode}
  & \textcolor{red}{\ding{184}}
  & \href{https://arxiv.org/abs/2512.03044}{\textcolor{blue}{\faExternalLink*}} \\

37 & GWM~\cite{gwm} & ICCV '25
  & \textcolor{orange}{\faImage}
  & 3DGS+DiT & S\textrightarrow R
  & \textcolor{green!60!black}{\faFileCode}
  & \textcolor{red}{\ding{184}}
  & \href{https://gaussian-world-model.github.io/}{\textcolor{blue}{\faExternalLink*}} \\

38 & GraspVLA~\cite{graspvla} & CoRL '25
  & \textcolor{cyan}{\faCube}
  & Sim & S\textrightarrow R
  & \textcolor{red}{\faWind}
  & \textcolor{red}{\ding{184}}
  & \href{https://pku-epic.github.io/GraspVLA-web}{\textcolor{blue}{\faExternalLink*}} \\

39 & AOMGen~\cite{wu2025aomgen} & CVPR Find. '26
  & \textcolor{orange}{\faImage}
  & Diff & S\textrightarrow R
  & \textcolor{red}{\faWind}~\textcolor{orange}{\faImage}
  & \textcolor{red}{\ding{184}}
  & \href{https://arxiv.org/abs/2512.18396}{\textcolor{blue}{\faExternalLink*}} \\

\bottomrule
\end{tabular}}
}
\vspace{-5pt}
\end{table*}

\subsubsection{Physics Perception and Visual-Dynamics Alignment}
Visual fidelity and correct kinematics alone are insufficient for closing the sim-to-real gap; physical parameters such as mass distribution, surface friction, elasticity, and restitution coefficients must also be inferred from observations and injected into the reconstructed representations.

For deformable bodies, PhysTwin~\cite{jiang2025phystwin} combines a spring-mass dynamics model, generative shape priors, and 3DGS rendering in a hybrid representation, using hierarchical sparse-to-dense optimization (zero-order for collision bounds, first-order for spring stiffness) to extract material parameters from sparse video without differentiable simulation. For rigid and contact-rich settings, RoboScape~\cite{RoboScape} jointly learns RGB video generation, temporal depth prediction, and keypoint tracking, embedding physical constraints that encode hidden dynamics and reduce the spatial hallucinations common in unconstrained video diffusion. The Gaussian World Model (GWM)~\cite{gwm} takes a complementary approach by using 3DGS as a representation for model-based reinforcement learning: a Gaussian VAE downsamples scenes into a compact latent space, and a latent Diffusion Transformer predicts future Gaussian states conditioned on robot actions, enabling real-time state prediction for planning.

For full Real-to-Sim-to-Real transfer, TwinAligner~\cite{twinaligner} decouples collision geometry (an SDF mesh for rigid-body contact) from visual rendering (overlaid 3DGS primitives), avoiding the optimization instability of end-to-end differentiable simulation. It aligns simulated dynamics to real-world trajectories via a gradient-free optimizer that matches velocity, inertia, and center-of-mass parameters, producing physics-aware digital twins that support zero-shot policy deployment on physical hardware. Real-is-Sim~\cite{realissim} pushes the digital twin toward real-time closed-loop operation: it maintains a 60\,Hz synchronized Embodied Gaussian twin so that policies are trained and executed entirely within the simulated counterpart while continuously aligning with the physical environment.

\subsection{Generative Data Augmentation}
\label{sec:data_aug}
Data augmentation for robotic manipulation has shifted from conventional 2D image-space editing toward explicit 3D and 4D spatiotemporal augmentation. The core motivation is scaling from \textit{few} manually collected demonstrations to \textit{many} synthetically generated, physically grounded samples. Early generative augmentation methods demonstrated this potential in 2D: ROSIE~\cite{yu2023rosie} used text-to-image models to paint semantically meaningful distractors and backgrounds into existing robot data, while GenAug~\cite{chen2023genaug} applied generative models to retarget manipulation behaviors to unseen visual contexts. However, these 2D augmentations do not respect multi-view geometric consistency, leading to generalization failures when policies encounter novel viewpoints or object configurations. By leveraging NeRF and 3DGS representations, recent methods synthesize augmented observations that are geometrically consistent across views and physically plausible across time.

\subsubsection{Viewpoint and Geometric Augmentation}
The most direct augmentation strategy uses 3D representations to synthesize novel-viewpoint observations from limited demonstrations. RoboSplat~\cite{novel_demo_gs} reconstructs demonstrations as 3DGS scenes and renders multi-view pseudo-observations under diverse camera poses, supporting content replacement and equivariant transformations. AOMGen~\cite{wu2025aomgen} extends this to articulated objects, generating multi-view RGB trajectories that respect underlying joint states. SplatSim~\cite{splatsim} replaces polygonal meshes with Gaussian splats within simulators, rendering photorealistic observations that reduce the visual domain gap and enable zero-shot sim-to-real transfer.

\subsubsection{Dynamic and Temporal Augmentation}
Beyond static viewpoint synthesis, recent work augments the temporal dimension by generating trajectories with varied execution dynamics. ExoGS~\cite{wang2026exogs} reconstructs real-world manipulation sequences into editable 3DGS assets with temporal dynamics, and incorporates a Mask Adapter that injects instance-level semantic cues to improve policy robustness under visual domain shifts. GAF~\cite{gaf} extends 3D Gaussians with explicit motion attributes to jointly model dynamic scene evolution and action prediction, using point-cloud-based matching to provide multi-view guidance and initial action hypotheses for downstream policy learning.

\begin{figure*}[t]
    \centering
    \includegraphics[width=0.95\linewidth]{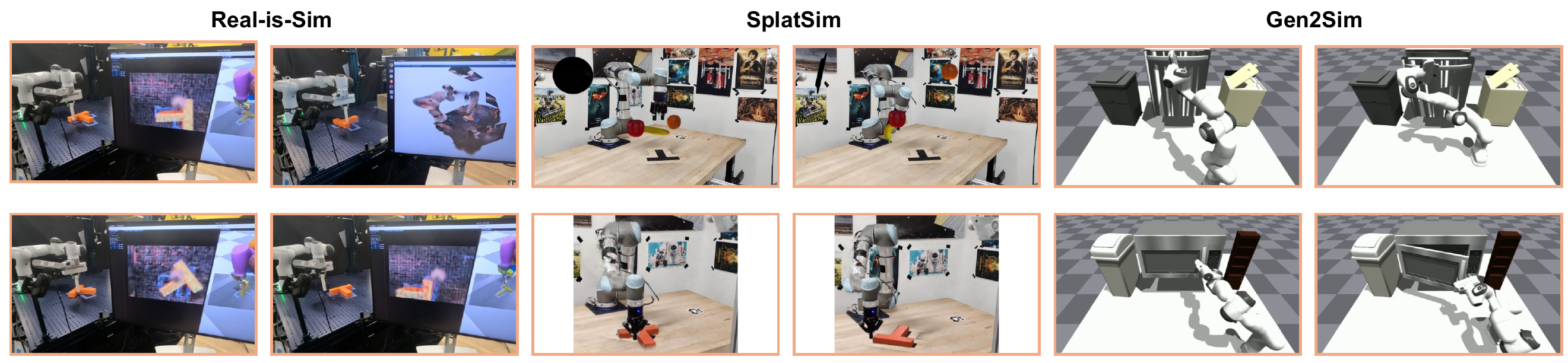}
    \caption{\textbf{Representative examples across the Sim2Real Bridge.} From left to right, \textsc{Real-is-Sim} shows synchronized digital twins, \textsc{SplatSim} shows 3DGS-based simulation and data augmentation, and \textsc{Gen2Sim} shows generated articulated-object tasks for policy learning in simulation.}
    \label{fig:sim2real-example}
\end{figure*}

\subsubsection{Physics-Aware Consistency}
As viewpoint and dynamic augmentation scale up, a central challenge is preserving the underlying physical and geometric consistency of the synthesized data. Naive transformations in image or latent space can introduce non-physical artifacts---object interpenetrations, broken spatial relationships, or inconsistent depth---that degrade policy robustness upon real-world deployment. Recent methods address this by enforcing explicit 3D structural constraints throughout the augmentation pipeline.

On the video generation side, RoboTransfer~\cite{liu2025robotransfer} conditions video diffusion models on global depth maps and surface normals to enforce geometric coherence across synthesized frames, mitigating the spatial hallucinations that arise in unconstrained video synthesis. For egocentric settings, EgoDemoGen~\cite{xu2025egodemogen} introduces a self-supervised double reprojection strategy that synthesizes viewpoint-robust first-person observations conditioned on kinematically retargeted actions, ensuring that the augmented views remain consistent with the underlying 3D scene geometry.

On the scene editing side, Splat-MOVER~\cite{shorinwa2024splat} leverages the explicit and semantically editable nature of 3DGS to maintain a geometrically consistent digital twin during multi-stage, open-vocabulary manipulation, enabling object rearrangement without breaking spatial coherence. Maniwhere~\cite{yuan2024learning} integrates spatial transformer networks for multi-view feature alignment, ensuring that augmented demonstrations across different viewpoints share a consistent 3D structure. SIGHT~\cite{gavryushin2025sight} further enforces physics-aware constraints by grounding synthesized trajectories in explicit 3D spatial representations, ensuring that generated action sequences respect physical boundaries and contact geometry. Together, these methods represent a shift from pixel-level perturbations toward geometry-grounded augmentation, where consistency with the underlying 3D structure is treated as a hard constraint rather than an emergent property, thereby improving the reliability of augmented data for downstream policy learning.

\subsection{Generative Task and Demonstration}
\label{sec:gen_task_demo}
Beyond augmenting existing observations, a complementary direction generates entirely new demonstrations and task-relevant trajectories at scale. By decoupling 3D asset generation from robotic kinematics, these methods synthesize large corpora of structured demonstration data, reducing the dependence on costly human teleoperation.

\subsubsection{Demonstration Scaling via Simulation and Trajectory Synthesis}
The most direct approach adapts a small set of human demonstrations to novel environments through spatial transformations and motion planning. MimicGen~\cite{mandlekar2023mimicgen} spatially transforms object-centric action segments from a few demonstrations to synthesize tens of thousands of trajectories across varying scene configurations. DemoGen~\cite{demogen} adapts single-source demonstrations to new object configurations using 3D point cloud editing and collision-free motion planning.

A complementary strategy uses LLMs to automatically construct complete simulation tasks from visual inputs. Gen2Sim~\cite{gen2sim} lifts open-world images to 3D assets, then employs LLMs to assign physics parameters and generate task decompositions with reward functions for RL training. GenSim2~\cite{gensim2} extends this direction with multimodal reasoning LLMs to create up to 100 articulated-object tasks with 200 objects, training a multi-task policy that transfers zero-shot to real robots. R2R2R~\cite{r2r2r} extracts 6-DoF object motion from a single video, reconstructs assets as 3DGS, and converts them to meshes for compatibility with parallelized rendering engines such as IsaacLab. ManipDreamer3D~\cite{manipdreamer3d} reconstructs a 3D occupancy map from a single image and uses a CHOMP-inspired planner to synthesize collision-free trajectories that subsequently guide a video diffusion process.

\subsubsection{World Model-Based Demonstration Generation}
\label{sec:world_model_demo}
Standard video diffusion models can produce visually diverse rollouts but frequently hallucinate physically impossible motions. World model-based methods address this by embedding structural and kinematic constraints into the generation pipeline. DreamGen~\cite{dreamgen} adapts image-to-video models to target robotic embodiments, generating photorealistic rollouts and recovering pseudo-action sequences via an Inverse Dynamics Model. AnchorDream~\cite{anchordream} conditions diffusion directly on rendered robotic motions, anchoring generation to the robot's kinematic structure to prevent spatial hallucinations.

PhysWorld~\cite{physworld} couples task-conditioned video generation with physical world reconstruction, grounding synthesized motions into executable actions through object-centric residual reinforcement learning. DRAW2ACT~\cite{draw2act} uses depth-aware, trajectory-conditioned video generation, extracting depth, semantic, and shape representations to ensure end-effector--object consistency. Video2Act~\cite{video2act} translates video diffusion outputs into real-time manipulation commands via an asynchronous dual-system architecture that pairs a slow video perception module with a fast Diffusion Transformer execution head. These methods collectively reduce dependence on costly human teleoperation for demonstration collection.

\subsubsection{Foundation-Scale Synthetic Training}
\label{sec:foundation_scale}
The approaches above converge toward scaling synthetic action data to foundation-model levels. GraspVLA~\cite{graspvla} curates SynGrasp-1B, a billion-frame grasping dataset generated in simulation with photorealistic rendering and extensive domain randomization. To leverage this synthetic corpus alongside uncurated real-world video, GraspVLA introduces Progressive Action Generation (PAG), which integrates autoregressive visual perception with flow-matching-based action generation in a unified architecture. Co-training on internet-scale semantic data (2D bounding boxes) and synthetic 6-DoF grasp poses enables the resulting model to transfer grasping skills to novel object categories, demonstrating that simulation-scale 3D generation can serve as prerequisite infrastructure for training open-vocabulary embodied foundation models.

%% file: sec/datasets.tex
\begin{figure*}[!t]
    \centering
    \includegraphics[width=0.95\linewidth]{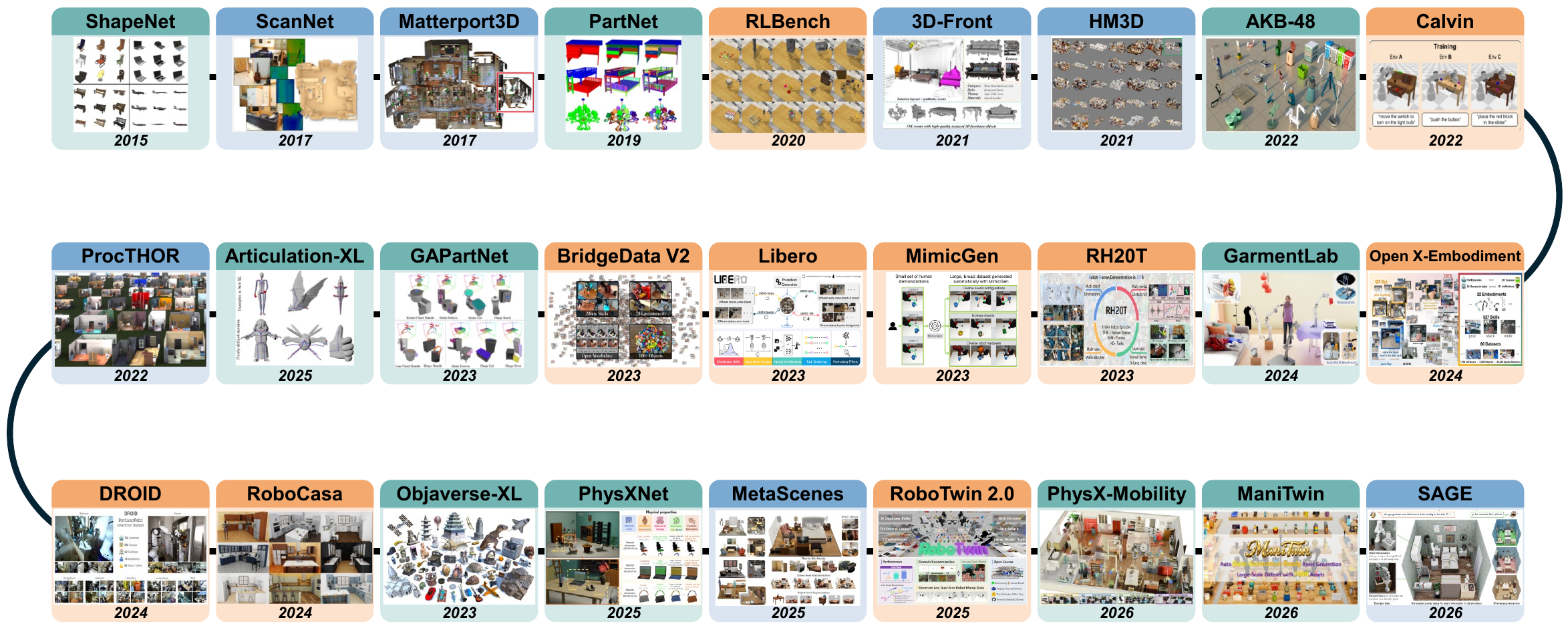}
    \caption{\textbf{Timeline of major datasets for 3D generation in embodied AI.} Datasets are organized chronologically and categorized by asset type, including object-level datasets, scene datasets, and embodied manipulation benchmarks.}
    \label{fig:datasets_timeline}
\end{figure*}

\section{Datasets and Evaluation}
\label{sec:dataset}

\subsection{Datasets}
\label{subsec:dataset}

Progress in 3D asset generation for embodied AI depends critically on datasets that provide ground-truth geometry, physical annotations, and interaction semantics. Unlike conventional 3D generation benchmarks that assess reconstruction accuracy or perceptual quality in isolation, embodied applications require datasets jointly capturing geometric structure, physical attributes, and simulator compatibility. This section surveys the datasets underpinning the three main research directions in this survey, organized by the nature of the assets rather than by chapter.

\subsubsection{Object Asset Datasets}
\label{sec:dataset_object}

\begin{table*}[!t]
\caption{Summary of datasets and benchmarks for 3D object generation in embodied AI.}
\label{tab:datasets}
\vspace{-5 pt}
\begin{minipage}{\textwidth}
    \footnotesize
    \begin{itemize}[leftmargin=1.2em, nosep, itemsep=2pt, label=\small$\bullet$]
    \item \textbf{Phys.}: physical property annotations (mass, friction, material, etc.).
    \item \textbf{Kin.}: kinematic/joint annotations.
    \item \textbf{Sim.}: direct simulator compatibility (URDF/MJCF export or native sim support).
    \item \cmark~=~provided, \pmark~=~partial, \xmark~=~not provided.
    \end{itemize}
\end{minipage}
\rowcolors{2}{gray!20}{white}
\resizebox{\textwidth}{!}{%
\begin{tabular}{@{}c l l l l l c c c c@{}}
\toprule
\textbf{\#} & \textbf{Dataset} & \textbf{Venue} & \textbf{Type}
  & \textbf{Scale} & \textbf{Annotations}
  & \textbf{Phys.} & \textbf{Kin.} & \textbf{Sim.}
  & \textbf{URL} \\
\midrule

1 & ShapeNet~\cite{chang2015shapenet} & arXiv '15
  & CAD Models & 51K objects / 55 cat.
  & Geometry, Semantics
  & \xmark & \xmark & \xmark
  & \href{https://shapenet.org/}{\textcolor{blue}{\faExternalLink*}} \\

2 & PartNet~\cite{mo2019partnet} & CVPR '19
  & CAD Models & 26.7K objects / 24 cat.
  & Part Seg. (573K instances)
  & \xmark & \xmark & \xmark
  & \href{https://partnet.cs.stanford.edu/}{\textcolor{blue}{\faExternalLink*}} \\

3 & PartNet-Mobility~\cite{xiang2020sapien} & CVPR '20
  & CAD Models & 2,346 objects / 46 cat.
  & Part Seg., Joint Params.
  & \xmark & \cmark & \cmark
  & \href{https://sapien.ucsd.edu/browse}{\textcolor{blue}{\faExternalLink*}} \\

4 & 3D-FUTURE~\cite{3d-future} & IJCV '21
  & CAD Models & 13.1K objects / 43 cat.
  & Geometry, Semantics, Texture
  & \xmark & \xmark & \xmark
  & \href{https://tianchi.aliyun.com/dataset/98063}{\textcolor{blue}{\faExternalLink*}} \\

5 & GSO~\cite{downs2022google} & ICRA '22
  & Real Scans & 1,030 objects
  & Geometry, PBR Material
  & \xmark & \xmark & \cmark
  & \href{https://huggingface.co/datasets/gso-bench/gso}{\textcolor{blue}{\faExternalLink*}} \\

6 & ABO~\cite{collins2022abo} & CVPR '22
  & Artist-designed & 7,953 objects / 63 cat.
  & Geometry, PBR Material, Real Dims.
  & \xmark & \xmark & \pmark
  & \href{https://amazon-berkeley-objects.s3.amazonaws.com/index.html}{\textcolor{blue}{\faExternalLink*}} \\

7 & AKB-48~\cite{liu2022akb} & CVPR '22
  & Real Scans & 2,037 objects / 48 cat.
  & Part Seg., Joint, Mass, Inertia, Friction
  & \cmark & \cmark & \cmark
  & \href{https://liuliu66.github.io/articulationobjects/}{\textcolor{blue}{\faExternalLink*}} \\

8 & Objaverse~\cite{deitke2023objaverse} & CVPR '23
  & Mixed & 800K+ objects
  & Geometry, Captions, Tags, Animations
  & \xmark & \xmark & \xmark
  & \href{https://huggingface.co/datasets/allenai/objaverse}{\textcolor{blue}{\faExternalLink*}} \\

9 & GAPartNet~\cite{geng2023gapartnet} & CVPR '23
  & CAD Models & 1,166 objects / 27 cat.
  & Part Seg., Poses (8,489 instances)
  & \xmark & \cmark & \cmark
  & \href{https://pku-epic.github.io/GAPartNet/}{\textcolor{blue}{\faExternalLink*}} \\

10 & ClothesNet~\cite{zhou2023clothesnet} & ICCV '23
  & CAD Models & 4,400+ garments / 11 cat.
  & Geometry, Boundary Lines, Keypoints
  & \xmark & \xmark & \pmark
  & \href{https://sites.google.com/view/clothesnet}{\textcolor{blue}{\faExternalLink*}} \\

11 & Objaverse-XL~\cite{deitke2023objaversexl} & NeurIPS '23
  & Mixed & 10.2M+ objects
  & Geometry, Texture, Tags
  & \xmark & \xmark & \xmark
  & \href{https://objaverse.allenai.org/}{\textcolor{blue}{\faExternalLink*}} \\

12 & GarmentLab~\cite{lu2024garmentlab} & NeurIPS '24
  & Sim. Env. & 9K+ objects / 11 garment cat.
  & Physics (PBD/FEM), Task Anno.
  & \cmark & \xmark & \cmark
  & \href{https://garmentlab.github.io/}{\textcolor{blue}{\faExternalLink*}} \\

13 & Articulation-XL~\cite{song2025magicarticulate} & CVPR '25
  & Mixed & 48K+ objects
  & Skeleton, Skinning Weights
  & \xmark & \cmark & \pmark
  & \href{https://huggingface.co/datasets/Seed3D/Articulation-XL2.0}{\textcolor{blue}{\faExternalLink*}} \\

14 & PhysXNet~\cite{cao2025physx3d} & NeurIPS '25
  & CAD Models & 26K objects
  & Joint, Mass, Material, Friction
  & \cmark & \cmark & \cmark
  & \href{https://huggingface.co/datasets/Caoza/PhysX-3D}{\textcolor{blue}{\faExternalLink*}} \\

15 & PhysX-Mobility~\cite{cao2025physx} & CVPR '26
  & CAD Models & 2K+ objects / 47 cat.
  & Geometry, Kinematics, Physics (per-part)
  & \cmark & \cmark & \cmark
  & \href{https://huggingface.co/datasets/Caoza/PhysX-Mobility}{\textcolor{blue}{\faExternalLink*}} \\

16 & DTC~\cite{dtc} & CVPR '25
  & Real Scans & 2K objects / 40 cat.
  & Geometry, 4K PBR, Evaluation Seqs.
  & \xmark & \xmark & \cmark
  & \href{https://www.projectaria.com/datasets/dtc/}{\textcolor{blue}{\faExternalLink*}} \\

17 & ManiTwin~\cite{manitwin} & arXiv '26
  & Mixed & 100K+ objects
  & Geometry, Manipulation Anno., Sim-ready
  & \pmark & \pmark & \cmark
  & \href{https://manitwin.github.io/}{\textcolor{blue}{\faExternalLink*}} \\
  
\bottomrule
\end{tabular}%
}
\end{table*}

As Table~\ref{tab:datasets} summarizes, object asset datasets have evolved along two axes: geometric scale and physical annotation depth. Early datasets such as ShapeNet~\cite{chang2015shapenet} and PartNet~\cite{mo2019partnet} provide the geometric foundation but lack physical or kinematic parameters. The transition toward simulation-ready datasets is marked by kinematic annotations in PartNet-Mobility~\cite{xiang2020sapien} and AKB-48~\cite{liu2022akb}, and by large-scale pretraining corpora such as Objaverse~\cite{deitke2023objaverse} and Objaverse-XL~\cite{deitke2023objaversexl} that trade physical annotations for geometric diversity. Most recently, PhysXNet~\cite{cao2025physx3d}, PhysX-Mobility~\cite{cao2025physx}, and Articulation-XL~\cite{song2025magicarticulate} address the critical gap in physical annotations, while DTC~\cite{dtc} and ManiTwin~\cite{manitwin} contribute photorealistic quality and manipulation-relevant metadata, respectively. This progression reflects a broader shift from assets designed for perception benchmarks to assets intended for direct use in simulation and manipulation pipelines. For deformable objects, ClothesNet~\cite{zhou2023clothesnet} and GarmentLab~\cite{lu2024garmentlab} provide garment geometry with simulation backends.

\subsubsection{Scene Datasets}
\label{sec:dataset_scene}

\begin{table*}[tbp]
\caption{Summary of scene-level datasets for embodied scene generation.}
\label{tab:scene_datasets}
\vspace{-5 pt}
\begin{minipage}{\textwidth}
    \footnotesize
    \begin{itemize}[leftmargin=1.2em, nosep, itemsep=2pt, label=\small$\bullet$]
    \item \textbf{Type}: synthetic scenes, real scans/reconstructions, scene graphs, procedural scenes, agent-generated scenes, auto-constructed scenes, task-oriented scenes, and domain-specific scenes.
    \item \textbf{Scene Info}: major scene-level information, such as layout, semantics, relations, task descriptions, or simulator-oriented metadata.
    \item \textbf{Sim}: direct simulator compatibility or embodied-oriented usability. \cmark~=~provided, \pmark~=~partial, \xmark~=~not provided.
    \item \textbf{Cat}: \textcolor{blue}{\ding{182}}~Source Scene, \textcolor{green!60!black}{\ding{183}}~Generated Scene, \textcolor{red}{\ding{184}}~Domain-oriented Scene.
    \end{itemize}
\end{minipage}

\rowcolors{2}{gray!20}{white}
\resizebox{\textwidth}{!}{%
\begin{tabular}{@{}c l c l l l c c c@{}}
\toprule
\textbf{\#} & \textbf{Dataset} & \textbf{Venue} & \textbf{Type}
  & \textbf{Scale} & \textbf{Scene Info}
  & \textbf{Sim.} & \textbf{Cat} & \textbf{URL} \\
\midrule
1 & Matterport3D~\cite{matterport3d}
  & 3DV '17
  & Real Scans
  & 90 scenes / 10.8K pans / 194K RGB-D
  & Geometry, Poses, Recon., 2D/3D Sem. Seg.
  & \pmark
  & \textcolor{blue}{\ding{182}}
  & \href{https://niessner.github.io/Matterport/}{\textcolor{blue}{\faExternalLink*}} \\

2 & ScanNet~\cite{scannet}
  & CVPR '17
  & Real Scans
  & 1.5K scans / 2.5M views
  & Geometry, Poses, Recon., Sem. Seg.
  & \pmark
  & \textcolor{blue}{\ding{182}}
  & \href{http://www.scan-net.org/}{\textcolor{blue}{\faExternalLink*}} \\

3 & Replica~\cite{replica}
  & arXiv '19
  & Real Reconstructions
  & 18 scenes
  & Geometry, HDR Texture, Sem./Inst. Labels
  & \cmark
  & \textcolor{blue}{\ding{182}}
  & \href{https://github.com/facebookresearch/Replica-Dataset}{\textcolor{blue}{\faExternalLink*}} \\

4 & 3DSSG~\cite{3dssg}
  & CVPR '20
  & Scene Graphs
  & 1.5K SGs / 48K objs / 544K rels
  & Semantics, Attributes, Relations
  & \xmark
  & \textcolor{blue}{\ding{182}}
  & \href{https://3dssg.github.io/}{\textcolor{blue}{\faExternalLink*}} \\

5 & 3D-FRONT~\cite{3d-front}
  & ICCV '21
  & Synthetic Scenes
  & 18.9K rooms / 6.8K houses
  & Layout, Semantics, Texture
  & \xmark
  & \textcolor{blue}{\ding{182}}
  & \href{https://tianchi.aliyun.com/specials/promotion/alibaba-3d-scene-dataset}{\textcolor{blue}{\faExternalLink*}} \\

6 & HM3D~\cite{HM3D}
  & NeurIPS D\&B '21
  & Real Scans
  & 1K scans
  & Geometry, Texture, Recon.
  & \cmark
  & \textcolor{blue}{\ding{182}}
  & \href{https://aihabitat.org/datasets/hm3d/}{\textcolor{blue}{\faExternalLink*}} \\

\midrule

7 & ProcTHOR-10K~\cite{procthor}
  & NeurIPS '22
  & Procedural Scenes
  & 10K houses
  & Layout, Interactive Objects
  & \cmark
  & \textcolor{green!60!black}{\ding{183}}
  & \href{https://procthor.allenai.org/}{\textcolor{blue}{\faExternalLink*}} \\

8 & MetaScenes~\cite{metascenes}
  & CVPR '25
  & Auto-constructed
  & 706 scenes / 15.4K objs
  & Geometry, Texture, Replacements, Phys. Plaus.
  & \cmark
  & \textcolor{green!60!black}{\ding{183}}
  & \href{https://meta-scenes.github.io/}{\textcolor{blue}{\faExternalLink*}} \\
9 & SAGE-10K~\cite{sage}
  & CVPR '26
  & Agent-generated
  & 10K scenes / 565K objs
  & Layout, Generated Objects, Sim-ready
  & \cmark
  & \textcolor{green!60!black}{\ding{183}}
  & \href{https://nvlabs.github.io/sage/}{\textcolor{blue}{\faExternalLink*}} \\

  \midrule

10 & MesaTask-10K~\cite{mesatask}
  & NeurIPS '25
  & Task-oriented Scenes
  & 10.7K tabletop scenes
  & Tasks, Layouts, Relations
  & \pmark
  & \textcolor{red}{\ding{184}}
  & \href{https://github.com/InternRobotics/MesaTask}{\textcolor{blue}{\faExternalLink*}} \\

11 & MarketGen~\cite{marketgen}
  & arXiv '25
  & Domain-specific Scenes
  & Supermarket envs / 1.1K+ assets
  & Layout Rules, Conditions, Benchmarks
  & \cmark
  & \textcolor{red}{\ding{184}}
  & \href{https://arxiv.org/abs/2511.21161}{\textcolor{blue}{\faExternalLink*}} \\

\bottomrule
\end{tabular}%
}
\end{table*}

Scene datasets (Table~\ref{tab:scene_datasets}) have progressed from passive collection to active construction. Source datasets such as 3D-FRONT~\cite{3d-front} and ScanNet~\cite{scannet} provide layout and semantic priors, while real-scan datasets (Matterport3D~\cite{matterport3d}, Replica~\cite{replica}, HM3D~\cite{HM3D}) serve as scene priors and simulation environments. Generated datasets---ProcTHOR-10K~\cite{procthor}, SAGE-10K~\cite{sage}, MetaScenes~\cite{metascenes}---emphasize scalability and simulator readiness over static collection. Their evolution also shows that scene datasets are no longer only passive resources, but increasingly function as controllable generators of embodied training environments. Domain-oriented datasets bind scene construction to downstream tasks: MesaTask-10K~\cite{mesatask} maps task descriptions to tabletop configurations, and MarketGen~\cite{marketgen} provides supermarket environments with benchmarks.

\subsubsection{Robot Demonstration Datasets}
\label{sec:dataset_demo}

\begin{table*}[tbp]
\caption{Summary of robot demonstration datasets relevant to 3D asset generation for embodied AI.}
\label{tab:demo_datasets}
\vspace{-5 pt}
\begin{minipage}{\textwidth}
    \footnotesize
    \begin{itemize}[leftmargin=1.2em, nosep, itemsep=2pt, label=\small$\bullet$]
    \item \textbf{Source}: Sim = simulation-only, Real = real-world-only, Both = paired real and simulated data.
    \item \textbf{Cat}: \textcolor{blue}{\ding{182}}~Real-World Corpus, \textcolor{green!60!black}{\ding{183}}~Sim-Based Benchmark, \textcolor{red}{\ding{184}}~Augmentation / Scaling System.
    \end{itemize}
\end{minipage}

\rowcolors{2}{gray!20}{white}
\resizebox{\textwidth}{!}{%
\begin{tabular}{@{}c l l l l l l c c@{}}
\toprule
\textbf{\#} & \textbf{Dataset} & \textbf{Venue} & \textbf{Source}
  & \textbf{Robot / Embodiment} & \textbf{Scale}
  & \textbf{Key Features}
  & \textbf{Cat} & \textbf{URL} \\
\midrule

1 & Open X-Embodiment~\cite{openxembodiment2024}
  & ICRA'24
  & Real
  & 22 embodiments (60 datasets)
  & 1M+ trajectories / 527 skills
  & Multi-embodiment aggregation, cross-robot transfer
  & \textcolor{blue}{\ding{182}}
  & \href{https://robotics-transformer-x.github.io/}{\textcolor{blue}{\faExternalLink*}} \\

2 & DROID~\cite{khazatsky2024droid}
  & RSS'24
  & Real
  & Franka Panda (18 robots)
  & 76K trajectories / 86 tasks
  & In-the-wild collection, 564 scenes, stereo RGB
  & \textcolor{blue}{\ding{182}}
  & \href{https://droid-dataset.github.io/}{\textcolor{blue}{\faExternalLink*}} \\

3 & RH20T~\cite{fang2024rh20t}
  & ICRA'24
  & Real
  & 4 robots / 7 configurations
  & 110K+ sequences / 147 tasks
  & Multimodal (RGB-D, F/T, audio), human-robot paired
  & \textcolor{blue}{\ding{182}}
  & \href{https://rh20t.github.io/}{\textcolor{blue}{\faExternalLink*}} \\

4 & BridgeData V2~\cite{walke2023bridgedata}
  & CoRL'23
  & Real
  & WidowX 250
  & 60K trajectories / 24 envs / 13 skills
  & Low-cost hardware, language annotations
  & \textcolor{blue}{\ding{182}}
  & \href{https://rail-berkeley.github.io/bridgedata/}{\textcolor{blue}{\faExternalLink*}} \\
  
\midrule

5 & RoboTwin~\cite{mu2024robotwin}
  & CVPR '25
  & Both
  & ALOHA dual-arm
  & 50 bimanual tasks / 731 objects
  & 3D generative digital twins, SAPIEN sim
  & \textcolor{green!60!black}{\ding{183}}
  & \href{https://arxiv.org/abs/2504.13059}{\textcolor{blue}{\faExternalLink*}} \\

6 & RoboTwin 2.0~\cite{chen2025robotwin}
  & ICML '26
  & Both
  & 5 bimanual embodiments
  & 100K+ trajectories / 50 tasks
  & Domain randomization, MLLM task generation
  & \textcolor{green!60!black}{\ding{183}}
  & \href{https://robotwin-platform.github.io/}{\textcolor{blue}{\faExternalLink*}} \\

7 & RoboCasa~\cite{nasiriany2024robocasa}
  & RSS'24
  & Sim
  & Franka (robosuite/MuJoCo)
  & 100K+ trajectories / 100 tasks
  & AI-generated textures \& objects, kitchen domain
  & \textcolor{green!60!black}{\ding{183}}
  & \href{https://robocasa.ai/}{\textcolor{blue}{\faExternalLink*}} \\

8 & ManiSkill2/3~\cite{tao2024maniskill3}
  & ICLR'23
  & Sim
  & Multi-robot (SAPIEN)
  & 20+ task families / 2K+ objects
  & GPU-parallelized, rigid \& soft body
  & \textcolor{green!60!black}{\ding{183}}
  & \href{https://maniskill.ai/}{\textcolor{blue}{\faExternalLink*}} \\

9 & RLBench~\cite{james2020rlbench}
  & RA-L'20
  & Sim
  & Franka Panda (CoppeliaSim)
  & 100 tasks / unlimited demos
  & Motion-planner demos, multi-view RGB-D
  & \textcolor{green!60!black}{\ding{183}}
  & \href{https://sites.google.com/view/rlbench}{\textcolor{blue}{\faExternalLink*}} \\

10 & LIBERO~\cite{liu2023libero}
  & NeurIPS '23
  & Sim
  & Franka Panda (robosuite)
  & 130 tasks / 6.5K demos
  & Lifelong learning, controlled distribution shifts
  & \textcolor{green!60!black}{\ding{183}}
  & \href{https://libero-project.github.io/}{\textcolor{blue}{\faExternalLink*}} \\

11 & CALVIN~\cite{mees2022calvin}
  & RA-L'22
  & Sim
  & Franka Panda (PyBullet)
  & 34 tasks / 24h play data
  & Language-conditioned, long-horizon chains
  & \textcolor{green!60!black}{\ding{183}}
  & \href{https://github.com/mees/calvin}{\textcolor{blue}{\faExternalLink*}} \\
  
\midrule

12 & MimicGen~\cite{mandlekar2023mimicgen}
  & CoRL'23
  & Sim
  & Franka Panda (robosuite)
  & 50K+ demos from $\sim$200 seeds
  & Trajectory augmentation, spatial adaptation
  & \textcolor{red}{\ding{184}}
  & \href{https://mimicgen.github.io/}{\textcolor{blue}{\faExternalLink*}} \\

\bottomrule
\end{tabular}%
}
\end{table*}

Robot demonstration datasets (Table~\ref{tab:demo_datasets}) supply state-action trajectories whose utility depends directly on the underlying 3D assets. Real-world corpora---Open X-Embodiment~\cite{openxembodiment2024}, DROID~\cite{khazatsky2024droid}, RH20T~\cite{fang2024rh20t}, BridgeData~V2~\cite{walke2023bridgedata}---provide high-fidelity trajectories but are costly to scale. Simulation-based benchmarks leverage 3D assets for demonstrations at scale: RoboTwin~\cite{mu2024robotwin} and RoboCasa~\cite{nasiriany2024robocasa} integrate generative models to create diverse environments, while ManiSkill2/3~\cite{tao2024maniskill3}, RLBench~\cite{james2020rlbench}, LIBERO~\cite{liu2023libero}, and CALVIN~\cite{mees2022calvin} offer controlled evaluation platforms. Together, these datasets connect asset generation with policy learning by turning geometric and physical realism into measurable differences in embodied data quality. MimicGen~\cite{mandlekar2023mimicgen} complements these by algorithmically augmenting existing demonstrations through spatial adaptation.

\subsection{Evaluation}
\label{subsec:evaluation}

Evaluating 3D assets for embodied AI goes beyond visual quality: outputs must not only \textit{look} correct but also \textit{behave} correctly under physics simulation and \textit{support} downstream robot learning. This motivates a three-level hierarchy in which higher levels subsume lower ones. Table~\ref{tab:eval_metrics} summarizes representative metrics across these levels.

\begin{table*}[tp]
\caption{Summary of evaluation metrics for 3D generation in embodied AI by evaluation level. \textcolor{green!60!black}{$\uparrow$}: higher is better; \textcolor{red}{$\downarrow$}: lower is better.}
\label{tab:eval_metrics}
\vspace{-5pt}
\footnotesize
\renewcommand{\arraystretch}{1.15}
\newcolumntype{Y}{>{\raggedright\arraybackslash}X}
\begin{minipage}{\textwidth}
\rowcolors{2}{gray!20}{white}
\begin{tabularx}{\textwidth}{@{} >{\raggedright\arraybackslash}m{1.6cm} >{\raggedright\arraybackslash}m{3.6cm} >{\centering\arraybackslash}m{0.5cm} Y >{\raggedright\arraybackslash}m{1cm} @{}}
\toprule
\textbf{Abbr.} & \textbf{Full Name} & & \textbf{Description} & \textbf{Ref.} \\
\midrule
\multicolumn{5}{@{}l}{\textit{\textbf{Geometry and Appearance Quality}}} \\
\midrule
\textbf{CD} & Chamfer Distance & \textcolor{red}{$\downarrow$} & Mean bidirectional nearest-neighbor distance between two point sets. & \cite{fan2017point} \\
\textbf{EMD} & Earth Mover's Distance & \textcolor{red}{$\downarrow$} & Minimum-cost optimal transport between two point distributions. & \cite{fan2017point} \\
\textbf{F-Score@$\tau$} & F-Score at Threshold $\tau$ & \textcolor{green!60!black}{$\uparrow$} & Harmonic mean of precision and recall of matched points at threshold $\tau$. & \cite{knapitsch2017tanks} \\
\textbf{IoU} & Intersection over Union & \textcolor{green!60!black}{$\uparrow$} & Volumetric overlap between predicted and ground-truth occupancy. & \cite{shapenets} \\
\textbf{FID} & Fr\'{e}chet Inception Distance & \textcolor{red}{$\downarrow$} & Distributional distance between multi-view generated and reference renders. & \cite{heusel2017gans} \\
\textbf{CLIP Score} & CLIP Similarity Score & \textcolor{green!60!black}{$\uparrow$} & Cosine similarity between CLIP embeddings of renders and text prompts. & \cite{radford2021learning} \\
\textbf{WT Ratio} & Watertight Ratio & \textcolor{green!60!black}{$\uparrow$} & Fraction of meshes that are closed, manifold, and self-intersection-free. & \cite{xiang2025structured} \\
\midrule
\multicolumn{5}{@{}l}{\textit{\textbf{Physical Plausibility and Sim-Readiness}}} \\
\midrule
\textbf{Stab. Rate} & Stability Rate & \textcolor{green!60!black}{$\uparrow$} & Fraction of objects remaining upright under gravity in simulation. & \cite{li2025dso} \\
\textbf{JT Acc.} & Joint Type Accuracy & \textcolor{green!60!black}{$\uparrow$} & Accuracy of predicted joint types (revolute/prismatic/fixed). & \cite{chen2024urdformer} \\
\textbf{JA Err.} & Joint Axis Error & \textcolor{red}{$\downarrow$} & Angular deviation (rad) of predicted vs.\ ground-truth joint axes. & \cite{chen2024urdformer} \\
\textbf{JO Err.} & Joint Origin Error & \textcolor{red}{$\downarrow$} & Euclidean distance (m) of predicted vs.\ ground-truth joint origins. & \cite{chen2024urdformer} \\
\textbf{JL Err.} & Joint Limit Error & \textcolor{red}{$\downarrow$} & Deviation in predicted range-of-motion bounds. & \cite{le2024articulate} \\
\textbf{Art. SR} & Articulation Success Rate & \textcolor{green!60!black}{$\uparrow$} & Binary success: position ${\leq}$50\,mm and angle ${\leq}$0.25\,rad. & \cite{le2024articulate} \\
\textbf{Penet. Vol.} & Penetration Volume & \textcolor{red}{$\downarrow$} & Intersection volume between components in multi-part assemblies. & \cite{luo2025physpart} \\
\textbf{CF Ratio} & Collision-Free Ratio & \textcolor{green!60!black}{$\uparrow$} & Fraction of assemblies free of inter-part collision. & \cite{luo2025physpart} \\
\textbf{Mat. Err.} & Material Error & \textcolor{red}{$\downarrow$} & Per-attribute error for physical properties (modulus, friction, density). & \cite{cao2025physx3d} \\
\midrule
\multicolumn{5}{@{}l}{\textit{\textbf{Embodied Task Performance}}} \\
\midrule
\textbf{Grasp SR} & Grasp Success Rate & \textcolor{green!60!black}{$\uparrow$} & Fraction of successful grasp-and-lift trials on generated objects. & \cite{seed2025seed3d} \\
\textbf{Art. Manip. SR} & Art. Manipulation SR & \textcolor{green!60!black}{$\uparrow$} & Fraction of successful articulated manipulation trials (e.g., open drawer). & \cite{wang2025embodiedgen} \\
\textbf{Nav. SR} & Navigation Success Rate & \textcolor{green!60!black}{$\uparrow$} & Fraction of navigation episodes reaching the target location. & \cite{savva2019habitat} \\
\textbf{SPL} & Success weighted by PL & \textcolor{green!60!black}{$\uparrow$} & Success weighted by inverse path length; penalizes long trajectories. & \cite{anderson2018spl} \\
\textbf{TCR} & Task Completion Rate & \textcolor{green!60!black}{$\uparrow$} & Fraction of task-conditioned environments where the task is completed. & \cite{mesatask} \\
\textbf{S2R SR} & Sim-to-Real Success Rate & \textcolor{green!60!black}{$\uparrow$} & Real-robot success rate after training exclusively on synthetic data. & \cite{graspvla} \\
\bottomrule
\end{tabularx}
\end{minipage}
\end{table*}

\subsubsection{Geometry and Appearance Quality}
Standard geometric metrics (CD, EMD, F-Score, IoU) assess shape fidelity and remain indispensable baselines inherited from 3D reconstruction. For textured assets, CLIP Score and FID evaluate semantic alignment and distributional visual quality, respectively. However, these metrics miss key embodied failure modes: low-CD meshes may still contain self-intersections or non-manifold edges, \textit{watertight ratio}~\cite{xiang2025structured} is only a binary proxy, and all regions are weighted equally despite the importance of contact-critical geometry. Likewise, appearance metrics such as FID and CLIP Score correlate weakly with physical fidelity and stable grasping.

\subsubsection{Physical Plausibility and Sim-Readiness}
Physical metrics are less standardized but increasingly central for embodied applications. Object-level metrics include \textit{stability rate}~\cite{li2025dso} for rigid bodies, kinematic accuracy (joint type, axis, origin, and limit errors)~\cite{chen2024urdformer, le2024articulate} for articulated objects, \textit{penetration volume} and \textit{collision-free ratio}~\cite{luo2025physpart} for assemblies, and \textit{material error}~\cite{cao2025physx3d} for physical properties. Scene-level metrics include collision rate, out-of-floor-plan rate~\cite{physcene}, object reachability~\cite{physcene}, stable-object ratio~\cite{sage}, and Physical Plausibility Score~\cite{pat3d}. Their main limitation is \textit{simulator dependence}: assets stable in MuJoCo may show artifacts in Isaac Sim, yet most works evaluate in only one engine. In addition, most metrics remain static or quasi-static and thus miss dynamic interactions such as impacts, sliding, and deformation.

\subsubsection{Embodied Task Performance}
Downstream task success is the most direct validation of generated asset quality, integrating geometric, physical, and semantic requirements into one scalar. At the object level, \textit{grasp SR}~\cite{seed2025seed3d} and \textit{articulated manipulation SR}~\cite{wang2025embodiedgen, cao2025physx} verify collision mesh and URDF accuracy for contact-rich interaction, while \textit{sim-to-real SR}~\cite{graspvla} measures whether policies trained on generated assets transfer to real robots. At the scene level, \textit{navigation SR} and \textit{SPL}~\cite{anderson2018spl} remain standard for Habitat and AI2-THOR benchmarks, and \textit{task completion rate}~\cite{mesatask} evaluates end-to-end task execution in generated environments. However, these metrics confound asset, simulator, and policy quality, and controlled ablations are still rare. They are also expensive, often requiring policy training, many rollouts, and simulator tuning, which limits comparison and fragments benchmarks.

%% file: sec/challenges.tex
\section{Challenges and Future Directions}
\label{sec:challenges}

Despite rapid progress in the surveyed research directions, significant challenges remain before 3D asset generation serves as a reliable, scalable foundation for embodied AI and robotic simulation.

\subsection{Physical Grounding at Scale}
The gap between geometric and physically grounded 3D generation appears at two levels: \emph{data} and \emph{model architecture}. On the data side, large-scale repositories such as Objaverse~\cite{deitke2023objaverse} and Objaverse-XL~\cite{deitke2023objaversexl} provide millions of geometric models but lack material properties, mass distributions, and kinematic parameters. Physically annotated datasets---AKB-48~\cite{liu2022akb}, PhysXNet~\cite{cao2025physx3d}, PhysX-Mobility~\cite{cao2025physx}---are far smaller, limiting the generalization of models like PhysX-3D~\cite{cao2025physx3d} and SOPHY~\cite{cao2025sophy}. Scaling annotation without proportional human labor requires LLM/VLM-assisted labeling (whose out-of-distribution accuracy remains unvalidated), self-supervised annotation via differentiable simulators, and synthetic bootstrapping through procedural pipelines such as Infinite Mobility~\cite{lian2025infinite}. On the model side, most generative architectures optimize visual metrics without awareness that outputs will undergo rigid-body simulation: non-manifold surfaces, self-intersecting meshes, and thin structures that render acceptably may be unusable in physics engines. Integrating differentiable physics constraints into training---as in DSO~\cite{li2025dso} for stability and PhysPart~\cite{luo2025physpart} for contact compatibility---is the most principled direction, but the cost of differentiable simulation grows substantially with scene complexity. A practical intermediate step is lightweight proxy validators that reject physically implausible outputs at generation time without full simulator rollouts. Moreover, vision--language foundation models show early promise in estimating coarse physical attributes (material category, relative density, friction regime) from appearance alone, potentially providing zero-shot physical priors that complement the sparse annotated datasets available today.

\subsection{Deformable and Dynamic Asset Generation}
Unlike rigid and articulated objects, deformable asset generation remains much less mature. Cloth, ropes, soft bodies, and food are central to robotic manipulation, yet their high-dimensional geometry and reliance on continuum mechanics (FEM, MPM, PBD) make generation and simulation fundamentally harder. Current progress concentrates on garments, where sewing-pattern-based pipelines~\cite{he2024dresscode, li2025dress} produce simulator-compatible rest shapes, but diversity is still limited by template coverage. General soft-body generation is largely open: methods like PhysGaussian~\cite{xie2024physgaussian} and PhysDreamer~\cite{zhang2024physdreamer} estimate material fields per instance but rely on costly optimization and do not scale. Key gaps include the absence of large-scale deformable datasets with material annotations, the speed--fidelity tension between deformable solvers and training-time requirements, and the mismatch between generative outputs (meshes/Gaussians) and particle-based discretizations. A promising lesson from garment generation is that the sewing pattern serves as a \emph{simulation-native intermediate representation} that is generatable and directly consumable by cloth solvers. Discovering analogous intermediates for general soft bodies---such as material point clouds annotated with constitutive parameters---could unlock scalable deformable asset generation beyond garments.

\subsection{Efficiency--Controllability Trade-off in Scene Generation}
Learning-based scene synthesis~\cite{atiss, diffuscene, physcene} is efficient but struggles with semantic consistency and long-range spatial coherence, while agentic approaches~\cite{sage, scenesmith, scenethesis} achieve stronger semantic alignment through iterative LLM-driven refinement at high computational cost. The key future direction is making agentic generation efficient enough for large-scale embodied AI training. Two concrete strategies stand out. First, \emph{amortized verification}: replacing full physics simulation at each agentic iteration with lightweight learned critics---as demonstrated by SAGE's semantic and physical plausibility scorers~\cite{sage}---can reduce per-step validation cost by orders of magnitude. Second, \emph{hierarchical generation}: using efficient diffusion or feed-forward models to produce a coarse scene layout and then restricting agentic refinement to task-critical regions avoids iterating over the entire scene, concentrating computation where it matters most.

\subsection{Evaluation Standardization}
As discussed in Sec.~\ref{subsec:evaluation}, evaluation protocols remain fragmented: geometric metrics ignore physical validity, physical metrics are used inconsistently, and downstream task evaluation depends on simulator and policy choices. Standardized benchmark suites are therefore needed for meaningful comparison. We advocate a three-level protocol covering (1)~\emph{format validity}, (2)~\emph{physics plausibility}, and (3)~\emph{downstream task performance}, together with \emph{cross-simulator consistency tests} to expose solver-dependent artifacts concealed by single-simulator evaluation.

\subsection{Closing the Sim-to-Real Loop}
A substantial sim-to-real gap persists across three dimensions: \textit{appearance} (lighting, reflectance, sensor noise), \textit{dynamics} (contact, deformation, inertia), and \textit{semantics} (object arrangement, task context, clutter). Closing these gaps requires tighter integration between generation and real-world feedback. Adaptive digital twin frameworks that continuously update simulation parameters from real interaction data~\cite{jiang2025phystwin, twinaligner} point in a promising direction. More broadly, the field's primary structural limitation is \emph{fragmentation}: generative models produce meshes or Gaussians for rendering, physics engines require watertight collision geometry with physical parameters, and policy learning demands task-relevant semantics---all developed in separate communities with incompatible representations. The long-term direction is unified foundation models jointly reasoning over geometry, physics, and task semantics. Representations like PhysGaussian~\cite{xie2024physgaussian}, where primitives simultaneously serve rendering and simulation, and pipelines like EmbodiedGen~\cite{wang2025embodiedgen} and PhysX-Anything~\cite{cao2025physx} that jointly produce geometry, kinematics, and physics from a single input, are early steps. Realizing such unification at the scale required for general embodied AI remains the field's defining open problem.

%% file: sec/conclusion.tex
\section{Conclusion}
\label{sec:conclusion}
This survey presented a simulation-centric review of 3D generation for embodied AI, organized around three roles: \emph{Data Generator}, producing simulation-ready assets; \emph{Simulation Environments}, building interactive worlds; and \emph{Sim2Real Bridge}, supporting real-world transfer. Within each role, we traced a progression from appearance-oriented generation toward physics-aware, simulator-compatible outputs: data generators increasingly produce assets with physical annotations and kinematic structures; scene-level methods integrate physical and semantic constraints into layout synthesis; and sim-to-real approaches leverage generative models to narrow the domain gap in both appearance and dynamics. Across all three, a consistent trend emerges: the goal of 3D generation has shifted from visual plausibility toward \emph{simulation readiness}, making generation a core infrastructure layer for embodied learning. Key challenges remain, including the scarcity of physical annotations, the gap between geometric realism and simulator deployability, limited support for deformable and dynamic assets, fragmented evaluation standards, and the persistent sim-to-real gap; most fundamentally, the current ecosystem remains modular and disjoint, with generative models, physics engines, and robot learning systems optimized separately and connected through brittle conversion pipelines. Looking forward, we expect these three roles to converge toward unified generation--simulation foundations that jointly model geometry, physics, and task semantics, and hope this survey provides a roadmap for that transition.

%% file: sec/biography.tex
\begin{IEEEbiography}
[{\includegraphics[width=1in,height=1.25in,clip,keepaspectratio]{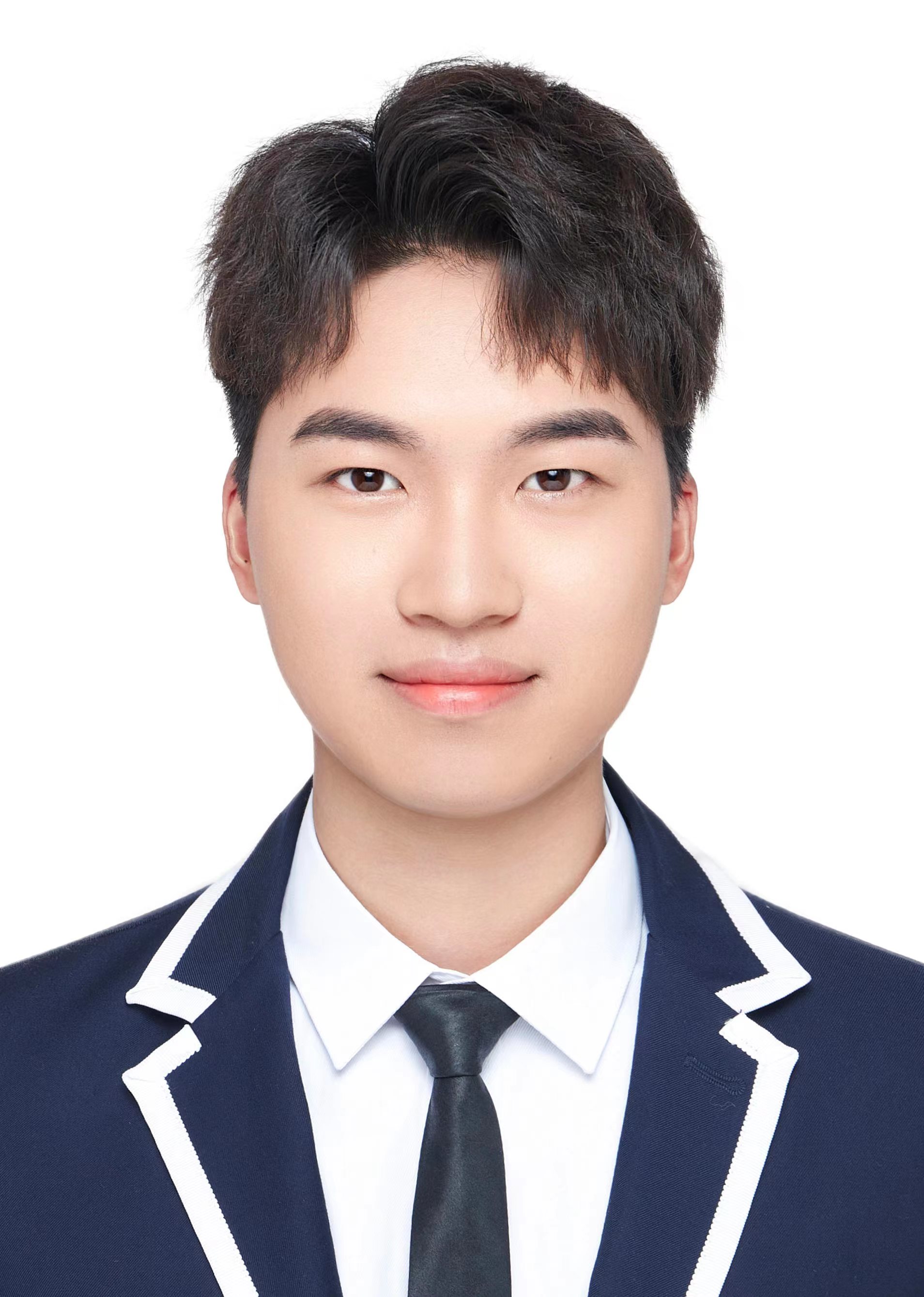}}]
{Tianwei Ye} received the bachelor's degree from the School of Electronic Information, Central South University, China, in 2024. He is currently a master's student with the Electronic Information School, Wuhan University, China. His research interests include shape analysis, 3D vision, and embodied AI.
\end{IEEEbiography}

\begin{IEEEbiography}
[{\includegraphics[width=1in,height=1.25in,clip,keepaspectratio]{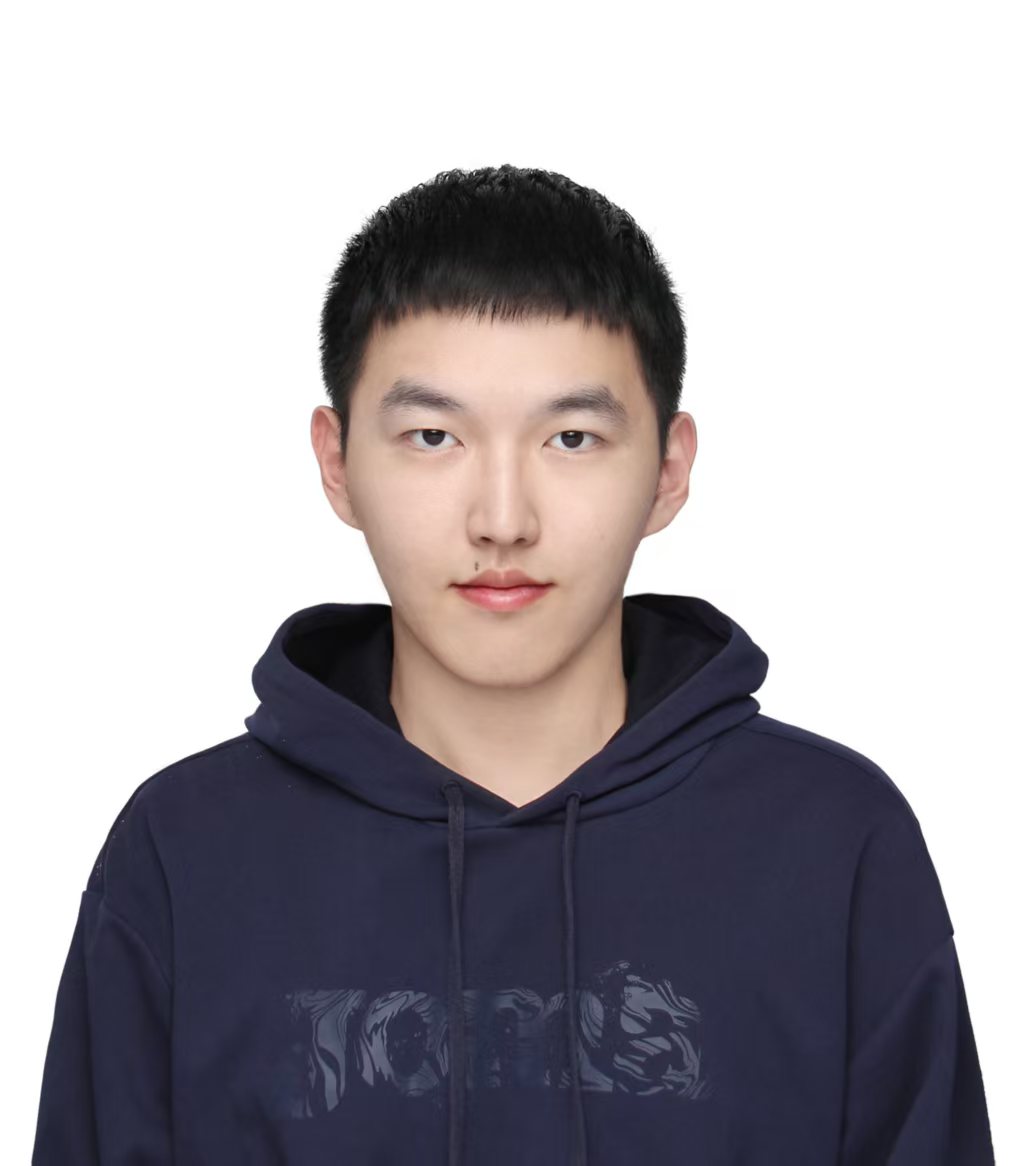}}]
{Yifan Mao} received his B.S. degree in 2024 from Jilin University, Changchun, China. He is currently an M.S. student with the School of Computer Science and Technology, Harbin Institute of Technology. His current research interests are 3D perception.
\end{IEEEbiography}

\begin{IEEEbiography}
[{\includegraphics[width=1in,height=1.25in,clip,keepaspectratio]{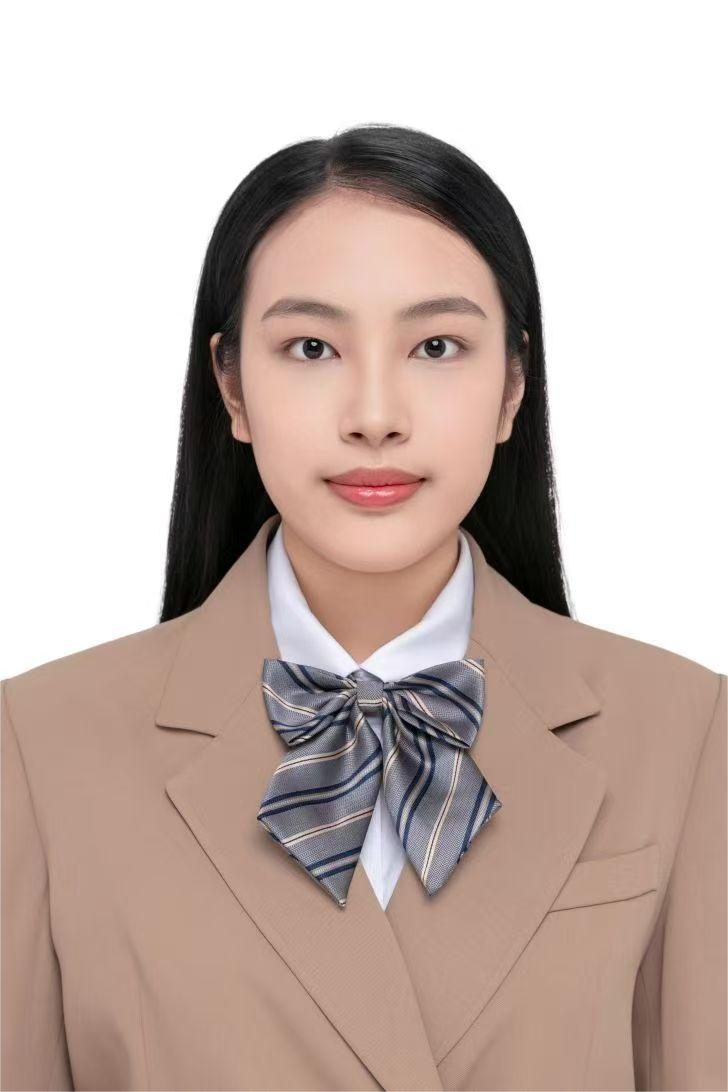}}]
{Minwen Liao} is currently working toward the B.E. degree in electronic information engineering at Xinjiang University, Urumqi, China. Her research interests include computer vision and robotics, with a particular focus on dexterous manipulation and world models. She has co-authored several papers accepted by top-tier conferences and journals, including IEEE/CVF International Conference on Computer Vision (ICCV) and the Conference on Robot Learning (CoRL).
\end{IEEEbiography}

\begin{IEEEbiography}
[{\includegraphics[width=1in,height=1.25in,clip,keepaspectratio]{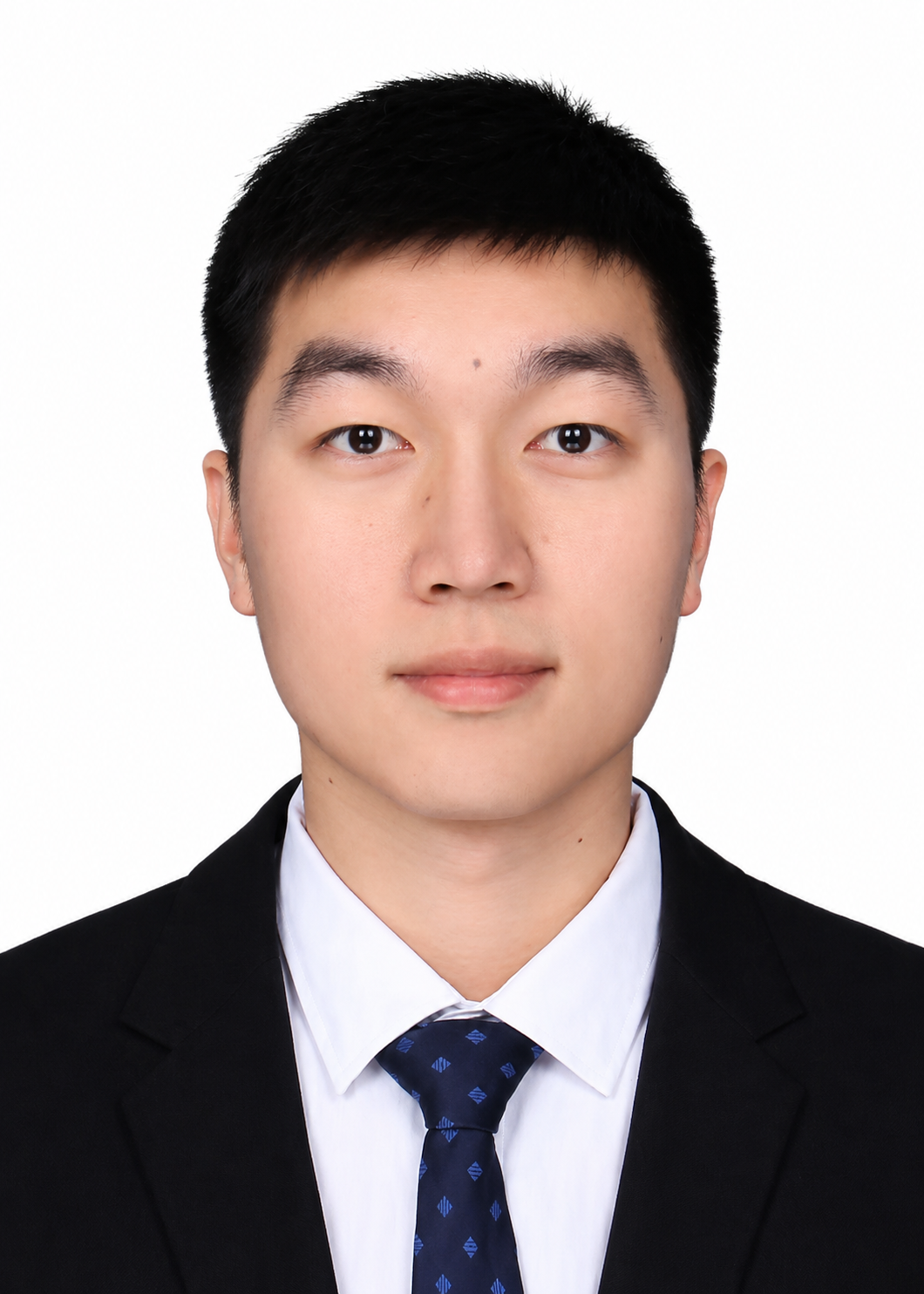}}]
{Jian Liu} received the B.E. degree in Software Engineering and the M.S. degree in Computer Science from Harbin Institute of Technology, Harbin, China, in 2022 and 2024, respectively. He is currently a Ph.D. student in Computer Science and Engineering at the Hong Kong University of Science and Technology, advised by Prof. Song Guo. His research interests include mesh generation, 3D generative models, and embodied AI.
\end{IEEEbiography}

\begin{IEEEbiography}
[{\includegraphics[width=1in,height=1.25in,clip,keepaspectratio]{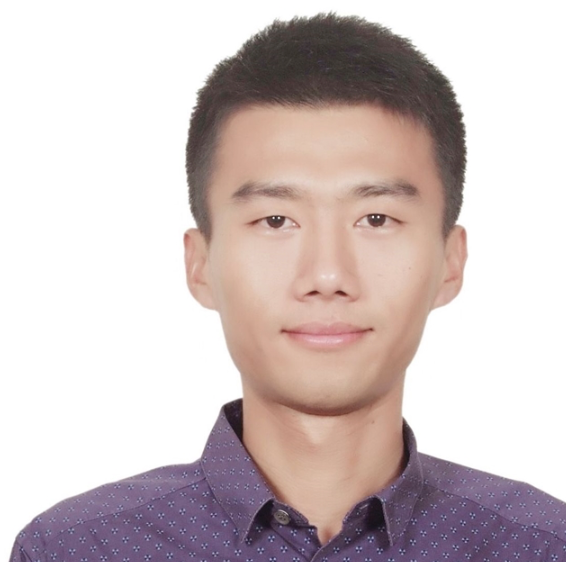}}]
{Chunchao Guo} is an Independent Researcher. His research focuses on advertising AI, generative large models, and their practical applications. He has published over 40 papers in top-tier conferences and journals such as CVPR, and won over 20 championships in authoritative technical competitions.
\end{IEEEbiography}

\begin{IEEEbiography}
[{\includegraphics[width=1in,height=1.25in,clip,keepaspectratio]{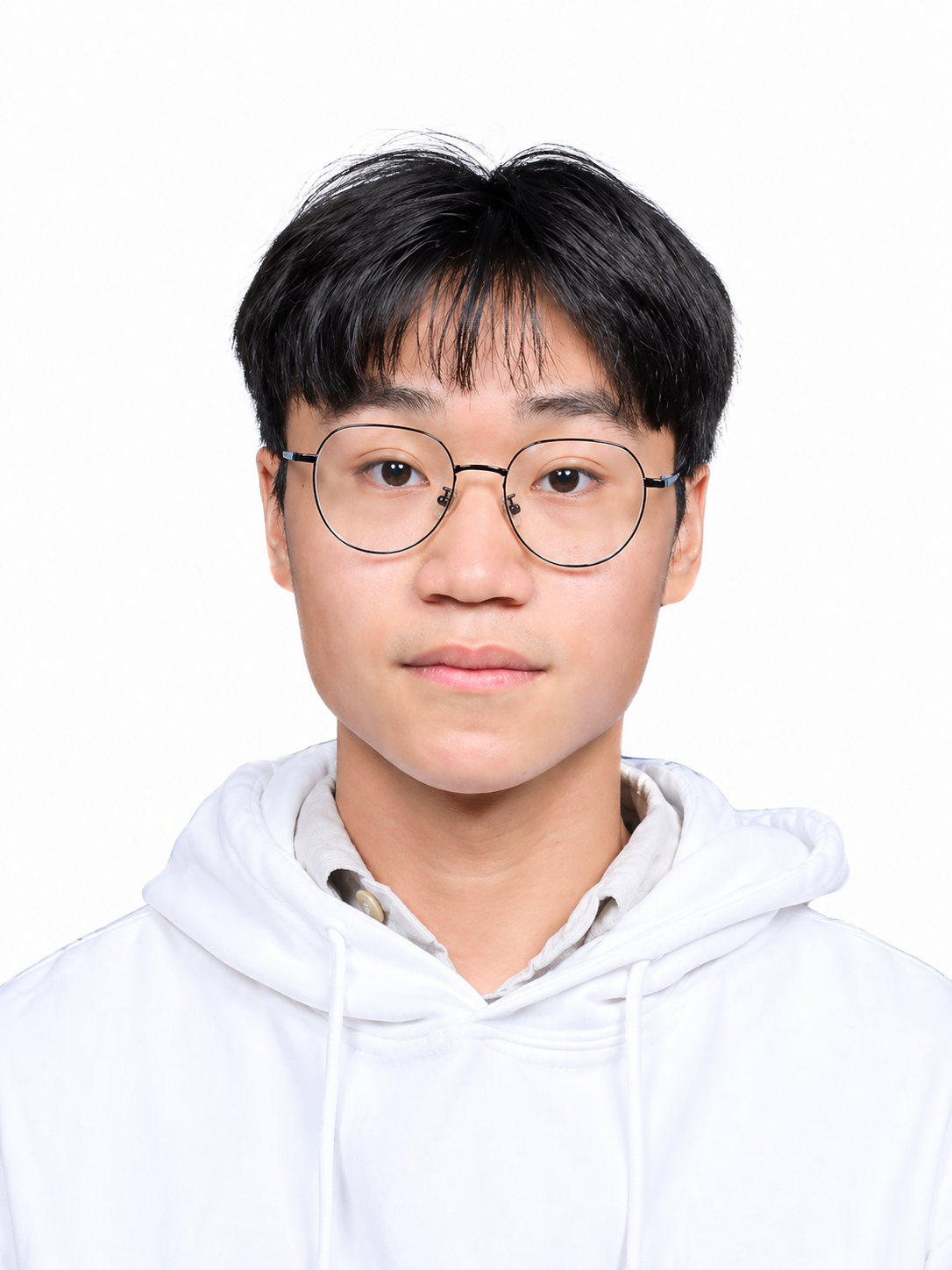}}]
{Dazhao Du} is currently a Ph.D. student at the Hong Kong University of Science and Technology. He received his bachelor's degree from the Department of Automation at Tsinghua University and his master's degree from the Institute of Software, Chinese Academy of Sciences. His research interests include MLLM, video understanding, and computer vision.
\end{IEEEbiography}

\begin{IEEEbiography}
[{\includegraphics[width=1in,height=1.25in,clip,keepaspectratio]{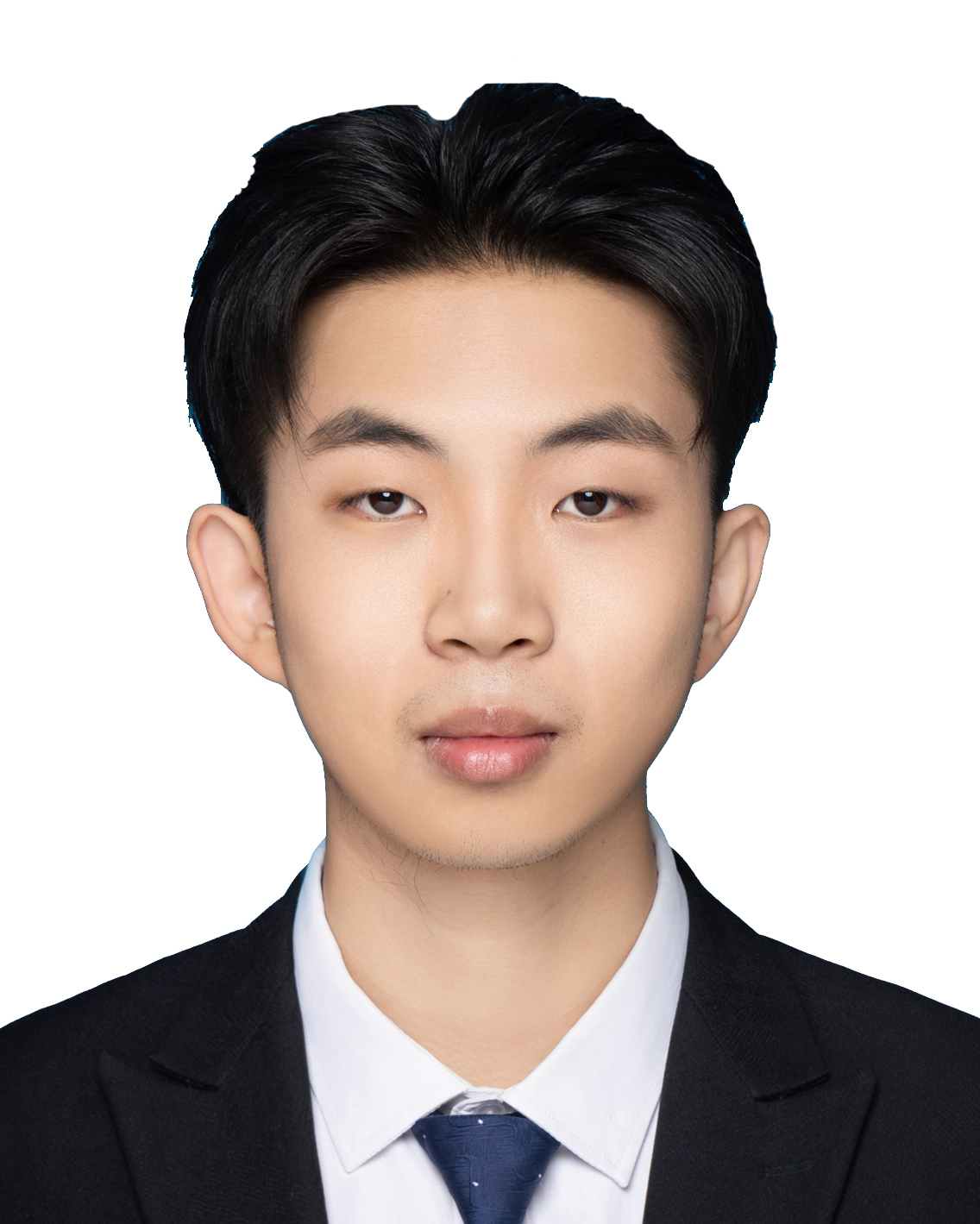}}]
{Quanxin Shou} received the B.E. degree from the School of Automation Science and Electrical Engineering, Beihang University, Beijing, China. He is currently pursuing the Ph.D. degree with the Department of Computer Science and Engineering, The Hong Kong University of Science and Technology, Hong Kong, under the supervision of Prof. Song Guo. His research interests include robot learning and multimodal large language models.
\end{IEEEbiography}

\begin{IEEEbiography}
[{\includegraphics[width=1in,height=1.25in,clip,keepaspectratio]{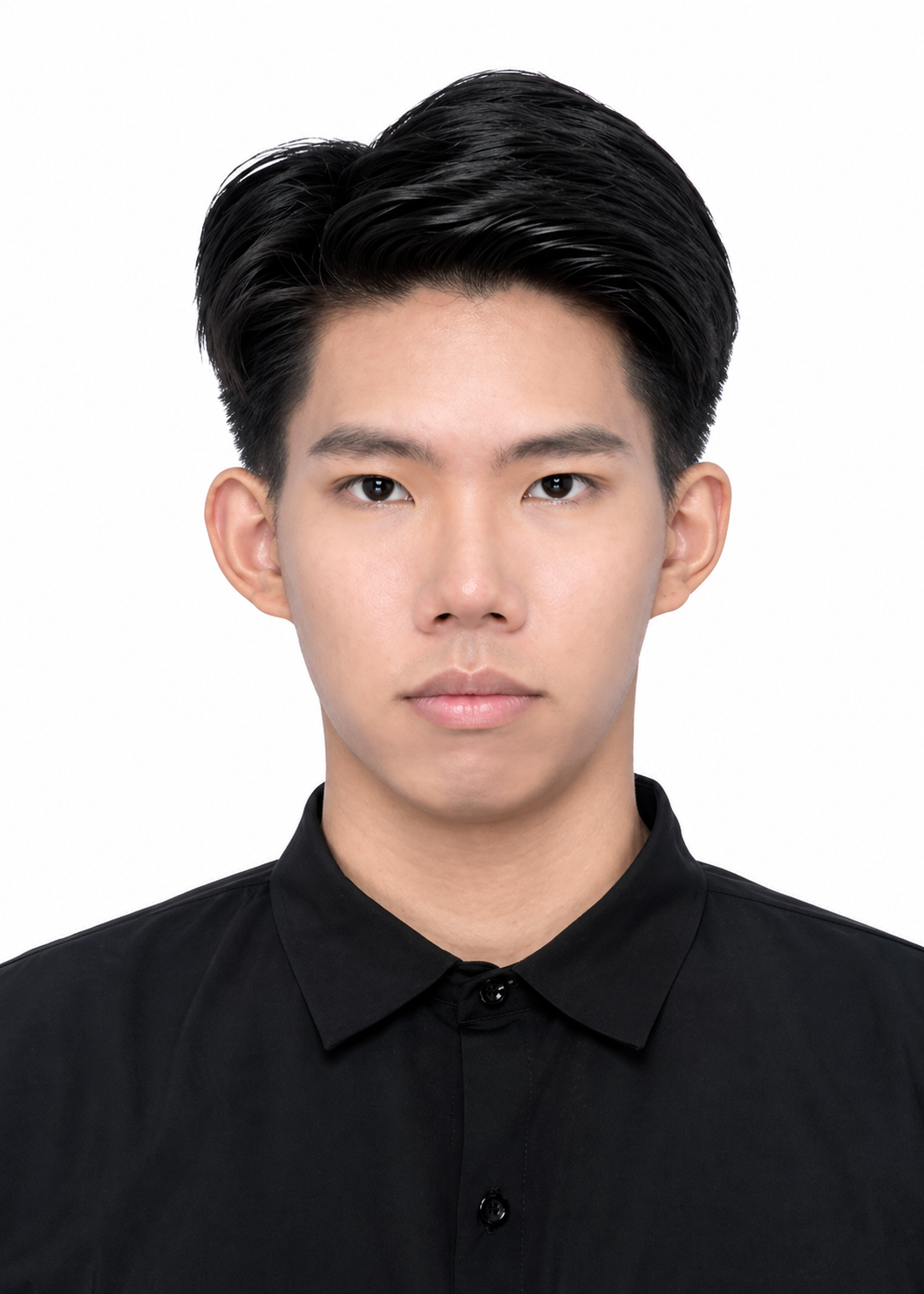}}]
{Fangqi Zhu} received the B.Eng. degree in Software Engineering from Wuhan University, Wuhan, China, in 2021, and the M.Eng. degree in Computer Science and Technology from Harbin Institute of Technology (Shenzhen), Shenzhen, China, in 2024. He is currently pursuing the Ph.D. degree with the Department of Computer Science and Engineering, The Hong Kong University of Science and Technology, Hong Kong, under the supervision of Prof. Song Guo. His research interests include world models and robot manipulation.
\end{IEEEbiography}

\begin{IEEEbiography}
[{\includegraphics[width=1in,height=1.25in,clip,keepaspectratio]{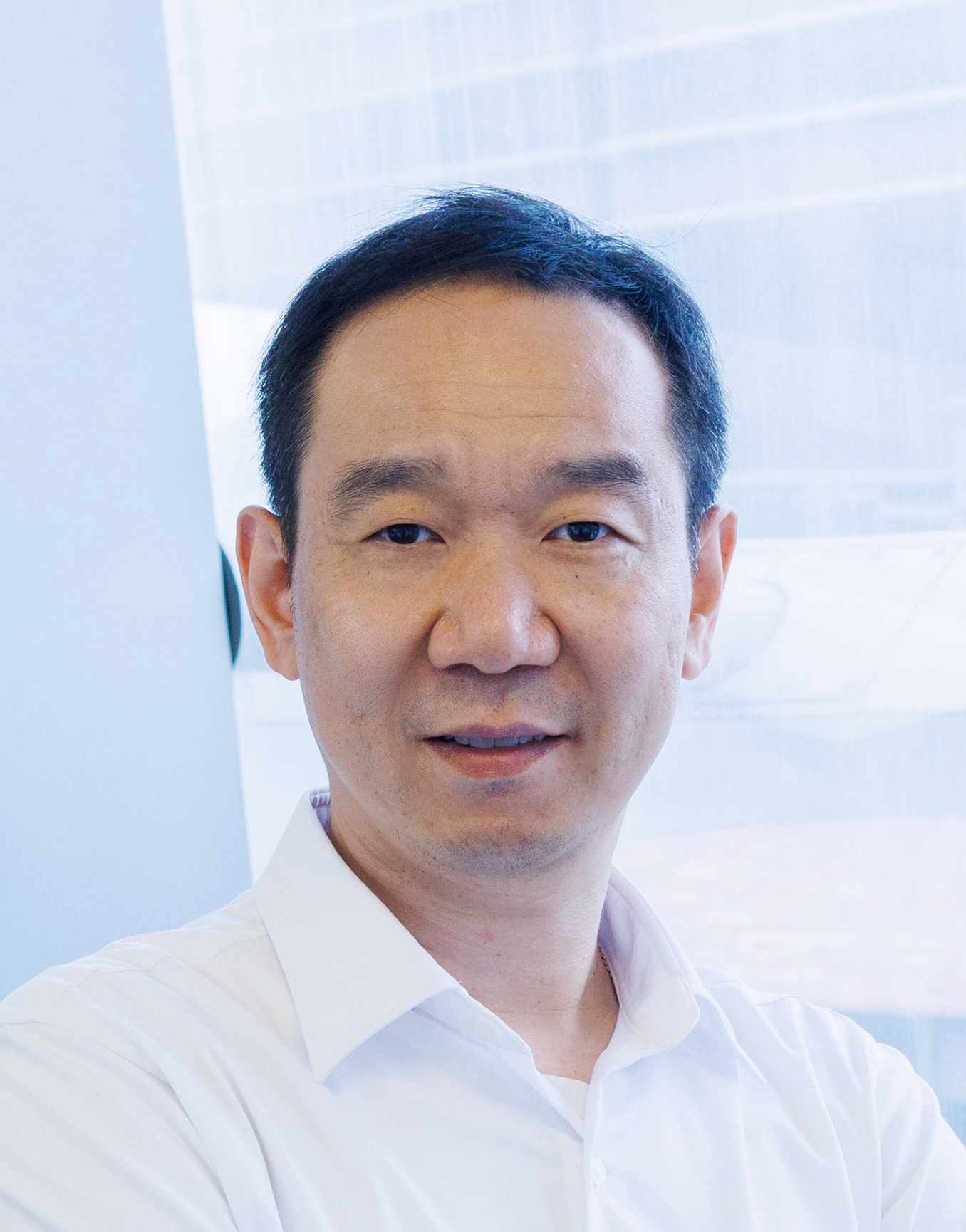}}]
{Song Guo} is a Chair Professor in the Department of Computer Science and Engineering at the Hong Kong University of Science and Technology. He also holds a Changjiang Chair Professorship awarded by the Ministry of Education of China. His research interests are mainly in Large Language Models, Edge AI, and Machine Learning Systems. As a Highly Cited Researcher (Clarivate Web of Science), he published many papers in top venues with wide impact and received over a dozen Best Paper Awards from IEEE/ACM conferences, journals, and technical committees. He is the recipient of the Edward J. McCluskey Technical Achievement Award in 2024, First Prize in Natural Science (China Electronics Society) in 2023, Gold Medal of Geneva Inventions Expo in 2024 \& 2023, etc. He is a Fellow of the Canadian Academy of Engineering (FCAE), Member of Academia Europaea (MAE), Fellow of the IEEE (FIEEE), Distinguished Member of the ACM, and Fellow of the Asia-Pacific Artificial Intelligence Association (FAAIA).
\end{IEEEbiography}